%% file: arxiv.tex
\documentclass{article} %
\newcommand{\arxivversion}{}
\newcommand{\codeurl}{https://abdullahamdi.com/llmseg/}
\PassOptionsToPackage{dvipsnames,table}{xcolor}
\usepackage{iclr2027_conference,times}

\input{math_commands.tex}

\usepackage{graphicx}
\usepackage{wrapfig}
\usepackage{amsmath}
\usepackage{amssymb}
\usepackage{amsfonts}
\usepackage{booktabs}
\usepackage{tabularx}
\usepackage{multirow}
\usepackage{xcolor}
\usepackage{microtype}
\usepackage{nicefrac}
\usepackage{pifont}
\usepackage{comment}
\usepackage{soul}
\usepackage{xspace}
\usepackage{subcaption}
\usepackage{tikz}
\usetikzlibrary{spy}          %
\usepackage{makecell}
\usepackage[most]{tcolorbox}  %
\usepackage{listings}         %
\usepackage{hyperref}
\usepackage{url}
\input{commands.tex}

\definecolor{Klein_Blue}{rgb}{0.0, 0.129, 0.6}
\usepackage{hyperref}       %
\hypersetup{
    colorlinks=true,
    linkcolor=magenta,
    urlcolor=magenta,
    citecolor=Klein_Blue
    }
\definecolor{PromptBg}{HTML}{F6F6F3}
\definecolor{PromptRule}{HTML}{4B527E}
\definecolor{PromptNote}{HTML}{1F7A3D}
\lstdefinestyle{promptstyle}{
  basicstyle=\fontsize{6}{6.5}\selectfont\ttfamily,
  morecomment=[l]{\#},
  commentstyle=\color{PromptNote},
  columns=fixed, keepspaces=true, showstringspaces=false,
  breaklines=false, aboveskip=0pt, belowskip=0pt, xleftmargin=0pt,
}
\newtcolorbox{promptbox}[2][]{
  enhanced, breakable,
  center title,
  left*=0pt, right*=0pt,
  boxsep=1pt, left=4pt, right=4pt, top=2pt, bottom=2pt,
  skin first=enhanced, skin middle=enhanced, skin last=enhanced,
  colframe=PromptRule!85, colback=PromptBg,
  fonttitle=\bfseries\rmfamily,
  fontupper=\fontsize{6}{6.5}\selectfont,
  before upper={\setlength{\topsep}{0pt}\setlength{\partopsep}{0pt}},
  title={\footnotesize\strut{#2}},
  #1}

\newcommand{\ours}{LLMSeg\xspace}          %
\newcommand{\samthree}{SAM~3\xspace}
\newcommand{\samagent}{\samthree Agent\xspace}
\newcommand{\gfl}{Gemini-3.7-Flash\xspace}  %
\newcommand{\gfli}{Gemini-3.1-Flash-Lite\xspace} %
\newcommand{\cgf}{cgF1\xspace}
\newcommand{\pmf}{pmF1\xspace}
\newcommand{\oursfull}{\ours (full)\xspace}          %
\newcommand{\oursnr}{\ours (w/o refine)\xspace}      %

\newcommand{\eqn}[1]{Eq.~(\ref{#1})}
\newcommand{\na}{{--}}                          %
\newcommand{\sd}[1]{{\scriptsize$\pm$#1}}
\newcommand{\ci}[2]{{\scriptsize[#1,\,#2]}}

\newlength{\qgw}
\newcommand{\qim}[1]{\includegraphics[width=\qgw]{imgs/#1}}
\newcommand{\qhd}[1]{{\scriptsize\textbf{#1}}}
\newcommand{\qlab}[2]{{\scriptsize\raggedright\textbf{#1}\newline\emph{``#2''}\par}}
\newlength{\qhdh}
\newcommand{\qcol}[6]{\begin{tabular}[b]{@{}c@{}}%
  \parbox[t][\qhdh][t]{\qgw}{\centering\scriptsize\textbf{#1}\\[-1pt]%
    {\tiny #2: \emph{``#3''}}}\\
  \qim{#4}\\ \qim{#5}\\ \qim{#6}\end{tabular}}
\newcommand{\qrow}[1]{\rotatebox{90}{\makebox[0.75\qgw]{\scriptsize\textbf{#1}}}}
\newcommand{\wlab}[2]{{\scriptsize\raggedright\textbf{#1}\newline #2\par}}
\newcommand{\bqry}[1]{{\scriptsize\emph{``#1''}}}

\title{\ours: Training-Free Text-Prompted Image Segmentation with B-Splines}

\author{Abdullah Hamdi\\
King Abdullah University of Science and Technology (KAUST)\\
}

\makeatletter
\def\@maketitle{\vbox{\hsize\textwidth\centering
{\LARGE\sc \@title\par}
\vskip 0.2in
{\bf \@author\par}
\vskip 0.3in minus 0.1in}}
\makeatother
\iclrfinalcopy %
\begin{document}

\maketitle
\lhead{} \renewcommand{\headrulewidth}{0pt} %
\thispagestyle{plain}

\input{sections/0.abstract.tex}

\begin{figure}[!ht]
  \centering
  \includegraphics[width=\textwidth]{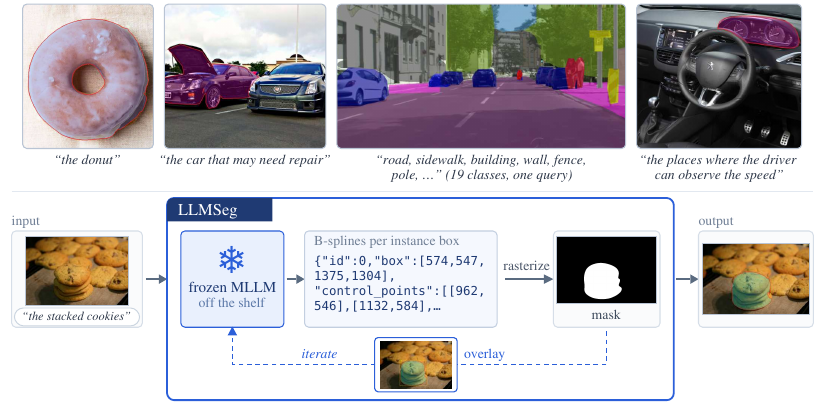}
  \vspace{-6pt}
  \caption{\small \textbf{Text-prompted segmentation from a frozen MLLM.} \ours segments an image from a free-form text query alone, with a
  frozen multimodal LLM as its only vision component: no segmentation head, no
  mask decoder, no training. \textit{Top:} four results, each under its query.
  \textit{Bottom:} the model emits geometry rather than pixels; deterministic
  code rasterizes it, and the rendered overlay is fed back so the model can
  inspect and correct its own boundary.}
  \label{fig:teaser}
\end{figure}

\setcounter{topnumber}{1}\setcounter{totalnumber}{1}\setcounter{bottomnumber}{0}
\input{sections/1.introduction.tex}
\input{sections/2.related.tex}
\input{sections/3.methodology.tex}

\input{sections/4.experiments.tex}

\input{sections/6.analysis.tex}
\input{sections/7.conclusion.tex}

\bibliography{egbib}
\bibliographystyle{iclr2027_conference}

\clearpage \clearpage 
\appendix
\input{sections/supp.tex}

\end{document}

%% file: math_commands.tex
\usepackage{amsmath,amsfonts,bm}

\def\eqref#1{equation~\ref{#1}}

\def\1{\bm{1}}

\DeclareMathAlphabet{\mathsfit}{\encodingdefault}{\sfdefault}{m}{sl}
\SetMathAlphabet{\mathsfit}{bold}{\encodingdefault}{\sfdefault}{bx}{n}

%% file: commands.tex
\definecolor{green}{RGB}{0,150,10}

\newcommand{\cmark}{\textcolor{blue}{\ding{51}}}%
\newcommand{\xmark}{\textcolor{red}{\ding{55}}}%

\newcommand{\figLabel}{Figure~}

\newcommand{\secLabel}{Section~}
\newcommand{\tblLabel}{Table~}
\newcommand{\mysection}[1]{\noindent\textbf{#1.}}

\newlength\mytmplen

\providecommand{\eg}{e.g.,\xspace}   %
\definecolor{GTAmber}{RGB}{214,133,0}  %
\newcommand{\gtamber}[1]{\textcolor{GTAmber}{#1}}

%% file: sections/0.abstract.tex
\begin{abstract}
Promptable segmentation models such as SAM~3 achieve remarkable open-vocabulary
performance but require dedicated architectures, large-scale annotation
engines, and task-specific training. We ask a different question: can a
general-purpose multimodal large language model (MLLM) segment images
\emph{directly}, without an added segmentation head, mask decoder, or training?
We present \ours, a training-free pipeline that turns a vision-capable LLM
into a text-prompted image segmenter. \ours decomposes segmentation into three
LLM-driven stages. First, \emph{detect} localizes all instances matching a
free-form text query and describes their coarse morphology. Next,
\emph{segment} traces each instance using closed periodic B-splines confined to
its detected box. These splines define an outer boundary and optional hole
loops. Finally, \emph{refine} iteratively proposes focus boxes around residual
errors and retraces confined splines. The process ends when the mask stabilizes
or the pass budget is exhausted.
The output is a binary mask with inspectable intermediate geometry.
We evaluate \ours across ten referring and reasoning segmentation splits
(ReasonSeg and RefCOCO/+/g). We also evaluate open-vocabulary instance
segmentation (LVIS, COCO, and SA-Co), semantic segmentation, and zero-shot
detection (ODinW13 and RF-100VL). With an off-the-shelf \gfl backend, \ours is
competitive with and often surpasses the specialist SAM~3 model on identical
items. It reaches 71.5 cIoU on ReasonSeg~val without additional
segmentation-specific training. Backend model updates can improve accuracy without
retraining the pipeline.
\ifdefined\arxivversion
The code is available at \href{\codeurl}{\codeurl}.
\fi
\end{abstract}

%% file: sections/1.introduction.tex
\section{Introduction} \label{sec:introduction}
\vspace{-5pt}
Text-prompted segmentation (producing a pixel mask for every instance that
matches a language query) is a foundational capability for multimodal AI.
The dominant approach builds ever-stronger \emph{specialist} models: the SAM series
\citep{kirillov2023segment,ravi2024sam2segmentimages} introduced promptable
visual segmentation. SAM~3 \citep{carion2025sam} extends this capability to
promptable \emph{concept} segmentation from noun phrases. It relies on a
dedicated detector-tracker architecture and a massive human- and
model-in-the-loop data engine.

In parallel, multimodal large language models (MLLMs) have acquired remarkable
general visual competence. Their capabilities include grounding, counting, and
coarse localization \citep{comanici2025gemini,GPT4,bai2025qwen3,deitke2025molmo}.
However, they are widely assumed to require a trained segmentation decoder,
such as SAM, to produce dense pixel masks
\citep{lai2024lisa,rasheed2024glamm,carion2025sam}.
We challenge this assumption. We present \ours (\figLabel\ref{fig:teaser}),
a fully training-free pipeline. A general-purpose MLLM is its \emph{only}
vision component. It returns \emph{geometry}: closed curves that deterministic
code rasterizes into a mask and keeps inside each box. The work splits into
three stages. \emph{Detect} localizes every instance the query matches and
returns its box. \emph{Segment} traces that instance with closed periodic
B-spline loops confined to the box. \emph{Refine} reads the rendered overlay
and retraces the errors that remain, each inside a signed \emph{focus box}.

This design provides an explicit \emph{parametric} intermediate representation:
cubic traces encode smooth curves, while degree-one traces encode straight edges.
Rasterization and refinement need not preserve a smooth final boundary. With
topology enabled, holes are specified as subtractive loops.
Reasoning queries (``the food that the bird likes to eat'') work off the shelf
because each query is interpreted by the same LLM that performs the segmentation.
Every stage emits human-readable JSON, making geometric decisions
inspectable. Since \ours adds no trained parameters, backend upgrades require no pipeline
retraining and usually come with segmentation accuracy gains.

\noindent\textbf{Contributions:}
\textbf{(i)} We propose \ours, a training-free promptable segmentation pipeline
in which an MLLM predicts parametric B-spline geometry confined to each detected box and
then refines it inside signed focus boxes.
\textbf{(ii)} We show for the first time at benchmark scale that a
general-purpose MLLM can perform dense text-prompted segmentation without any
\emph{trained segmentation component}. \ours with \gfl as the backend is
competitive with \samthree \citep{carion2025sam} on ReasonSeg, RefCOCO/+/g, LVIS,
COCO, SA-Co, ADE20K, Cityscapes, ODinW13, and RF-100VL, and reaches 71.5 cIoU on
ReasonSeg~val, the best among methods not fine-tuned for the task. 
Further analysis shows \ours's segmentation quality follows the backend strength without retraining, offering automatic improvement in segmentation with better backends.

%% file: sections/2.related.tex
\section{Related Work} \label{sec:related}
\vspace{-5pt}

\mysection{Segmentation models and MLLM-driven segmentation}
SAM \citep{kirillov2023segment} introduced promptable segmentation.
SAM~2 \citep{ravi2024sam2segmentimages} extended it to video.
\samthree \citep{carion2025sam} extended it to open-vocabulary concept
prompts.
GLIP \citep{li2022grounded}, GroundingDINO \citep{groundingdino},
OWLv2 \citep{minderer2024scalingopenvocabularyobjectdetection}, and
DINO-X \citep{ren2025dinoxunifiedvisionmodel} are trained specialists.
MLLM-based methods add language reasoning but retain a trained mask
component. LISA \citep{lai2024lisa} decodes an LLM's
\texttt{<SEG>} token with SAM. GLaMM \citep{rasheed2024glamm}, PSALM
\citep{zhang2024psalm}, and X-SAM \citep{wang2025x} also use trained
mask components. SAM4MLLM \citep{chen2024sam4mllm} and Seg-Zero
\citep{liu2025seg} train MLLMs to prompt SAM.

Training-free alternatives include RSVP \citep{lu2025rsvp} and
attention-based grounding \citep{kang2025your}. Agentic systems include
\samagent \citep{carion2025sam} and the piipeline in Colon-Bench
\citep{HamdiAbd_ColonBench_MICCAI2026}. These approaches still delegate mask
production to a segmentation model. In contrast, \ours uses only text
queries to a frozen, general-purpose MLLM. It requires no exemplars, input
boxes, or video prompts. The MLLM emits the final geometry, which our loop
revises directly.

\mysection{Geometry as text and contour representations}
Representing geometry as text is not new. VisionLLM
\citep{wang2023visionllm} represents boundaries as short polygons in a
trained multitask pipeline.
VisionLLM~v2 \citep{wu2024visionllm} uses task-specific trained decoders.
PixelLLM \citep{xu2024pixel} learns localization heads and obtains masks
through SAM. Text4Seg \citep{lan2025text4seg} generates segmentation tokens
but requires fine-tuning. General models such as Molmo
\citep{deitke2025molmo} and Gemini \citep{comanici2025gemini,geminiseg2025}
also provide pointing and coarse localization. Classical boundary
representations include spline snakes \citep{brigger2000bspline}. Trained
vertex predictors include Polygon-RNN++ \citep{acuna2018efficient} and
Curve-GCN \citep{ling2019fast}. \ours uses this output space without
training a contour predictor. A frozen, general-purpose backend emits
control points that define the mask prediction. Only deterministic
geometric processing follows. Confined signed regions then allow the
refinement process to correct local parts of the mask.

%% file: sections/3.methodology.tex
\section{Method} \label{sec:method}
\vspace{-5pt}

\begin{figure}[t]
  \centering
  \includegraphics[width=\textwidth]{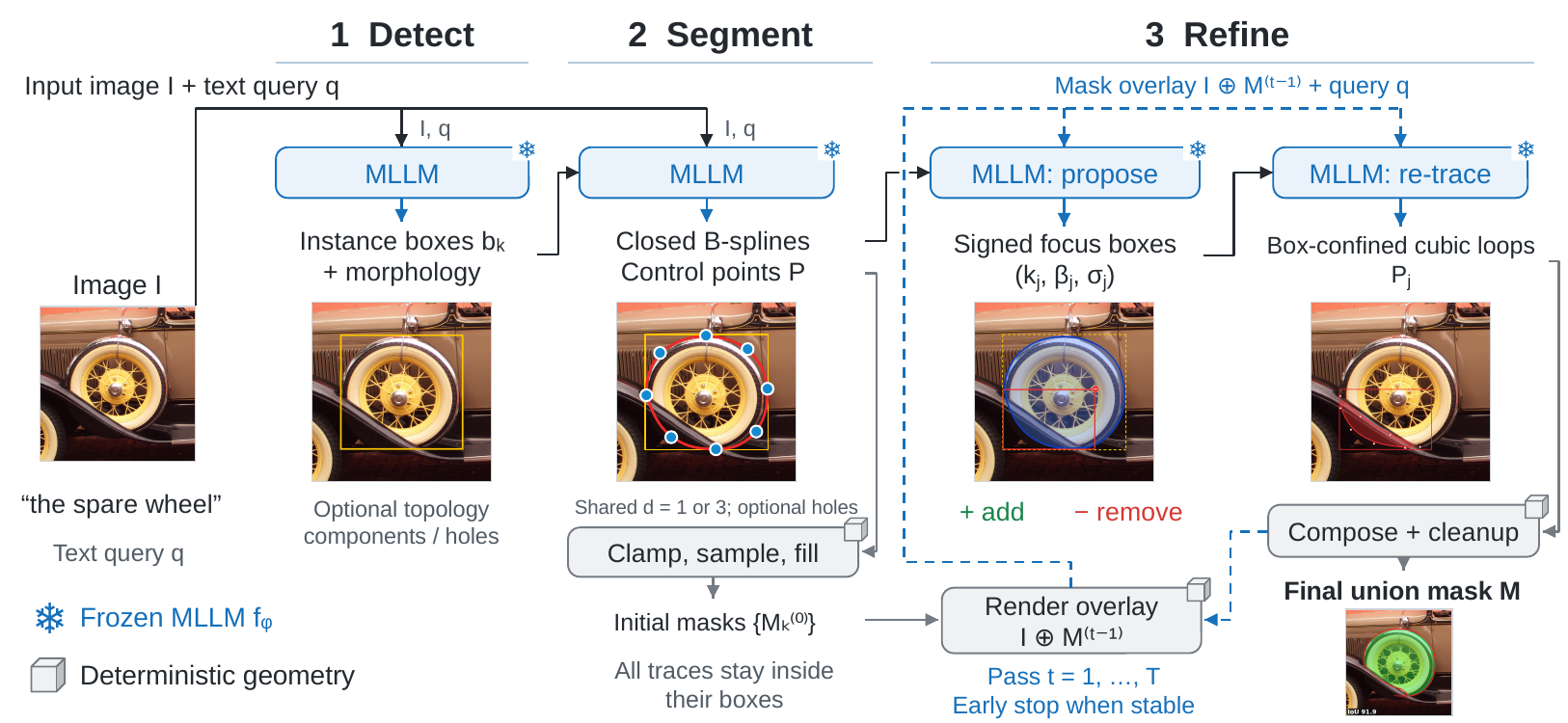}
  \vspace{-6pt}
  \caption{\small \textbf{\ours pipeline overview.}
  Detection localizes instances, segmentation traces box-confined B-splines,
  and refinement corrects the mask within signed focus boxes ($+$ adds pixels,
  $-$ removes them). Each MLLM call serves all instances (one is shown).}
  \label{fig:pipeline}
\end{figure}

\mysection{Overview}
Given an image $I$ and a free-form text query $q$, \ours predicts binary masks
for $N$ retained instances. We index these instances by $k=1,\ldots,N$.
Stage~2 initializes each mask as $M_k^{(0)}$. Refinement pass $t$ produces
$M_k^{(t)}$. Their union is $M^{(t)}=\bigcup_{k=1}^{N}M_k^{(t)}$.
The final output $M$ is the union at termination.
A frozen multimodal LLM $f_\phi$ is the only vision component. No parameters
are updated, and no segmentation head is attached. The pipeline accesses
$f_\phi$ only through fixed text prompts, with one prompt per stage call.
Each call is stateless and uses no tools. Deterministic code controls the
sequence. It also turns the returned geometry into masks.
\figLabel\ref{fig:pipeline} shows the three stages.

\subsection{Stage 1: detecting instances and their structure} \label{sec:stage1}
\vspace{-3pt}
A single detection call to $f_\phi$ takes $I$ and $q$ as input. For every
instance matching $q$, it returns an axis-aligned box
$b_k=[x_{k,1},y_{k,1},x_{k,2},y_{k,2}]$ and a confidence score. It also returns
a short \emph{morphology} description of the coarse shape and aspect.
Optionally, it provides a \emph{topology} descriptor that specifies the
component and hole counts. Near-duplicate boxes
are merged at IoU $\geq 0.9$. Instances are then ordered in reading order. At
most $N_{\max}$ are kept, so $0\leq N\leq N_{\max}$. Coordinates are given in
absolute pixels. The origin is at the top left. In set operations, $b_k$
denotes the pixel region inside that box. If $N=0$, the output is empty.

\begin{figure}[t]
  \centering
  \setlength{\tabcolsep}{1.5pt}
  \begin{tabular}{@{}cccc@{}}
    \includegraphics[height=2.15cm]{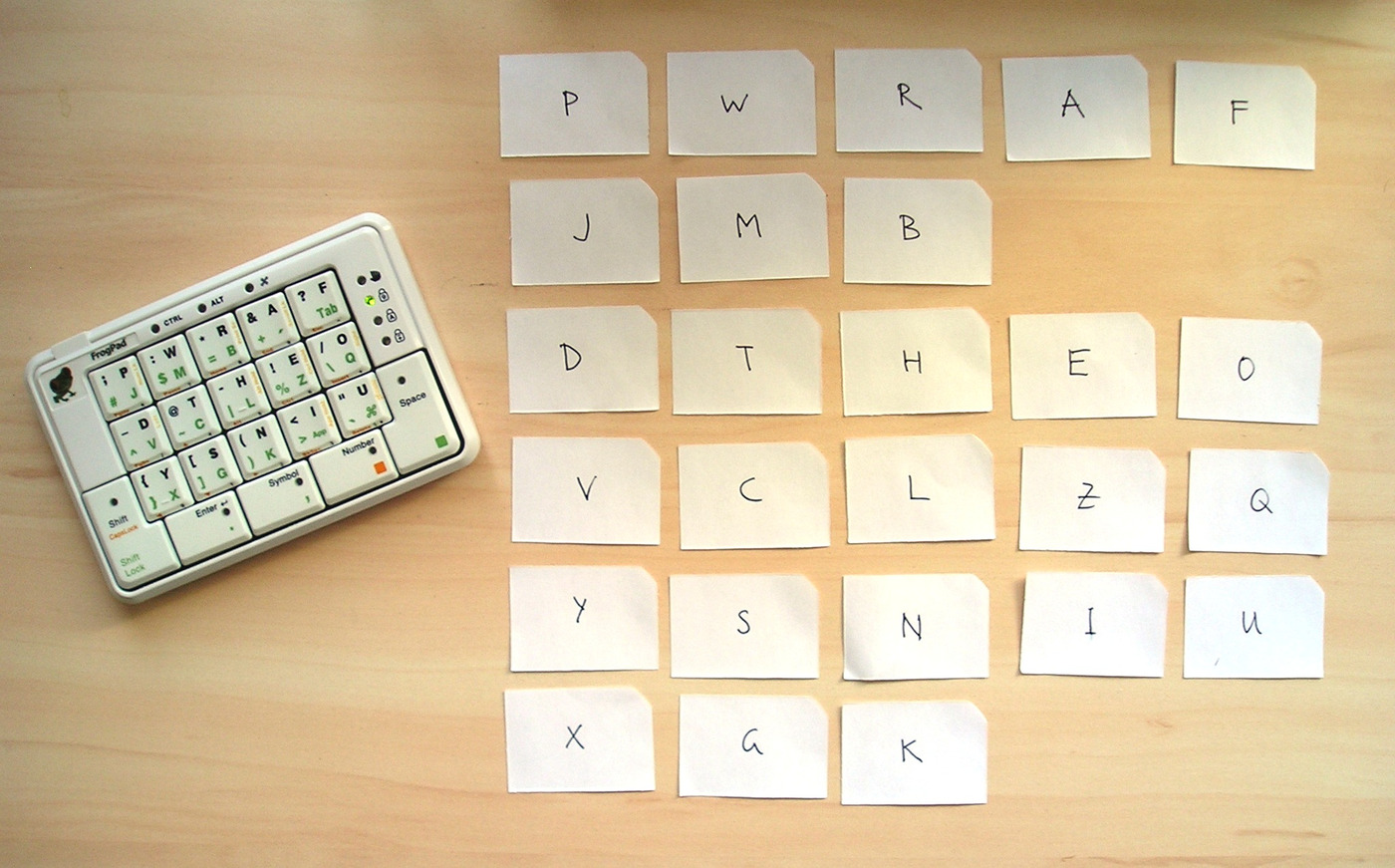} &
    \includegraphics[height=2.15cm]{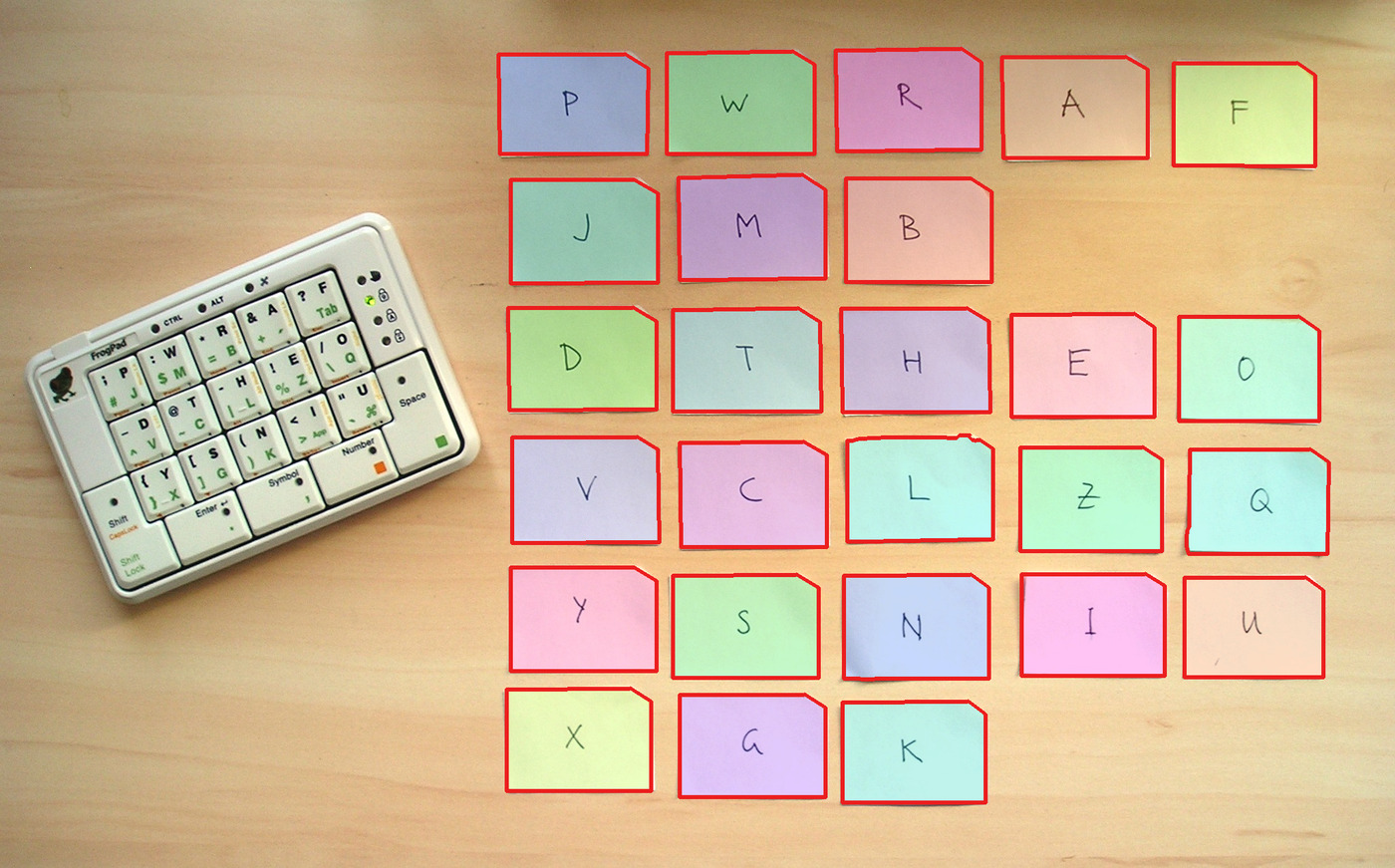} &
    \includegraphics[height=2.15cm]{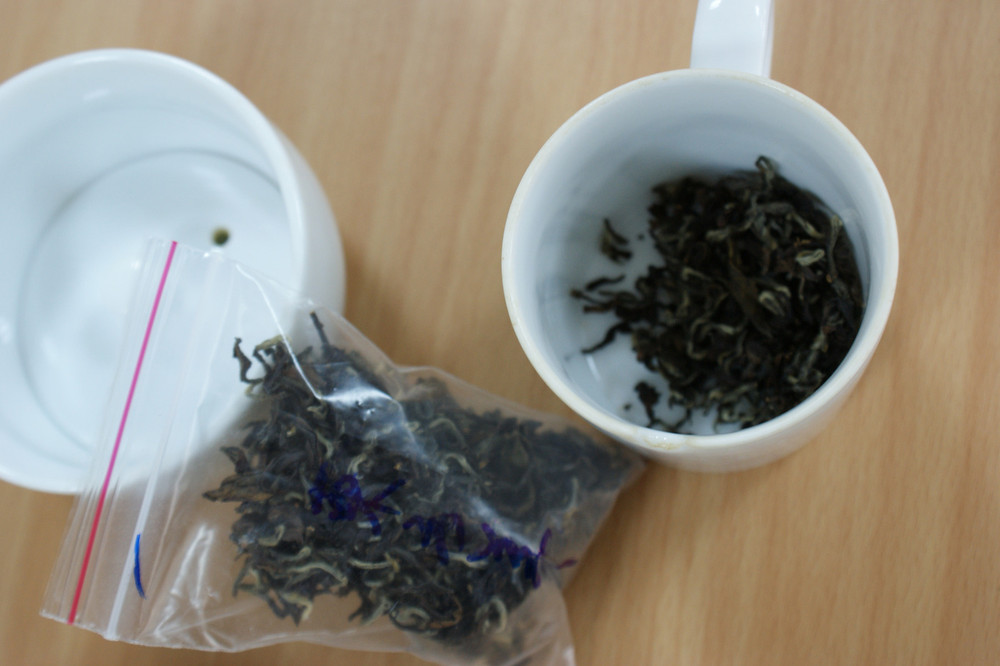} &
    \includegraphics[height=2.15cm]{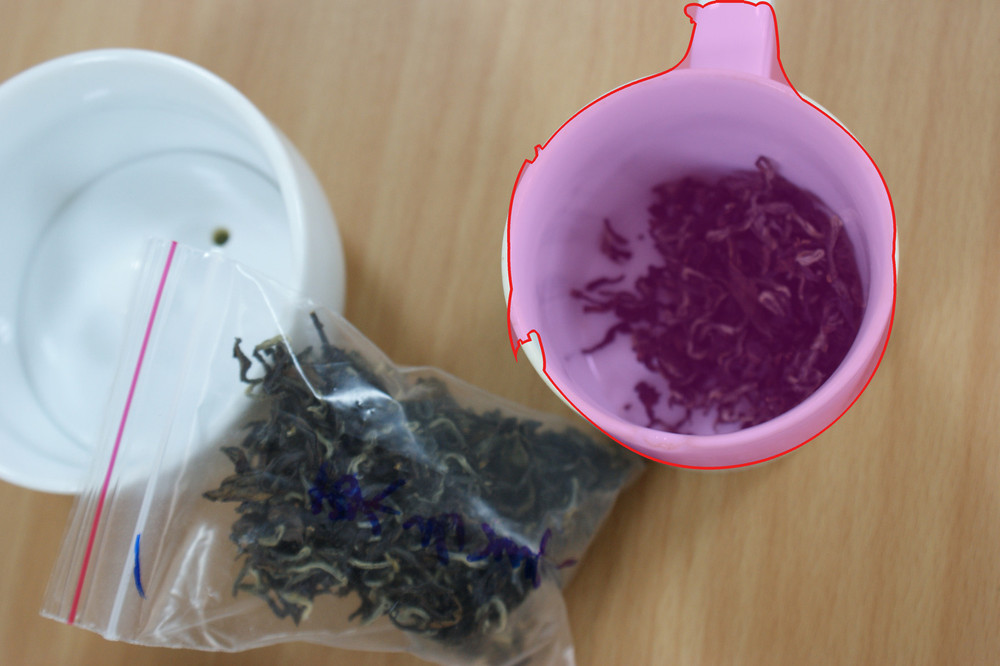} \\[2pt]
    {\small input} & {\small \ours} & {\small input} & {\small \ours} \\[1pt]
    \multicolumn{2}{@{}>{\centering\arraybackslash}p{0.50\linewidth}@{}}%
      {\scriptsize\textit{``In an educational setting, children often use different
       materials to learn about letters, numbers, and words. What object in
       the picture could be used as a visual aid for learning about letters
       and words?''}} &
    \multicolumn{2}{@{}>{\centering\arraybackslash}p{0.46\linewidth}@{}}%
      {\scriptsize\textit{``The general steps for brewing tea are to put tea leaves
       into a teacup and then pour hot water over them. What container in the
       picture is most likely to be used next for pouring hot water to make
       tea?''}} \\
  \end{tabular}
  \vspace{-6pt}
  \caption{\small \textbf{Reasoning queries work off the shelf.} Two ReasonSeg
  items segmented by \ours with \gfl. The same model reads the query, reasons about the image and
  outputs mask geometry, with no agent wrapper or query rewriting.}
  \label{fig:reasoning_qual}
\end{figure}

\begin{table}[t]
\centering
\resizebox{\linewidth}{!}{
\begin{tabular}{ll cc ccc ccc cc}
\toprule
& & \multicolumn{2}{c}{\textbf{ReasonSeg}} & \multicolumn{3}{c}{\textbf{RefCOCO}} & \multicolumn{3}{c}{\textbf{RefCOCO+}} & \multicolumn{2}{c}{\textbf{RefCOCOg}} \\
\cmidrule(lr){3-4}\cmidrule(lr){5-7}\cmidrule(lr){8-10}\cmidrule(lr){11-12}
\textbf{Method} & \textbf{MLLM/Backend} & val & test & val & testA & testB & val & testA & testB & val (U) & test (U) \\
\midrule
Seg-Zero                & Qwen2.5-VL 7B       & 62.0 & 52.0 & \na & \na & \na & \na & \na & \na & \na & \na \\
LISA                    & LLaVA 7B            & \na & \na & 74.9 & 79.1 & 72.3 & 65.1 & 70.8 & 58.1 & 67.9 & 70.6 \\
LISA                    & LLaVA1.5 13B        & 72.9 & 62.2 & \na & \na & \na & \na & \na & \na & \na & \na \\
GLaMM                   & Vicuna 7B           & 47.2 & \na & 79.5 & 83.2 & 76.9 & 72.6 & 78.7 & 64.6 & 74.2 & 74.9 \\
PSALM                   & Phi1.5 1.3B         & \na & \na & 83.6 & 84.7 & 81.6 & 72.9 & 75.5 & 70.1 & 73.8 & 74.4 \\
RICE                    & Qwen2.5-7B          & \na & \na & 83.5 & 85.3 & 81.7 & 79.4 & 82.8 & 75.4 & 79.8 & 80.4 \\
HyperSeg                & Phi2 2.7B           & 56.7 & \na & 84.8 & 85.7 & 83.4 & 79.0 & 83.5 & 75.2 & 79.4 & 78.9 \\
X-SAM                   & Phi3 3.8B           & 32.9 & 41.0 & 85.1 & 87.1 & 83.4 & 78.0 & 81.0 & 74.4 & 83.8 & 83.9 \\
\midrule
Grounded SAM            & \na                 & 14.5 & 16.4 & \na & \na & \na & \na & \na & \na & \na & \na \\
Gemini Seg              & Gemini2.5 Flash     & 13.3 & ~9.2 & \na & \na & \na & \na & \na & \na & \na & \na \\
SAM~3$^{*}$             & \na                 & ~9.9 & 11.3 & 42.0 & 42.8 & 44.2 & 33.6 & 37.0 & 32.9 & 40.0 & 39.7 \\
LISA                    & LLaVA1.5 13B        & 60.3 & 50.8 & \na & \na & \na & \na & \na & \na & \na & \na \\
Kang et al.             & LLaVA1.5 13B        & 60.5 & 49.9 & \textbf{76.1} & \textbf{78.9} & 72.8 & 64.1 & 67.1 & 57.3 & 67.7 & 69.0 \\
RSVP                    & GPT-4o              & 63.1 & 60.0 & \na & \na & \na & \na & \na & \na & \na & \na \\
\samagent               & Qwen2.5-VL 7B       & 49.1 & 53.5 & 59.4 & 64.3 & 55.0 & 51.4 & 57.0 & 44.9 & 57.2 & 58.8 \\
\samagent               & Qwen2.5-VL 72B      & 65.1 & 64.0 & 71.6 & 74.9 & 66.0 & 64.9 & 70.8 & 57.9 & 68.8 & 70.2 \\
\samagent               & Llama4 Maverick     & 61.5 & 60.9 & 71.7 & 76.5 & 66.5 & 65.1 & \underline{71.2} & 57.6 & 67.7 & 68.7 \\
\samagent               & Qwen3-VL 8B Think   & 57.4 & 67.3 & 68.6 & 72.3 & 63.9 & 59.3 & 64.5 & 55.6 & 66.4 & 66.9 \\
\samagent               & Qwen3-VL 235B Think & \underline{67.0} & \underline{71.8} & 73.9 & 77.0 & 69.7 & 66.9 & 70.9 & 62.2 & 72.1 & 72.9 \\
\samagent               & Gemini2.5 Pro       & 66.0 & \textbf{72.1} & \underline{75.5} & \underline{77.6} & 71.0 & \underline{67.3} & 71.1 & 63.4 & \underline{73.4} & 74.0 \\
SAM~3 Agent$^{*}$        & Gemma-4-31B         & 63.9 & 66.7 & 73.2 & 75.6 & 75.0 & 67.2 & \textbf{72.0} & 61.9 & 72.1 & \underline{74.7} \\
SAM~3 Agent$^{*}$        & \gfl                & 62.0 & 63.2 & 73.2 & 76.7 & \underline{75.5} & \textbf{68.4} & \textbf{72.0} & \textbf{68.5} & \textbf{76.5} & \textbf{76.9} \\
\rowcolor[HTML]{EFEFEF}
\oursnr                 & \gfl                & 64.6 & 58.2 & 65.5 & 65.7 & 72.9 & 59.7 & 61.4 & 61.1 & 64.8 & 69.3 \\
\rowcolor[HTML]{EFEFEF}
\oursfull               & \gfl                & \textbf{71.5} & 65.8 & 71.2 & 73.8 & \textbf{76.0} & 67.0 & 69.9 & \underline{64.9} & 70.9 & 74.5 \\
\bottomrule
\end{tabular}}
\caption{\small \textbf{Reasoning and referring expression segmentation.}
Results on ReasonSeg \citep{lai2024lisa} and RefCOCO/+/g
\citep{kazemzadeh2014referitgame,mao2016generation}. All values are cIoU. The
baselines are Seg-Zero \citep{liu2025seg}, LISA \citep{lai2024lisa}, GLaMM
\citep{rasheed2024glamm}, PSALM \citep{zhang2024psalm}, RICE
\citep{xie2025region}, HyperSeg \citep{wei2024hyperseg}, X-SAM
\citep{wang2025x}, Grounded SAM \citep{ren2401grounded}, Gemini Seg
\citep{geminiseg2025}, Kang et al.\ \citep{kang2025your}, RSVP
\citep{lu2025rsvp}, and both \samthree and \samagent \citep{carion2025sam}.
The \emph{top block} contains methods fine-tuned for referring or reasoning
segmentation. Unmarked rows are published
full-split numbers quoted from \samthree \citet{carion2025sam}. $^{*}$\,Our baseline
evaluation uses the complete ReasonSeg splits. On RefCOCO/+/g, it uses the
same 500-item random subset per split as \ours. 
\textbf{Bold} is the best and \underline{underline} the second best per column.}
\label{tab:reasonseg}
\end{table}

\subsection{Stage 2: from language to geometry} \label{sec:stage2}
\vspace{-3pt}
One batched call traces every instance at once. It returns an outer loop for
each instance. With topology conditioning, it also returns one
loop per predicted hole, each a list of control points.

\mysection{Representation} A loop is a \emph{closed periodic B-spline} of
degree $d$. It is defined over a control polygon
$P=(p_1,\dots,p_n)$ containing $n$ two-dimensional control points indexed by
$i$. The count $n$ can vary between loops. The polygon is wrapped so that
$p_{i+n}=p_i$:
\begin{equation}
  C_{P,d}(u) \;=\; \sum_{i=1}^{n} B_{i,d}(u)\; p_i ,
  \qquad u\in[0,n],\quad C_{P,d}(n)=C_{P,d}(0),
  \label{eq:bspline}
\end{equation}
where $u$ is the curve parameter and $B_{i,d}$ are the uniform periodic
B-spline basis functions of degree $d$ \citep{piegl1997nurbs}.
\eqn{eq:bspline} is a \emph{piecewise} polynomial consisting of $n$
degree-$d$ segments. Each $B_{i,d}$ is supported on only $d{+}1$ knot spans.
Thus, each control point influences only $d{+}1$ segments, and a misplaced
point changes a bounded arc rather than the whole curve. Stage~3 does not edit
these original control points. Its locality comes from adding or
subtracting focus-box regions.

We also let the model choose the degree. The same call that places the control
points first chooses $d \in \{1,3\}$ based on the object's form. It uses $d=3$
for rounded and organic silhouettes and $d=1$ for angular, straight-edged
ones. It then places the points as the selected degree requires. Cubic control
points are pushed outward toward the edge, whereas degree-1 vertices lie
directly on it. One degree is shared by all loops of all instances in the
stage-2 call.

\mysection{Rasterization by sampling} Masks are produced by sampling the curve
and filling the resulting polygon, not by evaluating an implicit function.
Let $\mathrm{clip}_b$ denote the per-coordinate clamp onto a box $b$. Let
$\mathrm{Fill}$ denote even-odd polygon fill \citep{foley1996computer}. One
loop then becomes the region
\begin{equation}
  \mathcal{R}_{b,d}(P) \;=\;
  \mathrm{Fill}\Big(
    \big(\mathrm{clip}_b(C_{P,d}(u_r))\big)_{r=1}^{m}
  \Big),
  \qquad u_r = \tfrac{r-1}{m-1}\,n ,
  \label{eq:raster}
\end{equation}
with $m$ samples indexed by $r$, uniformly spaced in the curve parameter
and including the repeated endpoint. For cubic loops, $m$ is proportional to
the perimeter of $b$, so small and large objects are rasterized at comparable
boundary resolution. For $d=1$, the control polygon is filled directly.
\eqn{eq:raster} applies $\mathrm{Fill}$ to an ordered sequence of boundary
samples. The control points are used as returned, and the sampled boundary is
clamped to $b$ before filling.
A trace can never leave its own detected box, so a hallucinated control point
cannot paint across the image. Overlapping instance boxes can still produce
overlapping masks. Consequently, detection quality upper-bounds segmentation
quality.
With topology conditioning off, the initial instance mask is
$M_k^{(0)}=\mathcal{R}_{b_k,d}(P_k^{\mathrm{out}})$, where
$P_k^{\mathrm{out}}$ is instance $k$'s outer control polygon.
When topology conditioning is enabled, regions rasterized from predicted hole
loops are subtracted from the outer region to obtain $M_k^{(0)}$.
The predicted hole count is not guaranteed to survive loop overlap,
rasterization, refinement, or cleanup.

\begin{figure}[t]
  \centering
  \setlength{\qgw}{0.158\linewidth}
  \setlength{\qhdh}{22pt}
  \setlength{\tabcolsep}{0pt}
  \renewcommand{\arraystretch}{0.9}
  \makebox[\linewidth][s]{%
  \begin{tabular}[b]{@{}c@{}}\\ \qrow{Input+\gtamber{GT}}\\ \qrow{Baseline}\\ \qrow{\ours}\end{tabular}%
  \hfill\qcol{Reasoning}{ReasonSeg}{\ldots likely to be the girl's trainer}{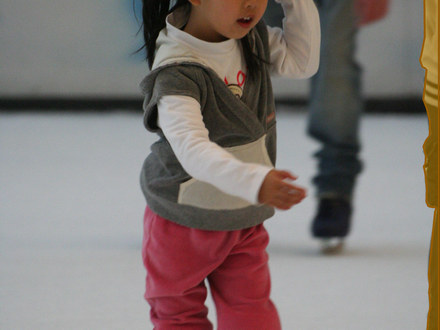}{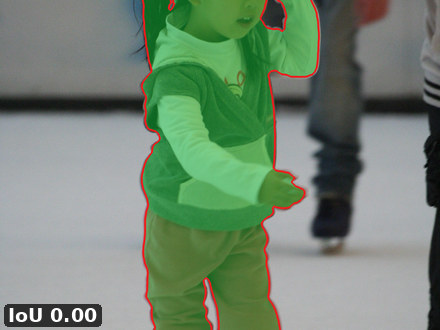}{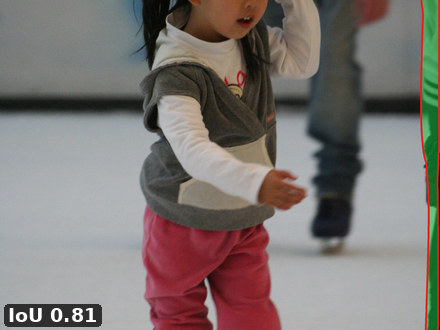}%
  \hfill\qcol{Many instances}{LVIS}{orange (fruit)}{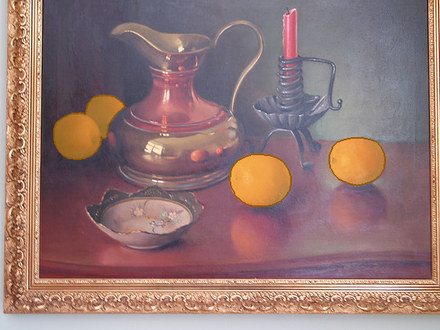}{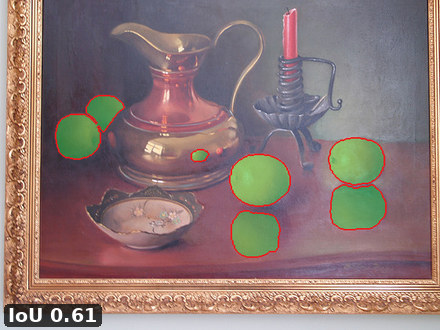}{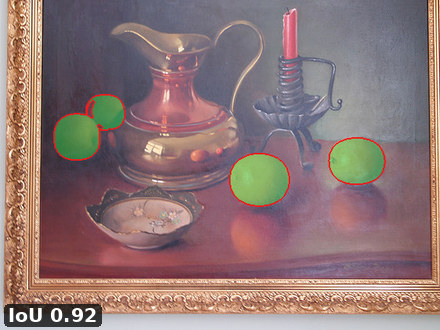}%
  \hfill\qcol{Hard negative}{SA-Co/Gold}{built-in bidet}{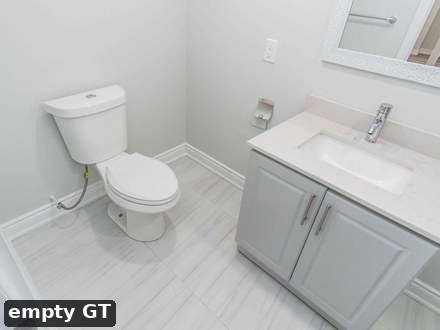}{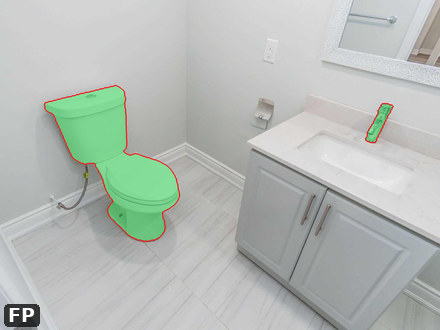}{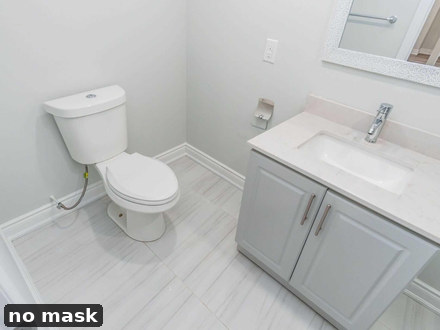}%
  \hfill\qcol{Rare domain}{RF-100VL X-ray}{Pneumonia}{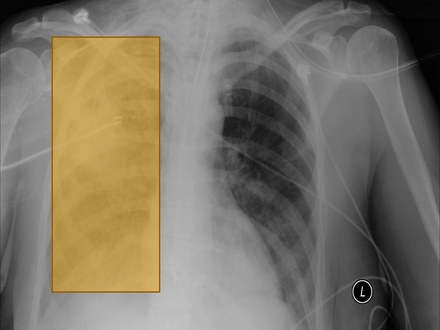}{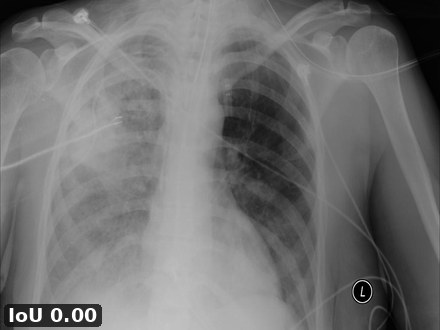}{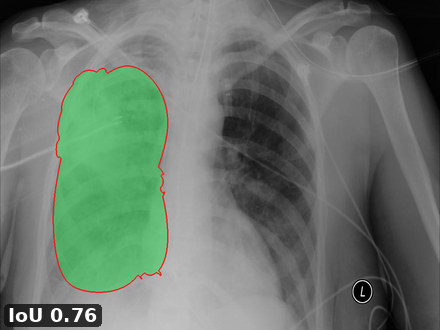}%
  \hfill\qcol{Referring}{RefCOCOg}{a pink and white striped sofa\ldots}{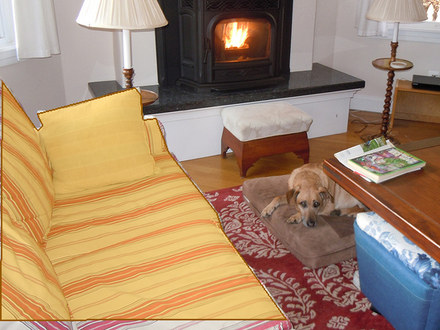}{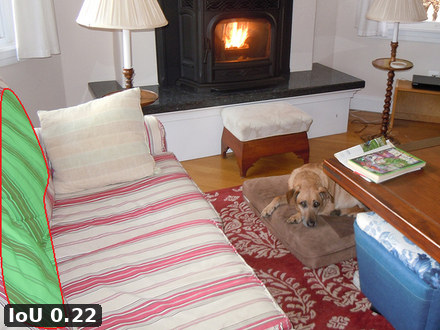}{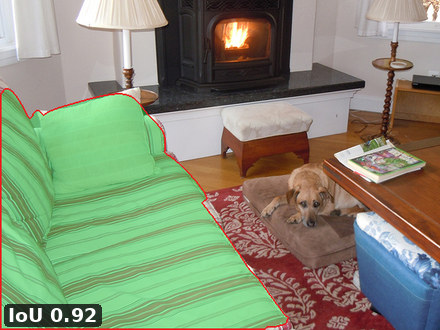}%
  \hfill\qcol{Failed first trace}{in the wild}{the dog}{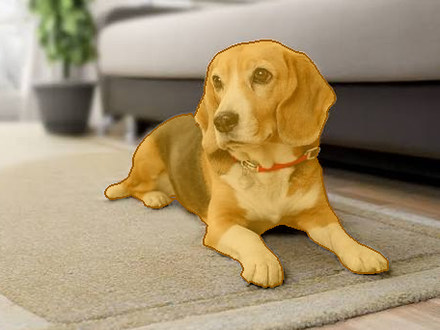}{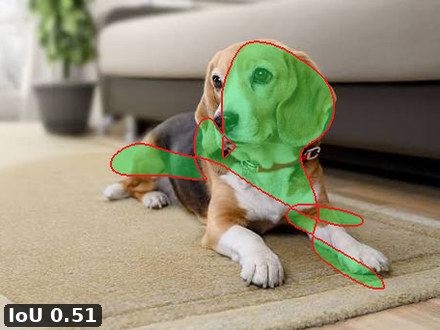}{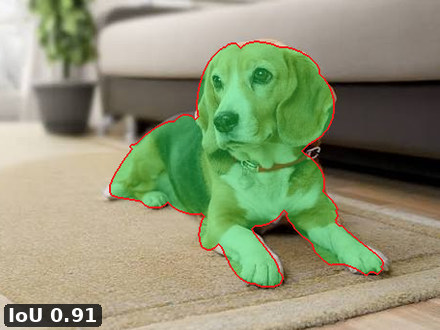}%
  }
  \vspace{-14pt}
  \caption{\small \textbf{\ours Qualitative comparisons.} We show compariosns of \ours with baselines on various setups. \emph{Top}: input with ground truth
  in \gtamber{amber}. \emph{Middle}: the baselines \samthree in the first four columns and \oursnr in the last two. \emph{Bottom}: \ours.}
  \label{fig:qual_main}
\end{figure}

\begin{table}[t]
\centering
\setlength{\tabcolsep}{3pt}
\fontsize{8.5}{10}\selectfont
\begin{tabular}{l cc c cccc cccc}
\toprule
& \multicolumn{3}{c}{\textbf{Semantic segmentation}}
& \multicolumn{8}{c}{\textbf{Zero-shot detection}} \\
\cmidrule(lr){2-4}\cmidrule(lr){5-12}
& \multicolumn{2}{c}{\textbf{ADE20K-150}} & \textbf{Cityscapes}
& \multicolumn{4}{c}{\textbf{ODinW13}} & \multicolumn{4}{c}{\textbf{RF-100VL}} \\
\cmidrule(lr){2-3}\cmidrule(lr){4-4}\cmidrule(lr){5-8}\cmidrule(lr){9-12}
\textbf{Method} & gIoU & cIoU & mIoU & AP$_0$ & gIoU & cIoU & \pmf & AP$_0$ & gIoU & cIoU & \pmf \\
\midrule
APE-D (LVIS-trained) & \na & \na & 44.2 & \na & \na & \na & \na & \na & \na & \na & \na \\
Gemini 2.5 Pro & \na & \na & \na & 33.7 & \na & \na & \na & 11.6 & \na & \na & \na \\
gDino-T & \na & \na & \na & 49.7 & \na & \na & \na & 15.7 & \na & \na & \na \\
gDino1.5-Pro & \na & \na & \na & \underline{58.7} & \na & \na & \na & \na & \na & \na & \na \\
\samthree & \na & \na & \textbf{65.2} & \textbf{61.0} & \na & \na & \na & 15.2 & \na & \na & \na \\
SAM~3$^{*}$ & \underline{57.9} & \textbf{67.3} & \underline{63.7} & 51.5 & 48.3 & \underline{51.5} & 39.6 & 14.2 & 13.8 & 21.7 & 11.4 \\
\rowcolor[HTML]{EFEFEF}
\oursnr & 56.6 & 60.5 & 48.8 & 52.0 & \underline{48.7} & 48.7 & \underline{50.0} & \underline{30.4} & \underline{42.1} & \underline{50.8} & \underline{58.7} \\
\rowcolor[HTML]{EFEFEF}
\oursfull & \textbf{59.2} & \underline{64.7} & 53.3 & 57.2 & \textbf{54.2} & \textbf{55.7} & \textbf{57.5} & \textbf{31.4} & \textbf{43.6} & \textbf{53.2} & \textbf{61.7} \\
\bottomrule
\end{tabular}
\caption{\small \textbf{Semantic segmentation and zero-shot detection.}
The baselines are APE-D \citep{Shen_2024_CVPR}, Gemini 2.5 Pro
\citep{comanici2025gemini}, and gDino-T \citep{groundingdino}.
\ours uses \gfl. Unmarked baselines are published full-split numbers.
$^{*}$\,Our evaluation uses complete splits, except on ADE20K, where we use
the same 1{,}000 random items as the shaded rows. ADE20K reports oracle-class
union-IoU and Cityscapes reports 19-class mIoU. Detection reports mean
per-subset box AP$_0$ and pooled mask metrics. \textbf{Bold} is the best and
\underline{underline} the second best per column.}
\label{tab:semantic}
\label{tab:apzero}
\end{table}

\subsection{Stage 3: iterative refinement} \label{sec:stage3}
\vspace{-3pt}
The initial outline often identifies the right object but misses boundary
detail. Each refinement pass asks the model to locate errors on the current
mask overlay. It then retraces those regions within smaller boxes. In pass
$t=1,\ldots,T$, both calls receive $I\oplus M^{(t-1)}$, the previous union
mask rendered over the image. Leaving the fixed prompts and structural context
implicit, we write the loop as
\begin{equation}
\begin{aligned}
  \{(k_j,\beta_j,\sigma_j)\}_j
    &= f_\phi(I\oplus M^{(t-1)},q), \\
  M_k^{(t)}
    &= \mathrm{Compose}\!\left(M_k^{(t-1)},
       \{(\sigma_j,\mathcal R_{\beta_j,3}(P_j))\}_{j:\,k_j=k}\right).
\end{aligned}
\label{eq:refine}
\end{equation}
Here $j$ indexes the focus boxes in the current pass. The value $k_j$ identifies
the instance corrected by the $j$th focus box. The box $\beta_j$ is confined
to that instance's box. The sign $\sigma_j\in\{+1,-1\}$ requests addition or
removal. A second call uses the same overlay, query, and box geometry. It
returns cubic control polygons for all retained boxes in one batch, where
$P_j$ is the control polygon of the $j$th focus box, which corrects instance
$k=k_j$.
Before this call, an effect gate discards boxes with negligible potential to
change the mask. If none remain, retracing is skipped.

Composition applies each correction to instance $k_j$ and cleans up small
artifacts. It carries a per-instance commitment field that is implicit in
\eqn{eq:refine}. This field resists opposing edits and reduces oscillation.
Boolean composition and cleanup need not leave a globally smooth spline boundary.
The updated union $M^{(t)}$ feeds the next pass. Both model calls batch all
instances, so each pass uses at most two MLLM calls regardless of $N$.
The gating and cleanup settings, the per-pixel composition rule, and a
worked example of each call are given in \secLabel\ref{app:pipeline}.
Refinement ends after at most $T$ passes, or once the
union mask changes only negligibly for $s$ consecutive passes (the patience
criterion).
\secLabel\ref{sec:experiments} gives additional details, and \secLabel\ref{sec:analysis} examines convergence.

%% file: sections/4.experiments.tex
\section{Experiments} \label{sec:experiments}
\vspace{-5pt}
\subsection{Experimental setup} \label{sec:setup}
\vspace{-3pt}
\mysection{Benchmarks and metrics}
We follow the image evaluation suite of \samthree \citep{carion2025sam}. We
use each benchmark's standard metrics.
\textbf{(i)}~\emph{Referring and reasoning segmentation}: We evaluate on
ReasonSeg \citep{lai2024lisa} (val/test) and all splits of RefCOCO, RefCOCO+,
and RefCOCOg \citep{kazemzadeh2014referitgame,mao2016generation}. We report
gIoU (mean per-item IoU) and cIoU (cumulative
$\nicefrac{\sum\text{I}}{\sum\text{U}}$).
\textbf{(ii)}~\emph{Open-vocabulary instance segmentation}: We evaluate on
LVIS \citep{lvis} and COCO \citep{coco}, and additionally on SA-Co/Gold
\citep{carion2025sam}, which is reported in the appendix
(\tblLabel\ref{tab:openvocab_full}). We report presence-gated \cgf and \pmf (with hard
negative prompts), positive-set IoU, and COCO-style box/mask AP.
\textbf{(iii)}~\emph{Semantic segmentation}: We report 19-class mIoU on
Cityscapes \citep{cordts2016cityscapes}, using deterministic mask compositing.
We use ADE20K \citep{ade20k} as a class-prompted oracle upper bound.
\textbf{(iv)}~\emph{Zero-shot detection}: We evaluate on ODinW13
\citep{odinw} and RF-100VL \citep{robicheaux2025roboflow100vl}. We report
AP$_0$, the mean per-subset box AP computed from masks.

\mysection{Protocol: complete splits and sampled subsets}
We run each split \emph{in full} whenever end-to-end evaluation is practical.
This includes ReasonSeg val (200) and test (779), Cityscapes val (500), and
every ODinW13 and RF-100VL subset ($6{,}237$ and $1{,}987$ items). In total,
we fully evaluate $9{,}703$ of the $17{,}703$ scored items. Four of the ten
benchmarks are run in full.
The remaining annotation files contain $10^4$ to $10^5$ items, making full
MLLM inference impractical. We instead draw uniform random samples with
seed~0. We do not take the first items because the annotation files are
strongly ordered. We sample $1{,}500$ items each from LVIS and COCO,
$1{,}000$ from ADE20K, and $500$ from each RefCOCO/+/g split. The item
counts above cover the ten benchmarks of the main tables. The appendix
SA-Co/Gold evaluation uses a further $2{,}500$ items covering all seven domains.
On these sampled splits, we run every baseline once on the \emph{full} split.
We then recompute its metrics on exactly the item set seen by \ours, using a
shared-key intersection. Negative sampling is also identical: 3 negatives per
positive for LVIS and COCO, with seed~0.

\mysection{\ours configuration}
The main results use \gfl as the backend for all
three stages. The backend selects the Stage-2 curve degree
$d\in\{1,3\}$ (\secLabel\ref{sec:method}). This degree is shared across all
instances in a call. Refinement curves are always cubic.
Refinement uses a budget of $T{=}8$ passes, a cap of $N_{\max}{=}200$
instances per image (matching \samthree's cap), and a focus-box budget of 6
per instance per pass.
The stopping criterion is
$|M^{(t)} \triangle M^{(t-1)}| < \varepsilon\,|\bigcup_{k=1}^{N} b_k|$,
where $\varepsilon{=}0.005$. The loop stops when this criterion holds for
$s{=}3$ consecutive passes with $t\geq1$. Here, $s$ is the stopping patience,
$|\cdot|$ counts pixels, and $\triangle$ denotes symmetric difference. The
denominator is the union of the instance boxes. By \eqn{eq:raster}, this is
the only region that a pass can alter. Normalizing by the full image would
make the same threshold behave very differently for small and large objects
(\secLabel\ref{sec:analysis}).

A full run uses 2 MLLM calls for detection and segmentation. Each refinement
pass adds at most 2 calls. Thus, the maximum is $2T{+}2=18$ calls, although
early stopping keeps the measured mean well below this limit. The stage-1
topology descriptor (\secLabel\ref{sec:method}) is optional. The main results
disable it for all three stages (\secLabel\ref{sec:analysis}). We use the same
settings on every benchmark and tune nothing per dataset.

\setcounter{topnumber}{2}\setcounter{totalnumber}{3}\setcounter{bottomnumber}{1}
\mysection{Baselines}
Our two main baselines are from the \samthree paper \citep{carion2025sam}.
\textit{\samthree}: We use the public checkpoint through Hugging Face
Transformers in fp32. We provide one text query per item and allow up to 200
instances. We score its output with the same metric code on the same items.
\samthree also accepts image-exemplar and box prompts and supports video. We
use none of these features, and \ours is text-only by design.
\textit{\samagent}: This agent baseline was introduced in the same paper. An
MLLM drives \samthree as a tool. The MLLM
proposes simple noun phrases for \samthree, inspects the returned numbered
masks, and selects those matching the query. We use Gemma-4-31B-it and \gfl
rather than the paper's Qwen and Gemini~2.5~Pro backends. The selected union
is scored identically. We also quote full-split numbers of \textit{published specialists} from
\citet{carion2025sam}.

\begin{table}[t]
\centering
\small
\setlength{\tabcolsep}{6pt}
\begin{tabular}{@{}l ccc ccc@{}}
\toprule
& \multicolumn{3}{c}{\textbf{LVIS}} & \multicolumn{3}{c}{\textbf{COCO}} \\
\cmidrule(lr){2-4}\cmidrule(lr){5-7}
\textbf{Method} & \cgf & \shortstack{box\\AP} & \shortstack{mask\\AP} & \cgf & \shortstack{box\\AP} & \shortstack{mask\\AP} \\
\midrule
OWLv2 & 20.1 & 35.2 & \na & \na & 38.2 & \na \\
OWLv2$^{\dagger}$ & 29.3 & 45.5 & 43.4 & \na & 46.1 & \na \\
gDino-T & 14.7 & 20.5 & \na & \na & 45.7 & \na \\
LLMDet-L & 35.1 & 42.0 & 36.3 & \na & 55.6 & \na \\
APE-D$^{\dagger}$ & \na & 59.6 & \underline{53.0} & \na & 58.3 & \na \\
DINO-X & \na & 52.4 & 38.5 & \na & 56.0 & \na \\
Gemini 2.5 Flash & 13.4 & \na & \na & \na & \na & \na \\
\samthree & 37.2 & 53.6 & 48.5 & \na & 56.4 & \na \\
SAM~3$^{*}$ & 50.1 & \textbf{69.1} & \textbf{62.2} & \underline{52.4} & \textbf{68.0} & \textbf{59.2} \\
SAM~3 Agent$^{*}$ & \underline{51.0} & 56.2 & 50.9 & 49.9 & 48.9 & 42.0 \\
\rowcolor[HTML]{EFEFEF}
\oursnr & 50.4 & 57.1 & 36.9 & 50.4 & 59.6 & 38.0 \\
\rowcolor[HTML]{EFEFEF}
\oursfull & \textbf{53.9} & \underline{64.2} & 47.2 & \textbf{56.1} & \underline{66.6} & \underline{50.5} \\
\bottomrule
\end{tabular}
\caption{\small \textbf{Open-vocabulary instance segmentation.}
The baselines are OWLv2 \citep{minderer2024scalingopenvocabularyobjectdetection},
gDino-T \citep{groundingdino}, LLMDet-L \citep{fu2025llmdet},
APE-D \citep{Shen_2024_CVPR}, DINO-X \citep{ren2025dinoxunifiedvisionmodel},
and Gemini 2.5 Flash \citep{comanici2025gemini}.
$^{*}$\,Evaluated on a matched subset. Unmarked rows are
published results from \samthree \citep{carion2025sam}.
$^{\dagger}$\,LVIS-trained. \cgf is presence-gated F1.}
\label{tab:openvocab}
\end{table}

\begin{figure}[t]
  \centering
  \includegraphics[width=\textwidth]{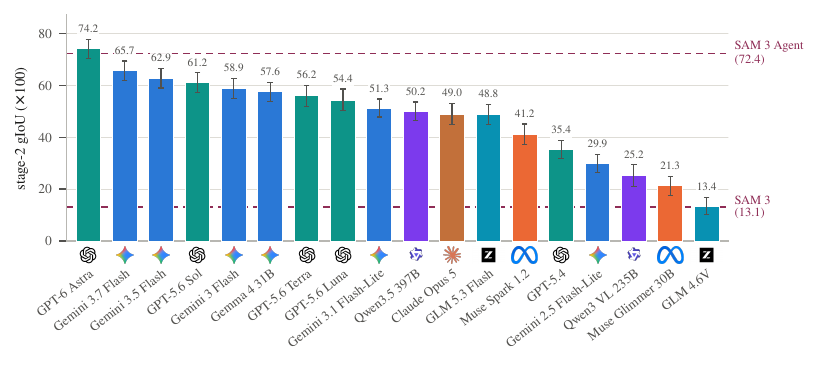}
  \vspace{-6pt}
  \caption{\small \textbf{\ours quality is the backend's quality.} Stage-2 gIoU
  on all 200 ReasonSeg val items. We show 18 of the 21 tested backends for
  readability. Whiskers show 95\% item-bootstrap intervals. Dashed lines mark
  \samthree and \samagent (\gfl). Recent MLLMs like GPT-6~Astra surpasses \samagent even without refinement.}
  \label{fig:backends}
\end{figure}

\subsection{Main results} \label{sec:results}
\vspace{-3pt}
Tables~\ref{tab:reasonseg} to~\ref{tab:openvocab} summarize ten benchmarks.
\figLabel\ref{fig:qual_main} shows one example per setting. Additional examples
appear in Figures~\ref{fig:qual_compare} to~\ref{fig:supp_passes}.

\mysection{Reasoning segmentation}
On ReasonSeg (\tblLabel\ref{tab:reasonseg}), \ours reaches 71.5 cIoU on
validation. This is the best result among methods not fine-tuned for the task.
Refinement adds 6.9 cIoU over the segment stage alone. The gIoU scores and
complete published results for both tasks appear in
\secLabel\ref{app:full_comparisons}.

\mysection{Referring expression segmentation}
On RefCOCO/+/g (\tblLabel\ref{tab:reasonseg}), \ours substantially outperforms
\samthree, while \samagent generally remains stronger. Refinement improves
cIoU across all eight splits. On RefCOCO testB, \ours reaches 76.0 cIoU, the
best zero-shot result on that split. This result shows that refinement narrows
the gap to \samagent.

\mysection{Open-vocabulary instance segmentation}
\ours reaches 53.9 and 56.1 \cgf on LVIS and COCO
(\tblLabel\ref{tab:openvocab}). It outperforms both \samthree baselines in
\cgf but trails \samthree in mask AP.

\mysection{Semantic segmentation and zero-shot in-the-wild detection}
\ours achieves 59.2 gIoU on ADE20K and 53.3 mIoU on Cityscapes
(\tblLabel\ref{tab:semantic}). On Cityscapes, it trails \samthree by 10.4
points. Dense scene labeling therefore remains a limitation. On zero-shot
detection, \ours reaches 57.2 AP$_0$ on ODinW13 and 31.4 on RF-100VL
(\tblLabel\ref{tab:apzero}). It exceeds \samthree on all four metrics in both
benchmarks. Its AP$_0$ on RF-100VL is $2.2\times$ that of \samthree, showing
strong cross-domain transfer.

%% file: sections/6.analysis.tex
\section{Analysis} \label{sec:analysis}
\vspace{-5pt}

\subsection{Backend progress transfers without segmentation training}
\vspace{-3pt}
We hold the pipeline fixed ($T{=}12$, topology on) and test 21 backends from
six developers. Stage-2 gIoU ranges from $13.4$ to $74.2$ on ReasonSeg val
(\figLabel\ref{fig:backends}). Recently-released GPT-6~Astra reaches $74.2$ without a
segmentation model, compared with $72.4$ for \samagent driven by \gfl.
The benefit is that \ours can use better backends without retraining. 
General benchmark capability is a weak
predictor of segmentation quality. Strong public scores do not reliably
translate into strong masks (\eg Claude Opus 5). The full ranking and generation
comparisons are in \secLabel\ref{app:backend_study}. The appendix also reports
the stage-2 and final results and the study protocol.

\makeatletter\global\@topnum=0\makeatother
\begin{figure}[t]
\centering
\begin{minipage}[b]{0.45\textwidth}
  \centering
  \includegraphics[width=\linewidth]{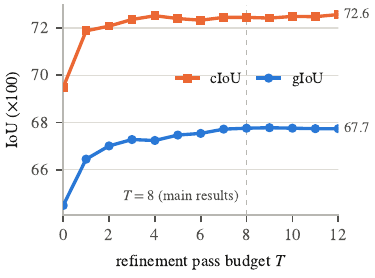}
  \vspace{-14pt}
  \caption{\small \textbf{Refinement budget.} We score the mask after the first
  $T$ passes at every budget on 200 ReasonSeg val items with \gfl. Gains
  saturate before $T{=}8$, the budget of the main results.}
  \label{fig:budget}
\end{minipage}\hfill
\begin{minipage}[b]{0.52\textwidth}
\centering
\small
\setlength{\tabcolsep}{4pt}
\resizebox{\linewidth}{!}{%
\begin{tabular}{@{}l cc cc@{}}
\toprule
 & \multicolumn{2}{c}{\textbf{ReasonSeg val}} & \multicolumn{2}{c}{\textbf{RefCOCO val}} \\
\cmidrule(lr){2-3} \cmidrule(lr){4-5}
\textbf{Configuration} & cIoU $\uparrow$ & calls $\downarrow$ & cIoU $\uparrow$ & calls $\downarrow$ \\
\midrule
w/o refinement & 67.2\,\sd{2.72} & 2.0 & 64.6 & 2.0 \\
w/o detection & 72.1\,\sd{0.87} & 8.8 & 78.8 & 10.4 \\
w/ topology & 72.6\,\sd{0.23} & 14.3 & 77.3 & 16.8 \\
Polygon loops & 73.1\,\sd{3.16} & 14.7 & 79.1 & 16.6 \\
Per-instance refinement & \textbf{74.9}\,\sd{0.91} & 21.3 & 77.7 & 18.8 \\
\midrule
\rowcolor[HTML]{EFEFEF}
Main conditioning & 74.7\,\sd{1.62} & 14.6 & \textbf{79.8} & 17.5 \\
\bottomrule
\end{tabular}}
\vspace{2pt}
\captionof{table}{\small \textbf{Design alternatives.} Final cIoU and mean MLLM calls per
image. ReasonSeg values are mean $\pm$ run SD over three runs. RefCOCO values
come from one run. Gray marks the main conditioning.}
\label{tab:ablations_main}
\end{minipage}
\end{figure}

\subsection{Refinement improves the boundary and usually stops early}
\vspace{-3pt}
With \gfl, refinement improves every benchmark in
Tables~\ref{tab:reasonseg} to~\ref{tab:openvocab}. Focus boxes let the model
correct a local boundary without redrawing the object. In the topology-enabled
ablation, removing refinement reduces cIoU by $5.4$ on ReasonSeg and $12.7$
on RefCOCO (\tblLabel\ref{tab:ablations_main}). This is the clearest component
effect. However, refinement is also the largest source of inference cost.
The loop has no convergence guarantee, so we measure both mask stability and
quality. In the $T{=}12$ ReasonSeg study, $79\%$ of items stop before the cap
(\figLabel\ref{fig:residual}a). The median item uses six passes, while $18.5\%$
reach the cap.
We score the mask after each pass budget (\figLabel\ref{fig:budget}). Four
passes recover $2.7$ of the eventual $3.2$ gIoU gain. Raising $T$ from $8$ to $12$ changes gIoU by under $0.1$. The main results
therefore use $T{=}8$. At $T{=}12$, early stopping cuts mean calls from $26$ to $14.7$.

\subsection{Ablation study}
\vspace{-3pt}
\tblLabel\ref{tab:ablations_main} compares the core design alternatives in
final-mask quality and inference calls with \gfl on 200 items per split. The shaded row uses the
main method's conditioning without topology. The  ReasonSeg reports three-run means and
standard deviations. RefCOCO has one run per configuration.

\mysection{Detection helps the initial trace} Removing detection lowers initial
cIoU from $68.3$ to $60.3$ on ReasonSeg. On RefCOCO, it lowers initial cIoU
from $66.0$ to $63.5$. After refinement, the no-detection configuration closes
these gaps with fewer total calls. Detection's initial advantage is larger on
the 63 ReasonSeg targets with disconnected ground-truth masks ($+11.3$ cIoU).
After refinement, 

\mysection{Batching bounds the cost as instance count grows} Per-instance cropped refinement
increases ReasonSeg calls from $14.3$ to $21.3$ (49\%). This yields a
$2.3$-point cIoU increase. On RefCOCO, it adds calls for only a $0.4$-point
increase. Batched tracing uses one call, regardless of the instance count, and hence it is our default choice. In contrast, crop-based
calls grow with the number of instances.

\mysection{Letting the model choose the degree helps} Forcing polygon loops
(B-splines of degree $1$) degrades performance on some instances. Letting the
model pick the degree improves it, specially on curved structures that are difficult to represent with straght lines ( see \figLabel\ref{fig:qual_polygon_supp} in the Appendix).

%% file: sections/7.conclusion.tex
\section{Conclusion} \label{sec:conclusion}
\vspace{-5pt}
We presented \ours, a training-free pipeline that transforms a multimodal LLM
into a text-prompted segmenter using box-confined B-spline geometry and
iterative focus-box refinement. Across ten benchmarks spanning four task
families, \ours is competitive with \samthree and performs better on reasoning
queries, rare domains, and some metrics. Its limitations include
detection errors, imprecise pixel-level boundaries, amorphous classes, and
inference cost. \ifdefined\arxivversion
The code, prompts, and evaluation harness are available at \href{\codeurl}{\codeurl}.
\else
We will release the code, prompts, and evaluation harness.
\fi

%% file: sections/supp.tex
\section*{Appendix}
\vspace{-4pt}

The appendix follows the order of the paper. \secLabel\ref{app:extended_related}
extends the related work. \secLabel\ref{app:pipeline} gives the method details
that the main text states only in words, and \secLabel\ref{app:eval} the
experimental setup: benchmarks, configuration, scoring, and the protocols for
the backend and cost studies. \secLabel\ref{app:results} holds the complete
result tables and qualitative grids. \secLabel\ref{app:analysis_extended}
covers the extended analysis (refinement, backends, ablations, topology,
localization, cost, and limitations), while
\secLabel\ref{app:prompts} reproduces the prompt templates verbatim.

\section{Extended Related Work} \label{app:extended_related}
\vspace{-4pt}

\mysection{Promptable and open-vocabulary segmentation}
SAM \citep{kirillov2023segment} introduced promptable segmentation with
points, boxes, and masks. SAM~2 \citep{ravi2024sam2segmentimages} extended it
to video, while \samthree \citep{carion2025sam} extended it to open-vocabulary
\emph{concept} prompts by training a DETR-based detector-tracker
\citep{carion2020end,kamath2021mdetr} with data from a purpose-built data
engine.
Open-vocabulary detection and segmentation methods
\citep{gu2021open,minderer2022simple,ding2022open,liang2023open} leverage
vision-language encoders such as CLIP \citep{CLIP}. GLIP
\citep{li2022grounded} and GroundingDINO \citep{groundingdino} cast detection
as phrase grounding, with OWLv2 \citep{minderer2024scalingopenvocabularyobjectdetection},
LLMDet \citep{fu2025llmdet}, APE \citep{Shen_2024_CVPR}, DINO-X
\citep{ren2025dinoxunifiedvisionmodel}, GLEE \citep{glee}, and T-Rex2
\citep{trex2} pushing scale and prompt modalities. All of these are
\emph{trained specialists}. \ours instead treats segmentation as an emergent
capability of a general MLLM, requiring no segmentation training at all, and
uses only text prompts (no exemplars, boxes, or video, unlike \samthree).

\mysection{MLLMs for grounding and segmentation}
A large family attaches a trained mask decoder to an (M)LLM. LISA
\citep{lai2024lisa} introduced reasoning segmentation by decoding a
\texttt{<SEG>} token with SAM. GLaMM \citep{rasheed2024glamm}, GSVA
\citep{xia2024gsva}, OMG-LLaVA \citep{zhang2024omg}, PSALM
\citep{zhang2024psalm}, EVF-SAM \citep{zhang2024evf}, HyperSeg
\citep{wei2024hyperseg}, and X-SAM \citep{wang2025x} refine this recipe.
SAM4MLLM \citep{chen2024sam4mllm} and Seg-Zero \citep{liu2025seg} train the
MLLM to emit prompts for SAM. Training-free alternatives still delegate the
pixels to a segmentation model. RSVP \citep{lu2025rsvp} and attention-based
grounding \citep{kang2025your} use the MLLM for localization only, and
\samagent \citep{carion2025sam} lets an MLLM drive \samthree as a tool,
similar in spirit to set-of-mark prompting \citep{yang2023setofmark}.
Native MLLM segmentation has appeared only as coarse per-token masks in
Gemini~2.5 \citep{comanici2025gemini,geminiseg2025}. \samthree reports that
these masks lag far behind those of specialists. \ours is, to our knowledge,
the first pipeline in which a \emph{frozen, general-purpose} MLLM produces the
final mask geometry itself, as parametric splines rather than pixels. No
downstream component is trained, and the pipeline achieves accuracy competitive
with trained systems. The prior work closest to this output space is discussed
next.

\mysection{MLLMs that emit coordinates as text}
Writing geometry down as tokens rather than decoding it is not itself new. What
is new here is that the tokens are the final answer. In prior work the emitted
text is an intermediate consumed by a trained head, and the pipeline around it
is trained. VisionLLM \citep{wang2023visionllm} is the closest precedent. It
represents an instance boundary as a short polygon, typically 16 points, emitted
by the language model, but keeps a trained decoding head per task, which both
ties generalization to the task set seen in training and caps mask resolution at
the vertex budget. VisionLLM~v2 \citep{wu2024visionllm} scales that recipe to
hundreds of vision-language tasks by routing a special per-task token to a
decoder fine-tuned for that task, while fine-tuning the MLLM for each task.
PixelLLM \citep{xu2024pixel} adds an MLP head to regress locations and an
encoder to embed an input box, giving the MLLM localization and grounding. It
trains the pipeline end to end and still obtains masks by prompting SAM with a
box. Text4Seg \citep{lan2025text4seg} makes the most extensive use of text as
an output space: it defines segmentation tokens over a row-ordered grid that is
encoded and decoded so that masks are produced by generating text, yet the MLLM
must be fine-tuned to generate them. Beyond segmentation,
\citet{jiang2026detect} cast detection as
next-point prediction over the same kind of structured coordinate text, training
the full pipeline with supervised fine-tuning followed by reinforcement learning,
and report detection and grounding but no masks. PixelVLA
\citep{liang2026pixelvla} carries pixel-level grounding into
vision-language-action models with trained pixel modules. Perceive Anything
\citep{lin2026perceive} instead couples SAM~2 with a modern LLM in a single
jointly trained pipeline for fine-grained region understanding, and its gains
are in recognition, explanation and captioning: the masks come from the SAM~2
branch, not from the language model. Frontier general-purpose models are
acquiring coarse localization on their own, with Molmo
\citep{deitke2025molmo} able to point at queried objects and the Gemini series
grounding boxes well \citep{comanici2025gemini,geminiseg2025}, but neither comes
close to a specialized segmentation network or \samthree at the pixel level.
\ours departs from all of the above in one consistent way: the backend is frozen
and general-purpose, nothing downstream of it is trained, and the control points
it writes out are the prediction itself rather than a prompt, a latent, or a
seed for a decoder.

\mysection{Agentic MLLM pipelines}
Beyond single-call inference, MLLMs increasingly act as agents that decompose
visual tasks, call tools, and verify intermediate results
\citep{fan2024videoagent,ge2023planting,yang2023setofmark}, including for
large-scale annotation \citep{carion2025sam}. Colon-Bench
\citep{HamdiAbd_ColonBench_MICCAI2026} shows such a workflow producing dense
lesion masks in a niche clinical domain where \samthree fails. However, the
masks are drawn by EdgeTAM \citep{zhou2025edgetam}, a compact SAM~2 tracker
invoked by the MLLM as a tool. The pixel-level competence therefore remains
with the tool. \ours follows this philosophy
with a fixed, budgeted loop: every stage is a structured JSON exchange, and
deterministic geometry code (rasterization, clamping, composition) provides
the reliability that free-form agent loops lack. The iterative
focus-box refinement can be seen as the model \emph{self-verifying} its own
rendered prediction, an ability that improves with backend reasoning strength
(\secLabel\ref{sec:analysis}).

\mysection{Contour and curve parameterizations of masks}
Describing a mask by its boundary rather than by a pixel grid has a decades-long
history. Active contour models \citep{kass1988snakes} evolve a spline under
internal smoothness and external image forces, and \citet{brigger2000bspline}
showed that the optimal solution to the curvature-constrained snake energy is a
cubic spline, the classical argument for the representation we adopt. Deep
learning recast the same idea as vertex prediction. Polygon-RNN
\citep{castrejon2017annotating} and Polygon-RNN++ \citep{acuna2018efficient}
emit polygon vertices autoregressively for interactive annotation. Curve-GCN
\citep{ling2019fast} predicts all vertices at once with a graph network and
supports spline as well as polygon outputs. Deep Snake \citep{peng2020deep}
deforms an initial contour with circular convolutions, and PolarMask
\citep{xie2020polarmask} regresses boundary rays in polar coordinates.
PolyTransform \citep{liang2020polytransform} and SharpContour
\citep{zhu2022sharpcontour} deform the polygon of an existing mask to sharpen
it, and E2EC \citep{zhang2022e2ec} learns the contour initialization end to end.
BoundaryFormer \citep{lazarow2022instance} regresses polygons with a
transformer supervised only through a differentiable rasterizer. Parametric
curves are standard in text spotting as well, where ABCNet
\citep{liu2020abcnet} localizes curved text with B\'ezier control points.
\ours shares its output space with this literature but little else. Each method
above is a network trained to regress vertices from image features over a fixed
label set, whereas \ours obtains control points as text from a frozen
general-purpose MLLM, without added contour or mask supervision or a closed
vocabulary. This representation makes that possible: a handful of control
points is short enough to serve as a language-model output, and focus-box
confinement lets the refinement loop add or subtract local regions without
retracing the whole outline.

\clearpage
\section{Method Details} \label{app:pipeline}
\vspace{-4pt}

\subsection{Structured replies and coordinate conventions}
\vspace{-2pt}

\mysection{Prompts and output schemas}
Each stage uses a single prompt with a strict structured-output schema. Detect
returns a list of detections, each with a box, a label, a confidence, and a
morphology description. When topology conditioning is enabled, it also returns
the component and hole counts. For each instance ID, Segment returns a list of
spline loops, each with a role (outer boundary or hole), a degree, and control
points. The model chooses a degree of $1$ or $3$, shared by every loop of every
instance in the call. The refinement schema has no degree field because focus
loops are always cubic. The two refinement calls return, respectively, typed
focus boxes tagged with their instance IDs and confined spline loops for every
surviving focus box. Each instance line in the segment prompt restates its box
as explicit numeric inequalities. Points are clamped in any case, but a
clamped-in point produces a worse outline than a correctly placed one.
\S\ref{app:prompts} gives the templates themselves.

\mysection{Coordinate conventions across backends}
Everything passed between stages is expressed in absolute pixels, with the
origin at the top left. Boxes use the order $[x_1,y_1,x_2,y_2]$. Gemini-family
models natively emit boxes in the order
$[y_{\min},x_{\min},y_{\max},x_{\max}]$ on a grid from $0$ to $1000$.
Requesting $x$-first order does not solve the problem: on some multi-instance
images, these models transpose their output, and no downstream check can catch
this because a swapped box is still well formed. We therefore request each
family's \emph{native} order and convert it afterward, transposing before
rescaling. All families return control points in $[x,y]$ order. If a loop is
transposed, we detect and repair it against its box: the swapped interpretation
lies almost entirely inside the box, whereas the raw interpretation does not.
Unlike boxes, points can be checked this way because their box is known ground
truth.

\mysection{Recovering truncated replies}
Batched replies covering several instances can hit output limits. Rather than
discard a truncated reply, we keep every instance completed before the cut and
treat the rest as missing. The refinement stage can still grow them from their
boxes. The prompts fix the emission order by ascending instance ID, so
truncation affects the tail rather than an arbitrary subset. We recover Detect
replies in the same way. An instance that survives with no loops is scored as
an empty mask and is never dropped from the item.

\subsection{Rasterization and composition} \label{app:implementation_details}
\vspace{-2pt}

\mysection{Rasterization}
A cubic loop is rasterized with
$m=\min\{2000,\max\{64,\operatorname{round}(2(w_b+h_b))\}\}$ samples
for a rasterization box $b$ of width $w_b$ and height $h_b$. Before clipping,
this corresponds to roughly one sample per pixel of the box perimeter. Samples
include both endpoints, as in \eqn{eq:raster}. Degree-1 loops use the closed
control polygon directly. The box is $b_k$ for an initial instance loop and
$\beta_j$ for a refinement loop.

\mysection{Composing a refinement pass}
$\mathrm{Compose}$ in \eqn{eq:refine} is deterministic. Naively reapplying
the signed regions each pass makes the mask oscillate: a boundary pixel added
by one pass is removed by the next and added again. We therefore carry a signed
\emph{commitment} field alongside each instance mask, marking every pixel as
committed to foreground, committed to background, or uncommitted, with a
magnitude that records the strength of agreement across passes. A refinement writes
freely into uncommitted pixels. One that agrees with an existing commitment
reinforces it. One that opposes it must first cancel one unit of commitment,
and only flips the pixel once the accumulated resistance is spent. Repeated
agreement makes a pixel progressively harder to overturn. Setting the unit an
opposing edit cancels ($\alpha$ below) to zero would freeze committed pixels
permanently.
\eqn{eq:softlock} gives the per-pixel rule.

Two deterministic steps bracket the composition. Before the re-tracing call,
an \emph{effect gate} drops any focus box whose largest achievable change to
the current mask is negligible, and a pass whose boxes are all dropped skips its
re-tracing call. Afterwards, a deterministic cleanup, described below, removes
specks and spurious holes.

\mysection{The soft lock, per pixel}
This expands $\mathrm{Compose}$ in \eqn{eq:refine}. Let $\Lambda_k^{(t)}$
be instance $k$'s signed commitment field after pass $t$, initialized to zero
before refinement. Fix a pass $t$, a focus box $j$ that corrects instance
$k=k_j$, and a pixel $x$ in its correction region
$\mathcal R_{\beta_j,3}(P_j)$. Write $a$ for the pixel's current binary mask
value and $\ell$ for its current signed commitment, initially
$a=\mathbf{1}[x\in M_k^{(t-1)}]$ and $\ell=\Lambda_k^{(t-1)}(x)$.
These values carry forward after each correction within the pass.
Let $\chi=\operatorname{sgn}\ell$ and $v=|\ell|-\alpha$, the resistance
remaining after an opposing edit erodes it by $\alpha$ ($\alpha=1$
throughout).
With $\sigma=\sigma_j$, one correction updates the pixel to
\begin{equation}
  (a',\ell') =
  \begin{cases}
    (\mathbf{1}[\sigma=+1],\;\sigma)
      & \ell=0 \quad\text{(free: commit)},\\[2pt]
    (\mathbf{1}[\sigma=+1],\;\ell+\sigma)
      & \chi=\sigma \quad\text{(agree: reinforce)},\\[2pt]
    (a,\;\chi\,v)
      & \chi=-\sigma,\; v>0 \quad\text{(oppose: erode)},\\[2pt]
    (\mathbf{1}[\sigma=+1],\;\sigma)
      & \chi=-\sigma,\; v\leq0 \quad\text{(oppose: flip)}.
  \end{cases}
  \label{eq:softlock}
\end{equation}
Here $\mathbf{1}[A]$ is $1$ when statement $A$ is true and $0$ otherwise.
Primes denote a single correction, not a complete refinement pass.
After all corrections and the cleanup below, the resulting mask and field
are $M_k^{(t)}$ and $\Lambda_k^{(t)}$, and the union is $M^{(t)}$.
Negative regions are applied before positive ones within a pass. After
composition, nearby positive islands disconnected from the main mask are
bridged. A deterministic cleanup then applies a minor closing and opening,
sized as percentages of the instance box's minimum dimension, and removes
specks and spurious holes using thresholds scaled by object area. When the
stage-1 topology descriptor is used, its component and hole counts gate this
cleanup so that objects predicted to have holes retain them.

\subsection{One item, step by step} \label{app:walkthrough}
\vspace{-2pt}

\begin{figure}[p]
  \centering
  \setlength{\tabcolsep}{1pt}
  \setlength{\qgw}{0.204\linewidth}
  \renewcommand{\arraystretch}{0.7}
  \begin{tabular}{@{}m{0.098\linewidth}@{\hspace{3pt}}cccc@{}}
    & \qhd{Input} & \qhd{Stage 1: box} & \qhd{Stage 2: control points} & \qhd{Stage 2: mask} \\[1pt]
    \wlab{Initialize}{1 component, 0 holes} & \qim{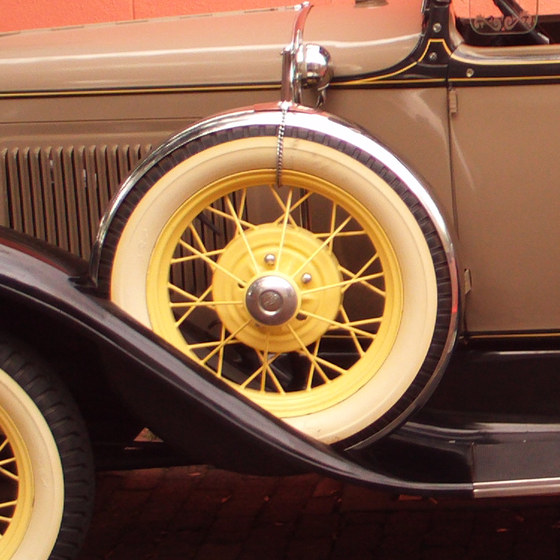} & \qim{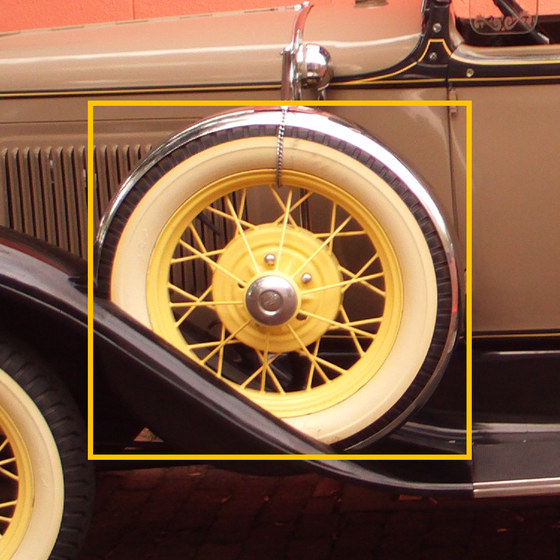} & \qim{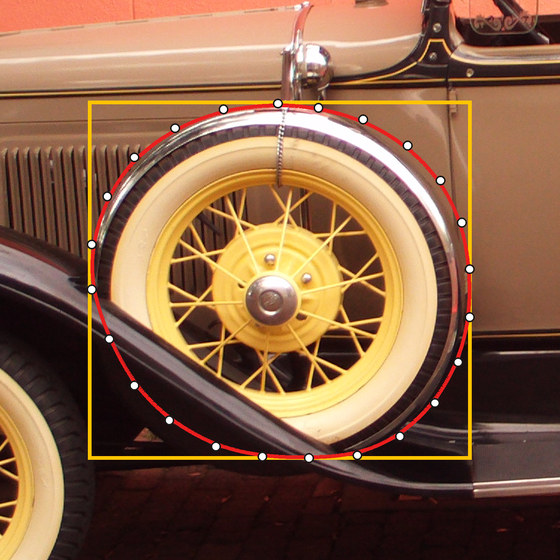} & \qim{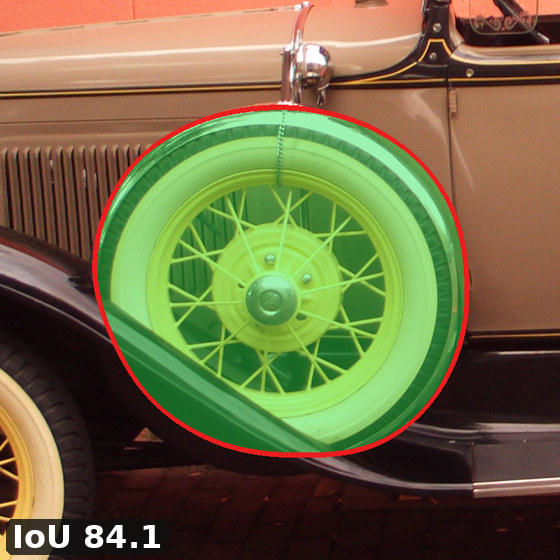} \\
  \end{tabular}

  \vspace{1pt}
  \begin{tabular}{@{}m{0.098\linewidth}@{\hspace{3pt}}cccc@{}}
    & \qhd{(a) focus boxes} & \qhd{(b) re-traced loops} & \qhd{(c) mask after the pass} & \qhd{(d) what changed} \\[1pt]
    \wlab{Pass 1}{1 box, $-$} & \qim{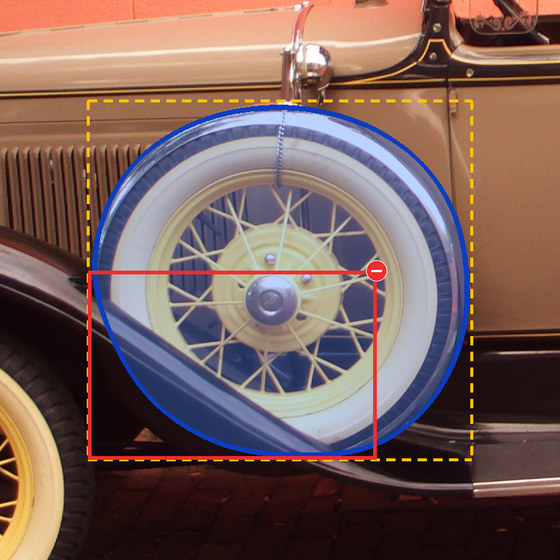} & \qim{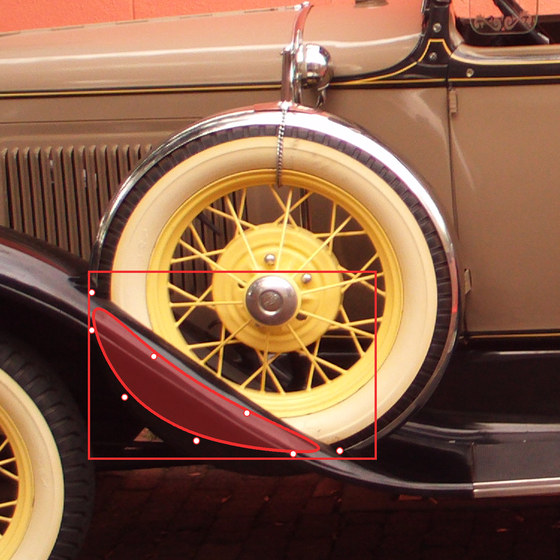} & \qim{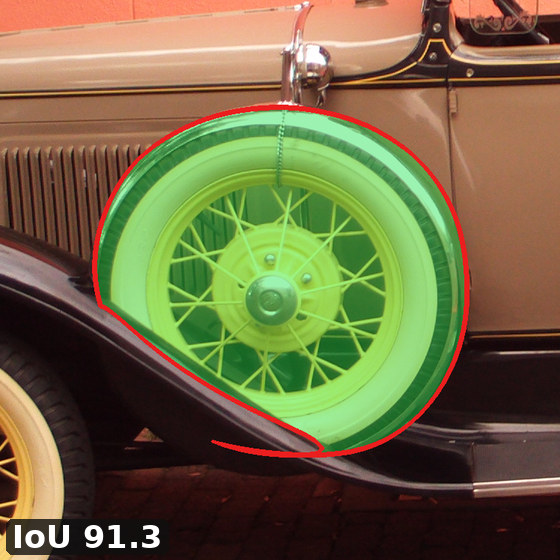} & \qim{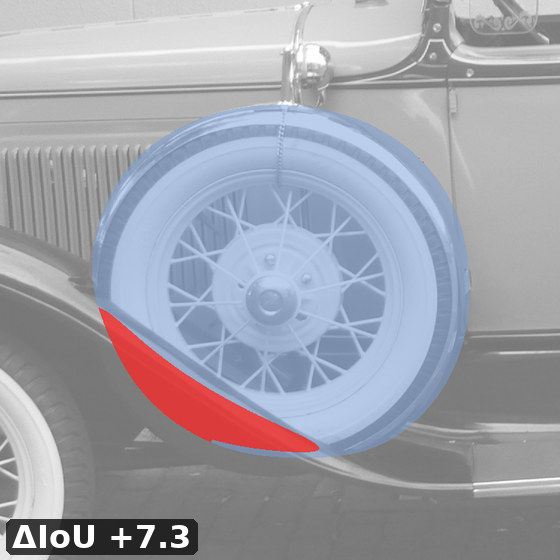} \\
    \wlab{Pass 3}{1 box, $+$} & \qim{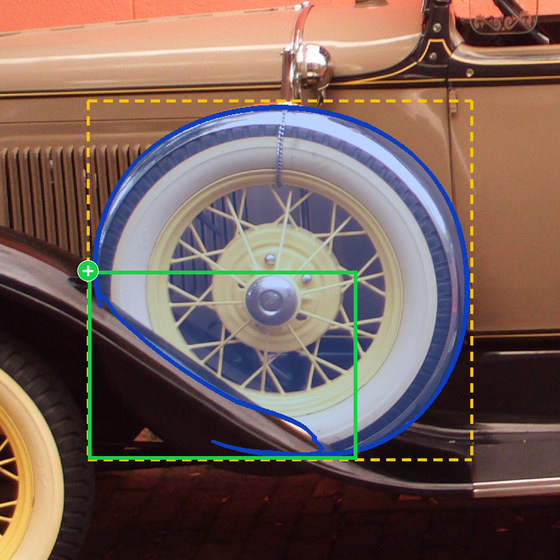} & \qim{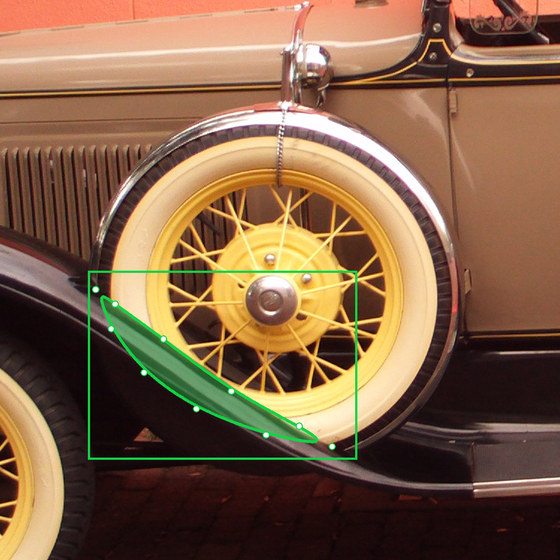} & \qim{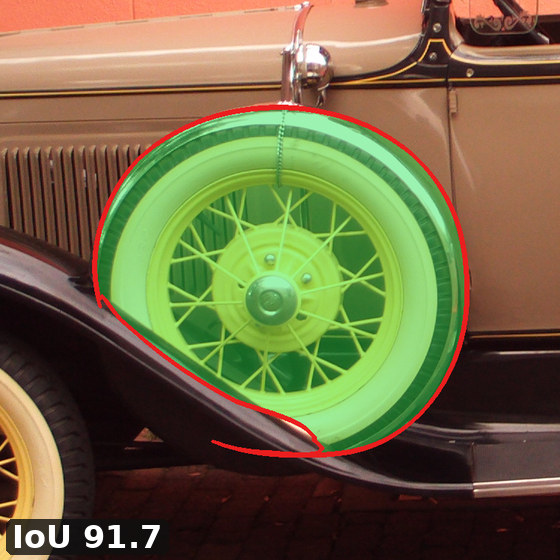} & \qim{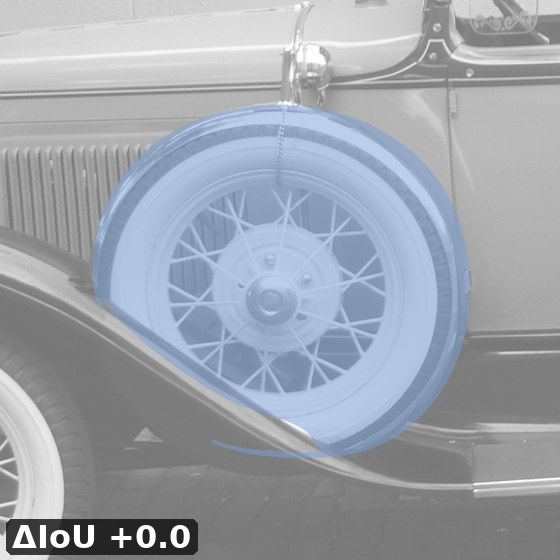} \\
    \wlab{Pass 4}{1 box, $-$} & \qim{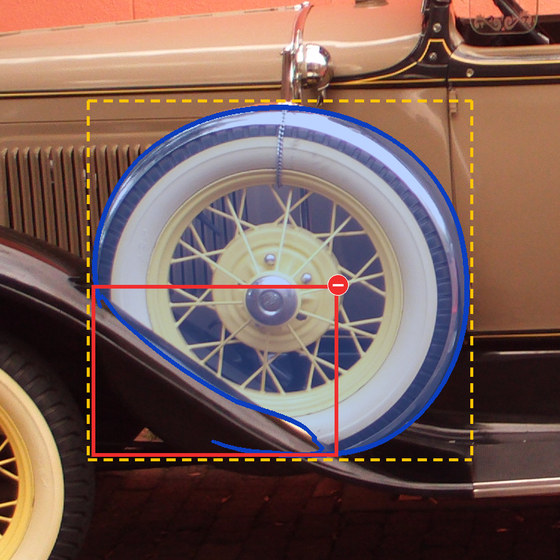} & \qim{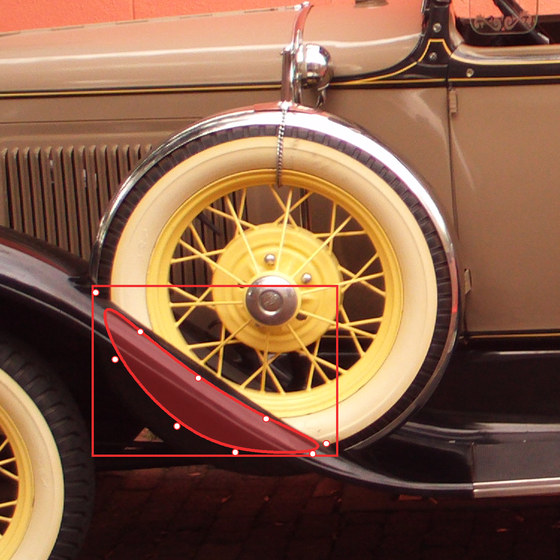} & \qim{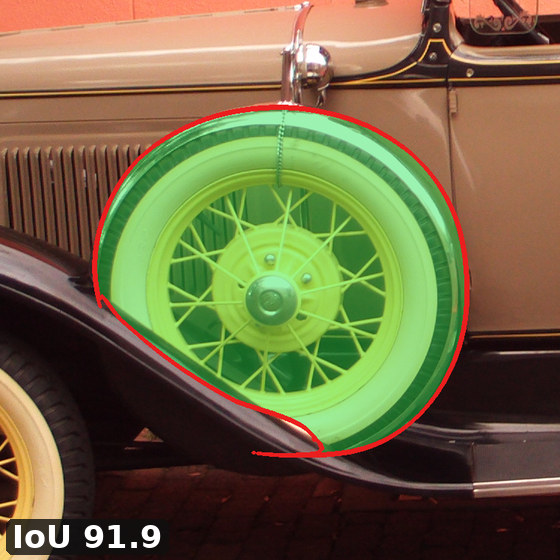} & \qim{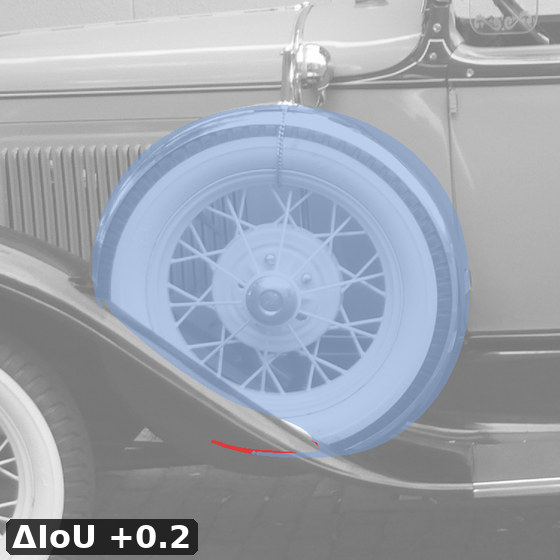} \\
  \end{tabular}

  \vspace{1pt}
  \begin{tabular}{@{}m{0.098\linewidth}@{\hspace{3pt}}cccc@{}}
    & \qhd{One loop, enlarged} & \qhd{Converged mask} & \qhd{Ground truth} & \qhd{IoU after each pass} \\[1pt]
    \wlab{Result}{stopped at 4 passes} & \qim{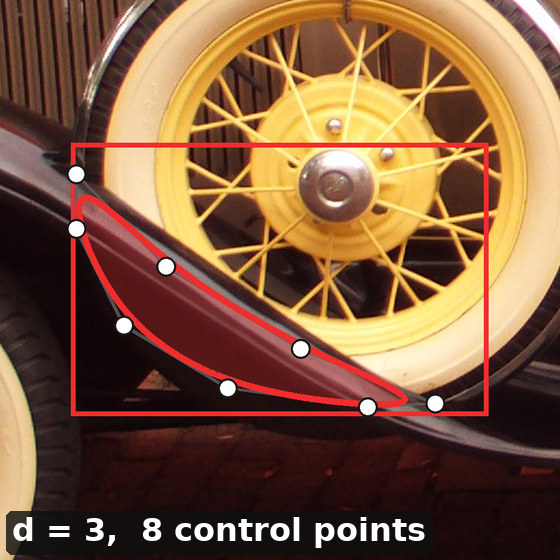} & \qim{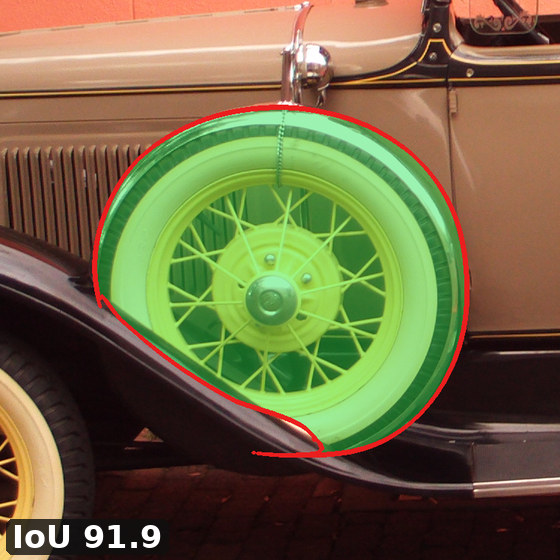} & \qim{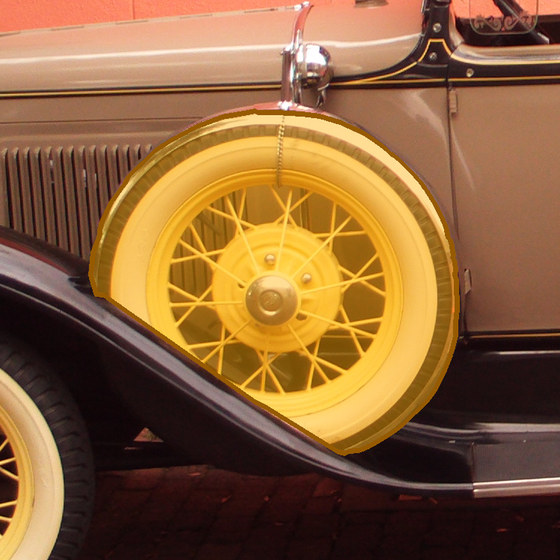} & \qim{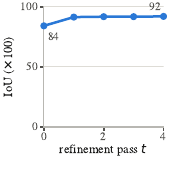} \\
  \end{tabular}
  \vspace{-6pt}
  \caption{\small \textbf{A step-by-step view of \ours on one ReasonSeg test item},
  \emph{``the spare wheel''}, with the \gfl backend. \emph{Top:} stage 1's
  box, then stage 2's control points and the resulting closed cubic B-spline.
  \emph{Middle:} three of the four passes, each of which proposes signed
  focus boxes over the current mask and composes the re-traced loops. The net
  effect is shown in green (added) and red (removed). \emph{Bottom:} one loop
  enlarged, the converged mask, the ground truth, and the IoU after each pass.}
  \label{fig:walkthrough}
\end{figure}

\figLabel\ref{fig:walkthrough} follows a single ReasonSeg item through every
call the pipeline makes, on the query \emph{``the spare wheel''}. Stage 2 is
already competent here, scoring $84.1$ IoU, and the loop takes it to $91.9$:
this is the ordinary case, not a rescue. What makes it a useful example is
\emph{what} is wrong with the stage-2 trace. The wheel is mounted behind a
black fender that cuts across its lower-left, and the outline runs straight
past the occlusion, so the entire error is one wedge of car body inside an
otherwise good boundary. Nothing in the figure is drawn for illustration. Every
box, control point and loop is the geometry the model actually returned for
that item, and every mask is rebuilt from those records by the pipeline's own
composition code, which reproduces the stage-2 and final masks exactly.

Three things stand out in the figure. First, the spline is
cubic: stage 2 chose $d{=}3$ and placed $24$ control points around a circle,
all inside the supplied box. \eqn{eq:bspline} turns them into a smooth closed
curve rather than a $24$-gon. Second, the loop is mostly subtractive. Pass 1
removes $15{,}118$ pixels of fender, $9.4\%$ of the object's area, and
increases IoU by $7.3$ points. The two later passes shave slivers off the same
edge. Third, pass 3 is the interesting one. It proposes \emph{adding} a wedge
back along that lower-left edge, but the mask is effectively unchanged:
composing the proposal against the commitment field left by the first two
passes changes only one pixel. Against an empty field, the same proposal would
have added $7{,}791$ pixels (a further $4.8\%$ of the object). This is the
signed resistance of \eqn{eq:refine} doing its job: it refuses an edit that
contradicts earlier commitments instead of letting the boundary oscillate.
The loop then stopped itself after $4$ of the $12$ allowed passes.

\clearpage
\section{Experimental Details} \label{app:eval}
\vspace{-4pt}

\subsection{Benchmarks}
\vspace{-2pt}

We evaluate on five families of benchmarks (\tblLabel\ref{tab:benchmarks}).
An item is always one (image, query) pair, and the query is the only prompt any
system receives. ReasonSeg \citep{lai2024lisa} names its target by role or
function rather than category, so a system must resolve the query before it
can segment the target. RefCOCO and RefCOCO+ \citep{kazemzadeh2014referitgame} come
from a two-player game on COCO images, so their expressions disambiguate one
instance from same-class neighbours. The testA split contains images crowded with people,
whereas testB contains images crowded with other categories. RefCOCOg
\citep{mao2016generation} was collected non-interactively. LVIS \citep{lvis}
uses federated annotations, so we establish absence only from its per-image
verified-absent list and draw our negative prompts from that list. COCO
\citep{coco} is exhaustively annotated, so an unannotated category is absent by
construction. SA-Co/Gold \citep{carion2025sam} spans seven domains and is
annotated three times (we use the first annotation). These three benchmarks
report the presence-gated \cgf\ of \citet{carion2025sam}. ADE20K-150
\citep{ade20k} and Cityscapes \citep{cordts2016cityscapes} are semantic
benchmarks. ODinW13 \citep{odinw} and RF-100VL
\citep{robicheaux2025roboflow100vl} comprise 13 and 100 small detection
datasets, respectively. They are prompted with their category names and
provide boxes rather than masks.

\begin{table}[h]
\centering
\small
\setlength{\tabcolsep}{4pt}
\resizebox{\textwidth}{!}{
\begin{tabular}{l l r r r l l}
\toprule
\textbf{Benchmark (splits)} & \textbf{Query (words)} & \textbf{Split} & \textbf{Neg.} & \textbf{Scored} & \textbf{Ground truth} & \textbf{Metric} \\
\midrule
ReasonSeg val / test & reasoning sentence (18.3 / 20.7) & 200 / 779 & 2.0 / 0.3 & all & one mask per query & gIoU, cIoU \\
\midrule
RefCOCO val / testA / testB & referring expression (3.6) & 10{,}834 / 5{,}657 / 5{,}095 & 0 & 500 each & one instance & gIoU, cIoU \\
RefCOCO+ val / testA / testB & no location words (3.6) & 10{,}758 / 5{,}726 / 4{,}889 & 0 & 500 each & one instance & gIoU, cIoU \\
RefCOCOg val / test & written description (8.5) & 4{,}896 / 9{,}602 & 0 & 500 each & one instance & gIoU, cIoU \\
\midrule
LVIS val & category name (1.4) & 129{,}547 & 45.9 & 1{,}500 & every instance & \cgf, \pmf, AP \\
COCO val2017 & category name (1.2) & 29{,}631 & 50.6 & 1{,}500 & every instance & \cgf, \pmf, AP \\
SA-Co/Gold & open noun phrase (2.6) & 168{,}202 & 84.5 & 2{,}500 & every instance & \cgf, \pmf \\
\midrule
ADE20K-150 val & class name (1.0) & 16{,}103 & 0 & 1{,}000 & all pixels of class & gIoU, cIoU \\
Cityscapes val & 19 class names in one query & 500 & 0 & 500 & 19-class label map & mIoU \\
\midrule
ODinW13 test & category name & 6{,}237 (13 subsets) & 0 & all & box, filled & AP$_0$ \\
RF-100VL test & category name & 1{,}987 (100 subsets) & 0 & all & box, filled & AP$_0$ \\
\bottomrule
\end{tabular}}

\vspace{3pt}
\caption{\small \textbf{Benchmarks.} Query type with mean query length in words, the
number of items in each split after loading, the share of items whose correct
answer is an empty mask, the number of items we score, and the reported metric.
An item is one (image, query) pair. Negative shares are percentages of the total number of items in the split.}
\label{tab:benchmarks}
\end{table}

\subsection{Configuration of \ours} \label{app:config}
\vspace{-2pt}

The main results use \gfl as the backend for all
three stages. For each call, the backend chooses a stage-2 curve degree
$d\in\{1,3\}$, shared across the instances in that call
(\secLabel\ref{sec:method}). Refinement curves are always cubic. Their
rasterization is described in \secLabel\ref{app:implementation_details}.
The budget constants and stopping rule are given in
\secLabel\ref{sec:experiments}. A full run costs 2 MLLM calls (detect and
segment) plus at most 2 per refinement pass, so at most $2T{+}2 = 18$. Measured
over ReasonSeg val, the reported configuration averages $13.4$ calls per image (median $14$), because
the effect gate and early stopping terminate many loops well
before $T$ (\figLabel\ref{fig:residual}). Mean wall-clock time is
$159$\,s per image. Runs at $T{=}12$ use more passes and average $14.3$ to
$15.0$ calls per image, depending on the run (\secLabel\ref{app:convergence},
\secLabel\ref{app:ablations_other}, \secLabel\ref{app:cost}). The stage-1 topology descriptor
(\secLabel\ref{sec:method}) is an option rather than a fixed part of the
configuration, and the main results leave it \emph{off} on all three stages.
\secLabel\ref{sec:analysis} and \secLabel\ref{app:ablations_other} report both settings, and the difference
between them is within run-to-run variation. These settings are
used unchanged on every benchmark. Nothing is tuned per dataset.

\mysection{Study configuration}
The backend study (\secLabel\ref{app:backend_study}) and the ablations
(\secLabel\ref{app:ablations_other}) were run with a budget of $T{=}12$ and
topology conditioning on. We call this the study configuration. It differs
from the reported configuration in exactly these two settings: rows inside one
study are comparable with each other, while a comparison between a study row
and a main-results number mixes both differences. In the ablation tables the
reference row uses the reported setting (topology conditioning off) at $T{=}12$,
and the \emph{+ topology conditioning} row is the study configuration.

\subsection{Evaluation protocol and matched-subset scoring}
\vspace{-2pt}

\mysection{Evaluation protocol}
Four of the 200 ReasonSeg val items and two of the 779 test items have a
legitimately \emph{empty} ground-truth mask: the queried thing is absent from
the image. We score these items rather than excluding them, so a system that
correctly predicts nothing is credited, while one that hallucinates a region is
penalized. This distinction matters because dropping these items raises the
reported gIoU of systems that do not handle absence well and makes our val
numbers incomparable to the published ReasonSeg results we quote.
For LVIS/COCO/ADE20K, presence-gated metrics use 3, 3, and 0 sampled negative
prompts per image, respectively, with seed 0 shared across all systems.
Cityscapes mIoU is computed from corpus-level, per-class intersections and
unions over the 19 classes: \samthree is queried one class at a time, while
\ours labels all classes in an image with a single query and groups spline
instances by label. All backends use the same deterministic composition: stuff
masks are painted before thing masks, with larger masks first within each
group, so smaller masks overwrite larger ones. Uncovered pixels remain
unassigned. No confidence argmax is used. ADE20K is class-prompted, with each
item defined as (image, class). Its score is a union-IoU oracle upper bound and
is not comparable to published argmax mIoU.

\mysection{Matched-subset scoring}
Item keys are constructed in the data loader before items are split by backend,
so every backend sees the same items. gIoU/cIoU/\cgf/\pmf/mIoU are recomputed
from per-item components over the key intersection. AP is re-evaluated with
COCOeval on the matched items only, with boxes derived from masks for all
systems. For \ours, pipeline failures (detection or segmentation errors) count
as zero.

Five evaluations use their entire split, so their key intersection is the full
split and their results can be compared with published full-split results.
These are ReasonSeg val and test, Cityscapes val mIoU, and AP$_0$ on both
detection collections, averaged over all 13 ODinW13 subsets and the 99 of 100
RF-100VL subsets on which AP is defined. Sampling affects box and mask AP on LVIS and
COCO. Because AP ranks detections across categories, an image can enter the
evaluation only if every ground-truth category it contains was queried. Our
items are uniformly sampled (image, category) pairs, so usually only one
category is queried for each sampled image, and the image is excluded: only 48
of $1{,}449$ LVIS images and 62 of $1{,}304$ COCO images survive. The surviving
images are skewed toward those with few categories. These AP results are
not comparable to published full-split federated AP. We use them only
to compare systems on the same images. Every other reported metric is computed
per item and is unaffected by this sampling issue.

\subsection{Reproducing the \samagent baseline}
\vspace{-2pt}

\samagent is the agent baseline of \citet{carion2025sam}, not part of \ours.
Our reproduction follows their reference agent implementation. The MLLM
iterates over three tools: one runs \samthree on a simple noun phrase and
returns the resulting masks as a numbered overlay, another selects the mask
numbers that match the full query, and the third reports that nothing in the
image matches.
The loop is capped at 8 rounds, and the union of the selected masks is scored
through the same path as every other system. We run it with Gemma-4-31B,
Gemini-3.5-Flash and \gfl rather than the paper's Qwen2.5-VL and Gemini~2.5~Pro
backends, so our \samagent scores are somewhat below the published ones
(\tblLabel\ref{tab:reasonseg}). The Gemini-3.5-Flash run is reported only in
the cost table (\tblLabel\ref{tab:cost}).

\subsection{Backend study protocol} \label{app:backends}
\vspace{-2pt}

\mysection{Protocol}
The backend study in \secLabel\ref{sec:analysis} holds \ours completely fixed
and changes only which MLLM answers its three calls. All 21 backends run the
full pipeline on all 200 ReasonSeg val items in the study configuration
(\secLabel\ref{app:config}), with every other setting fixed. Metrics are
recomputed from the per-item records rather than copied from any summary, so
gIoU is the mean per-item IoU and cIoU is
$\nicefrac{\sum\text{I}}{\sum\text{U}}$ over all 200 items, with pipeline
failures scored $0$. Confidence intervals are percentile bootstraps over items
(2000 resamples). Each call has a wall-clock budget. Backends with high
serving latency (marked $\dagger$ in \tblLabel\ref{tab:backends}) receive a
longer budget so that provider latency, rather than model capability, does not
determine their scores. The budget limits only how long a call may take. It
does not alter any reply. Qwen3-VL-235B is included despite failing to return a
usable reply on $49\%$ of items, so its score reflects reply rate as well as
segmentation quality. Every row of \tblLabel\ref{tab:backends} covers the same
200 items. The two reference lines in \figLabel\ref{fig:backends} show
\samthree and \samagent scored with the same code on the same items. These are
the ReasonSeg val entries from \tblLabel\ref{tab:reasonseg_full}. The \samagent
line uses \gfl as its driver, and its gIoU varies by only $0.7$ points
($72.4$, $72.5$, $73.1$) across the three MLLMs we ran it with, compared with
$10.8$ points for \ours across one Gemini Flash product line.

\mysection{Serving the backends}
The eight Gemini and Gemma models are called through Google's native Gemini
API, and the other thirteen through OpenRouter\footnote{\url{https://openrouter.ai}},
a gateway for models from many developers. The route changes only how a request
is transported, not the prompt, parser, or geometry code. Images are sent at
their original resolution because returned coordinates use the image's pixel
frame. Sampling parameters remain at each provider's defaults, so every
backend is evaluated as its developer serves it. A call that fails or exceeds
its budget is a pipeline failure scored $0$. OpenRouter's public listings also
provide the prices used in \secLabel\ref{app:cost}.

\mysection{The capability axes}
Public benchmark coverage is uneven: no single published benchmark is reported
for all 21 backends, and the available benchmarks use incomparable scales. We
build each capability axis by z-scoring every benchmark \emph{within
this set of backends} and averaging the axis's benchmarks for which a model
reports a score. Missing values are never imputed: models with no published
score for an axis are omitted from that axis's panel. The four axes are
\textbf{software engineering} (AA Coding Index, SWE-bench Verified, SWE-bench
Pro, LiveCodeBench),
\textbf{reasoning and knowledge} (AA Intelligence Index, GPQA Diamond, HLE),
\textbf{agentic and tool use} (AA Agentic Index, OSWorld, Terminal-Bench), and
\textbf{vision and multimodal} (MMMU, MMMU-Pro, CharXiv). Two consequences follow. Each axis is \emph{relative}: adding a backend
re-centres it, so its zero carries no meaning beyond this set. And the
underlying numbers are vendor-reported, so the slopes and rank correlations we
report are indicative of which skills matter, not claims of statistical
significance.

\subsection{Cost accounting} \label{app:cost_protocol}
\vspace{-2pt}

\begin{figure}[h]
  \centering
  \includegraphics[width=\textwidth]{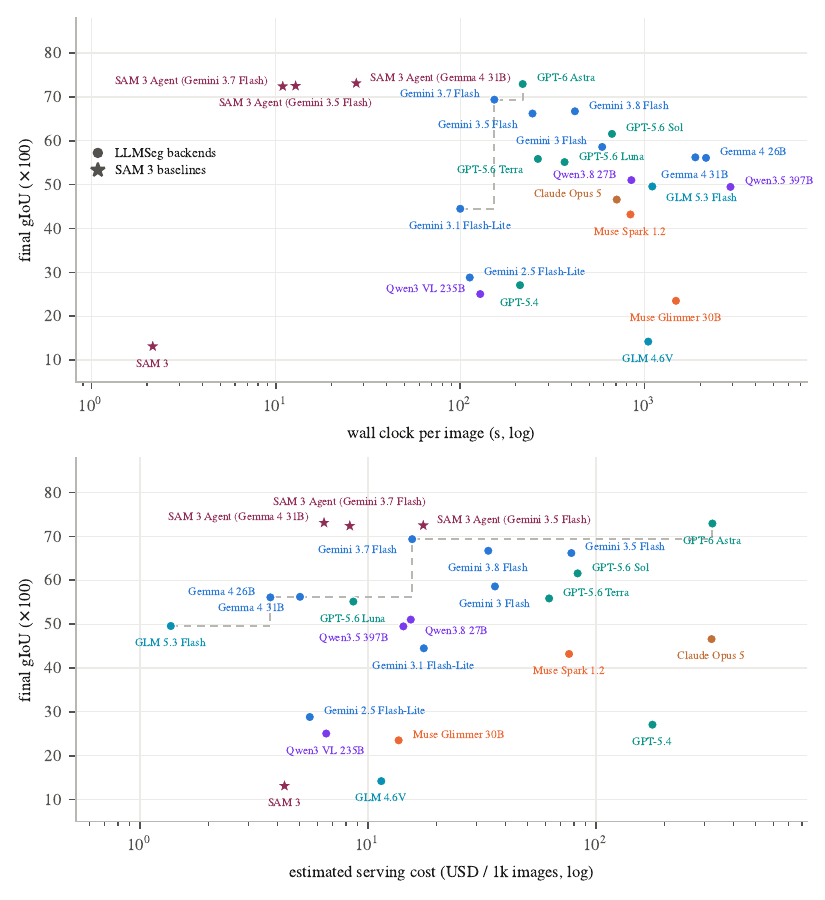}
  \vspace{-6pt}
  \caption{\small \textbf{Quality against what it costs.} ReasonSeg val, 200 items,
  every \ours backend and baseline, plotted against wall-clock time (\emph{top}) and
  estimated serving cost (\emph{bottom}). Both axes are logarithmic and the dashed staircase is the
  cheapest-so-far frontier over the \ours backends. Circles are \ours backends,
  stars the \samthree and \samagent baselines, which the frontier excludes.}
  \label{fig:cost}
\end{figure}

\mysection{Protocol}
\tblLabel\ref{tab:cost} and \figLabel\ref{fig:cost} price the same 200
ReasonSeg val items scored in \tblLabel\ref{tab:backends}, covering all
twenty-one \ours backends and both the \samthree and \samagent baselines.
Nothing here was
rerun: every number is either measured from records those runs already wrote
or quoted from the published model table. \emph{Wall clock} is the median
seconds per item among the items a backend answered, based on per-item
timers recorded during the runs. We use the median rather than the mean because
a run that hits its per-call budget on a handful of items can have a mean that
reflects the timeout rather than the backend. The \emph{call count} includes
both types of model invoked for each image. The MLLM call count is reconstructed from each
item's recorded pass list. It counts one detect and one segment call, plus one focus
call per refinement pass and one batched spline call for each pass whose focus
boxes survived the effect gate. For \samagent, it is the mean number of agent
rounds. The \samthree figure is the number of forward passes of the
$860$\,M-parameter model. It counts one pass for
\samthree itself and one per phrase the agent chooses to segment. It counts
none for \ours, which calls no segmentation model. \samthree and \samagent were timed with \samthree running on one RTX-class GPU. Prices and time-to-first-token
come from the providers' published figures. Call counts come from the single
study run of each backend, so they can differ slightly from repeated-run means
elsewhere (for \gfl, $15.0$ here against $14.3$ in \tblLabel\ref{tab:ablations}).

\mysection{The dollar column is an estimate, and a lower bound}
The spend column is an estimate rather than a billed measurement. We price
every row using one shared budget: $1900$ input tokens per call (the stage
prompt templates, which measure $522$ to $953$ tokens, plus one image with a
full-resolution allowance of about $1.1$k) and $175$ output tokens per call
(the mean structured payload recorded across all twenty-one runs). The column
depends only on the published price and the measured call count, so it ranks
backends consistently under these assumptions. Its
absolute values are lower bounds because reasoning tokens, which several
providers bill for, are not included.

\mysection{\samthree is not free either}
We ran \samthree on a local GPU, so we incurred no charge, but a forward
pass has a market price. Pricing it at zero would flatter \ours. We charge
each \samthree pass at $\$0.0043$ per image, the pay-as-you-go rate for a
serverless endpoint serving the same checkpoint,\footnote{Segmind serverless
inference, pay-as-you-go credits, retrieved 2026-09.} which puts \samthree at
$\$4.30$ per thousand images for one pass and adds the cost of $1.4$ passes to
each \samagent image, in addition to its MLLM calls. \ours calls no segmentation
model, so it carries no such term.

\clearpage
\section{Additional Results} \label{app:results}
\vspace{-4pt}

\subsection{Complete reasoning and referring comparisons} \label{app:full_comparisons}
\vspace{-2pt}

\tblLabel\ref{tab:reasonseg} compares selected baselines on ReasonSeg and
RefCOCO/+/g using cIoU. The complete tables, including gIoU and the training
columns, are reproduced here.

\begin{table}[!htbp]
\centering
\resizebox{\linewidth}{!}{
\begin{tabular}{ll cc cc cc}
\toprule
& & \multicolumn{2}{c}{\textbf{Training}} & \multicolumn{2}{c}{\textbf{val}} & \multicolumn{2}{c}{\textbf{test}} \\
\cmidrule(lr){3-4}\cmidrule(lr){5-6}\cmidrule(lr){7-8}
\textbf{Method} & \textbf{MLLM} & RES & ReasonSeg & gIoU & cIoU & gIoU & cIoU \\
\midrule
\multicolumn{8}{l}{\textit{Published, full splits} \citep{carion2025sam}} \\
SEEM~\citep{zou2023segment}            & \na             & \xmark & \xmark & 25.5 & 21.2 & 24.3 & 18.7 \\
Grounded SAM~\citep{ren2401grounded}   & \na             & \xmark & \xmark & 26.0 & 14.5 & 21.3 & 16.4 \\
OVSeg~\citep{liang2023open}            & \na             & \xmark & \xmark & 28.5 & 18.6 & 26.1 & 20.8 \\
GLaMM~\citep{rasheed2024glamm}         & Vicuna 7B       & \cmark & \xmark & 47.4 & 47.2 & \na & \na \\
SAM4MLLM~\citep{chen2024sam4mllm}      & Qwen-VL 7B      & \cmark & \xmark & 46.7 & 48.1 & \na & \na \\
SAM4MLLM~\citep{chen2024sam4mllm}      & LLaVA1.6 8B     & \cmark & \xmark & 58.4 & 60.4 & \na & \na \\
Seg-Zero~\citep{liu2025seg}            & Qwen2.5-VL 3B   & \cmark & \xmark & 58.2 & 53.1 & 56.1 & 48.6 \\
Seg-Zero~\citep{liu2025seg}            & Qwen2.5-VL 7B   & \cmark & \xmark & 62.6 & 62.0 & 57.5 & 52.0 \\
X-SAM~\citep{wang2025x}                & Phi3 3.8B       & \cmark & \cmark & 56.6 & 32.9 & 57.8 & 41.0 \\
HyperSeg~\citep{wei2024hyperseg}       & Phi2 3B         & \cmark & \cmark & 59.2 & 56.7 & \na & \na \\
Kang et al.~\citep{kang2025your}       & LLaVA1.5 7B     & \xmark & \xmark & \na & 52.4 & \na & 48.7 \\
Kang et al.~\citep{kang2025your}       & LLaVA1.5 13B    & \xmark & \xmark & \na & 60.5 & \na & 49.9 \\
LISA~\citep{lai2024lisa}               & LLaVA 7B        & \cmark & \xmark & 44.4 & 46.0 & 36.8 & 34.1 \\
LISA~\citep{lai2024lisa}               & LLaVA 7B        & \cmark & \cmark & 52.9 & 54.0 & 47.3 & 34.1 \\
LISA~\citep{lai2024lisa}               & LLaVA 13B       & \cmark & \xmark & 48.9 & 46.9 & 44.8 & 45.8 \\
LISA~\citep{lai2024lisa}               & LLaVA 13B       & \cmark & \cmark & 56.2 & 62.9 & 51.7 & 51.1 \\
LISA~\citep{lai2024lisa}               & Llama2 13B      & \cmark & \cmark & 60.0 & 67.8 & 51.5 & 51.3 \\
LISA~\citep{lai2024lisa}               & LLaVA1.5 7B     & \cmark & \xmark & 53.6 & 52.3 & 48.8 & 47.1 \\
LISA~\citep{lai2024lisa}               & LLaVA1.5 7B     & \cmark & \cmark & 61.3 & 62.9 & 55.6 & 56.9 \\
LISA~\citep{lai2024lisa}               & LLaVA1.5 13B    & \xmark & \xmark & 57.7 & 60.3 & 53.8 & 50.8 \\
LISA~\citep{lai2024lisa}               & LLaVA1.5 13B    & \cmark & \cmark & 65.0 & 72.9 & 61.3 & 62.2 \\
RSVP~\citep{lu2025rsvp}                & LLaVA1.6 7B     & \xmark & \xmark & 59.2 & 56.7 & 56.9 & 50.7 \\
RSVP~\citep{lu2025rsvp}                & Qwen2-VL 7B     & \xmark & \xmark & 58.6 & 48.5 & 56.1 & 51.6 \\
RSVP~\citep{lu2025rsvp}                & Gemini1.5-Flash & \xmark & \xmark & 56.9 & 49.2 & 57.1 & 59.2 \\
RSVP~\citep{lu2025rsvp}                & GPT-4o          & \xmark & \xmark & 64.7 & 63.1 & 60.3 & 60.0 \\
Gemini Seg~\citep{geminiseg2025}       & Gemini2.5 Flash & ? & ? & 28.3 & 13.3 & 30.6 & ~9.2 \\
\samagent~\citep{carion2025sam}        & Qwen2.5-VL 7B   & \xmark & \xmark & 62.2 & 49.1 & 63.0 & 53.5 \\
\samagent~\citep{carion2025sam}        & Qwen2.5-VL 72B  & \xmark & \xmark & 74.6 & 65.1 & 70.8 & 64.0 \\
\samagent~\citep{carion2025sam}        & Llama4 Maverick & \xmark & \xmark & 68.5 & 61.5 & 67.1 & 60.9 \\
\samagent~\citep{carion2025sam}        & Qwen3-VL 8B Think  & \xmark & \xmark & 68.5 & 57.4 & 70.2 & 67.3 \\
\samagent~\citep{carion2025sam}        & Qwen3-VL 235B Think & \xmark & \xmark & 76.6 & \textbf{67.0} & 73.7 & 71.8 \\
\samagent~\citep{carion2025sam}        & Gemini2.5 Pro   & \xmark & \xmark & \textbf{77.0} & 66.0 & \textbf{74.0} & \textbf{72.1} \\
\midrule
\samthree (text query)~\citep{carion2025sam} & \na       & \xmark & \xmark & 13.1 & ~9.9 & ~8.4 & 11.3 \\
\samagent~\citep{carion2025sam}        & Gemma-4-31B     & \xmark & \xmark & \textbf{73.1} & 63.9 & \textbf{72.2} & \textbf{66.7} \\
\samagent~\citep{carion2025sam}        & \gfl            & \xmark & \xmark & 72.4 & 62.0 & \textbf{72.2} & 63.2 \\
\rowcolor[HTML]{EFEFEF}
\oursnr                                & \gfl            & \xmark & \xmark & 64.9 & 64.6 & 62.7 & 58.2 \\
\rowcolor[HTML]{EFEFEF}
\oursfull                              & \gfl            & \xmark & \xmark & 69.6 & \textbf{71.5} & 68.3 & 65.8 \\
\bottomrule
\end{tabular}}
\caption{\small \textbf{Reasoning segmentation on ReasonSeg} \citep{lai2024lisa}.
Top block: published numbers quoted from \citet{carion2025sam} on the full
splits, reproducing its complete baseline list. Bottom block (below the rule):
our own evaluations (200 validation items and 779 test items). Both blocks cover the full splits.
\samthree and \samagent are both baselines from \citet{carion2025sam}. Only the
shaded rows are ours. The two training columns follow \citet{carion2025sam}:
RES = fine-tuned for referring expression segmentation, ReasonSeg = fine-tuned
on the ReasonSeg training split. A method is zero-shot when both entries are
\xmark. \oursnr is our pipeline stopped after the segment stage. The best
zero-shot result in each column is shown in \textbf{bold}, separately for each
block.}
\label{tab:reasonseg_full}
\end{table}

\begin{table}[!htbp]
  \centering
  \resizebox{\linewidth}{!}{
  \begin{tabular}{ll c ccc ccc cc}
  \toprule
  & & & \multicolumn{3}{c}{\textbf{RefCOCO}} & \multicolumn{3}{c}{\textbf{RefCOCO+}} & \multicolumn{2}{c}{\textbf{RefCOCOg}} \\
  \cmidrule(lr){4-6}\cmidrule(lr){7-9}\cmidrule(lr){10-11}
  \textbf{Method} & \textbf{MLLM / backbone} & \textbf{RES} & val & testA & testB & val & testA & testB & val (U) & test (U) \\
  \midrule
  \multicolumn{11}{l}{\textit{Published, full splits, cIoU} \citep{carion2025sam}} \\
  LISA~\citep{lai2024lisa}                  & LLaVA 7B          & \cmark & 74.9 & 79.1 & 72.3 & 65.1 & 70.8 & 58.1 & 67.9 & 70.6 \\
  GSVA~\citep{xia2024gsva}                  & 13B               & \cmark & 79.2 & 81.7 & 77.1 & 70.3 & 73.8 & 63.6 & 75.7 & 77.0 \\
  GLaMM~\citep{rasheed2024glamm}            & Vicuna 7B         & \cmark & 79.5 & 83.2 & 76.9 & 72.6 & 78.7 & 64.6 & 74.2 & 74.9 \\
  SAM4MLLM~\citep{chen2024sam4mllm}         & LLaVA1.6 7B       & \cmark & 79.6 & 82.8 & 76.1 & 73.5 & 77.8 & 65.8 & 74.5 & 75.6 \\
  SAM4MLLM~\citep{chen2024sam4mllm}         & LLaVA1.6 8B       & \cmark & 79.8 & 82.7 & 74.7 & 74.6 & 80.0 & 67.2 & 75.5 & 76.4 \\
  GLEE~\citep{glee}                         & Plus              & \cmark & 79.5 & \na  & \na  & 68.3 & \na  & \na  & 70.6 & \na \\
  GLEE~\citep{glee}                         & Pro               & \cmark & 80.0 & \na  & \na  & 69.6 & \na  & \na  & 72.9 & \na \\
  DETRIS~\citep{huang2025densely}           & DETRIS-L          & \cmark & 81.0 & 81.9 & 79.0 & 75.2 & 78.6 & 70.2 & 74.6 & 75.3 \\
  UniLSeg~\citep{liu2024universal}          & UniLSeg-20        & \cmark & 80.5 & 81.8 & 78.4 & 72.7 & 77.0 & 67.0 & 78.4 & 79.5 \\
  UniLSeg~\citep{liu2024universal}          & UniLSeg-100       & \cmark & 81.7 & 83.2 & 79.9 & 73.2 & 78.3 & 68.2 & 79.3 & 80.5 \\
  PSALM~\citep{zhang2024psalm}              & Phi1.5 1.3B       & \cmark & 83.6 & 84.7 & 81.6 & 72.9 & 75.5 & 70.1 & 73.8 & 74.4 \\
  EVF-SAM~\citep{zhang2024evf}              & RC                & \cmark & 82.1 & 83.7 & 80.0 & 75.2 & 78.3 & 70.1 & 76.8 & 77.4 \\
  EVF-SAM~\citep{zhang2024evf}              & Extra Data        & \cmark & 82.4 & 84.2 & 80.2 & 76.5 & 80.0 & 71.9 & 78.2 & 78.3 \\
  RICE~\citep{xie2025region}                & Qwen2.5-7B        & \cmark & 83.5 & 85.3 & 81.7 & \textbf{79.4} & 82.8 & 75.4 & 79.8 & 80.4 \\
  MLCD-seg~\citep{an2024multi}              & Qwen2.5-7B        & \cmark & 83.6 & 85.3 & 81.5 & \textbf{79.4} & 82.9 & \textbf{75.6} & 79.7 & 80.5 \\
  HyperSeg~\citep{wei2024hyperseg}          & Phi2 2.7B         & \cmark & 84.8 & 85.7 & \textbf{83.4} & 79.0 & \textbf{83.5} & 75.2 & 79.4 & 78.9 \\
  X-SAM~\citep{wang2025x}                   & Phi3 3.8B         & \cmark & \textbf{85.1} & \textbf{87.1} & \textbf{83.4} & 78.0 & 81.0 & 74.4 & \textbf{83.8} & \textbf{83.9} \\
  \addlinespace[1pt]
  GL-CLIP~\citep{yu2023zero}                & ResNet-50         & \xmark & 32.7 & 35.3 & 30.1 & 37.7 & 40.7 & 34.9 & 41.6 & 42.9 \\
  GL-CLIP~\citep{yu2023zero}                & ViT-B/32          & \xmark & 32.9 & 34.9 & 30.1 & 38.4 & 42.1 & 32.7 & 42.0 & 42.0 \\
  CaR~\citep{sun2024clip}                   & ViT-B/16          & \xmark & 33.6 & 35.4 & 30.5 & 34.2 & 36.0 & 31.0 & 36.7 & 36.6 \\
  Ref-Diff~\citep{ni2023ref}                & VAE               & \xmark & 37.2 & 38.4 & 37.2 & 37.3 & 40.5 & 33.0 & 44.0 & 44.5 \\
  TAS~\citep{suo2023text}                   & ResNet-50         & \xmark & 39.9 & 42.9 & 35.9 & 44.0 & 50.6 & 36.4 & 47.7 & 47.4 \\
  TAS~\citep{suo2023text}                   & ViT-B/32          & \xmark & 39.8 & 41.1 & 36.2 & 43.6 & 49.1 & 36.5 & 46.6 & 46.8 \\
  IteRPrimeE~\citep{wang2025iterprime}      & \na               & \xmark & 40.2 & 46.5 & 33.9 & 44.2 & 51.6 & 35.3 & 46.0 & 45.1 \\
  Pseudo-RIS~\citep{yu2024pseudo}           & CRIS              & \xmark & 39.8 & 44.8 & 33.0 & 42.2 & 46.3 & 34.5 & 43.7 & 43.4 \\
  Pseudo-RIS~\citep{yu2024pseudo}           & ETRIS             & \xmark & 41.1 & 48.2 & 33.5 & 44.3 & 51.4 & 35.1 & 46.0 & 46.7 \\
  LGD+DINO~\citep{li2025lgd}                & ViT-B/32          & \xmark & 49.5 & 54.7 & 41.0 & 49.6 & 58.4 & 38.6 & 50.3 & 51.1 \\
  VLM-VG~\citep{wang2025learning}           & ResNet-50         & \xmark & 47.7 & 51.8 & 44.7 & 41.2 & 45.9 & 34.7 & 46.6 & 47.1 \\
  VLM-VG~\citep{wang2025learning}           & ResNet-101        & \xmark & 49.9 & 53.1 & 46.7 & 42.7 & 47.3 & 36.2 & 48.0 & 48.5 \\
  HybridGL~\citep{liu2025hybrid}            & ViT-B/32          & \xmark & 49.5 & 53.4 & 45.2 & 43.4 & 49.1 & 37.2 & 51.3 & 51.6 \\
  \citet{kang2025your}                      & LLaVA-1.5 7B      & \xmark & 74.2 & 76.5 & 70.4 & 62.5 & 65.2 & 56.0 & 64.2 & 68.1 \\
  \citet{kang2025your}                      & LLaVA-1.5 13B     & \xmark & \underline{76.1} & \underline{78.9} & \underline{72.8} & 64.1 & 67.1 & 57.3 & 67.7 & 69.0 \\
  \samagent~\citep{carion2025sam}           & Qwen2.5-VL 7B     & \xmark & 59.4 & 64.3 & 55.0 & 51.4 & 57.0 & 44.9 & 57.2 & 58.8 \\
  \samagent~\citep{carion2025sam}           & Qwen2.5-VL 72B    & \xmark & 71.6 & 74.9 & 66.0 & 64.9 & 70.8 & 57.9 & 68.8 & 70.2 \\
  \samagent~\citep{carion2025sam}           & Llama4 Maverick   & \xmark & 71.7 & 76.5 & 66.5 & 65.1 & 71.2 & 57.6 & 67.7 & 68.7 \\
  \samagent~\citep{carion2025sam}           & Qwen3-VL 8B Think & \xmark & 68.6 & 72.3 & 63.9 & 59.3 & 64.5 & 55.6 & 66.4 & 66.9 \\
  \samagent~\citep{carion2025sam}           & Qwen3-VL 235 Think& \xmark & 73.9 & 77.0 & 69.7 & 66.9 & 70.9 & 62.2 & 72.1 & 72.9 \\
  \samagent~\citep{carion2025sam}           & Gemini 2.5 Pro    & \xmark & 75.5 & 77.6 & 71.0 & \underline{67.3} & \underline{71.1} & \underline{63.4} & \underline{73.4} & \underline{74.0} \\
  \midrule
  \multicolumn{11}{l}{\textit{Our runs, matched 500-item subsets, gIoU}} \\
  \samthree (text query) & \na          & \xmark & 38.7 & 40.9 & 40.9 & 32.1 & 35.0 & 28.0 & 40.0 & 38.7 \\
  \samagent              & Gemma-4-31B  & \xmark & 74.9 & 75.6 & \textbf{76.7} & 69.8 & 72.9 & 67.1 & 75.5 & 76.5 \\
  \samagent              & \gfl         & \xmark & \textbf{75.3} & \textbf{76.9} & 76.1 & \textbf{71.2} & \textbf{74.2} & \textbf{71.2} & \textbf{78.4} & \textbf{77.3} \\
  \rowcolor[HTML]{EFEFEF}
  \oursnr                & \gfl         & \xmark & 66.0 & 66.2 & 72.1 & 61.9 & 63.3 & 64.0 & 67.1 & 68.5 \\
  \rowcolor[HTML]{EFEFEF}
  \oursfull              & \gfl         & \xmark & 71.6 & 73.0 & 74.8 & 68.2 & 70.8 & 67.6 & 72.7 & 73.0 \\
  \midrule
  \multicolumn{11}{l}{\textit{Our runs, matched 500-item subsets, cIoU}} \\
  \samthree (text query) & \na          & \xmark & 42.0 & 42.8 & 44.2 & 33.6 & 37.0 & 32.9 & 40.0 & 39.7 \\
  \samagent              & Gemma-4-31B  & \xmark & \textbf{73.2} & 75.6 & 75.0 & 67.2 & \textbf{72.0} & 61.9 & 72.1 & 74.7 \\
  \samagent              & \gfl         & \xmark & \textbf{73.2} & \textbf{76.7} & 75.5 & \textbf{68.4} & \textbf{72.0} & \textbf{68.5} & \textbf{76.5} & \textbf{76.9} \\
  \rowcolor[HTML]{EFEFEF}
  \oursnr                & \gfl         & \xmark & 65.5 & 65.7 & 72.9 & 59.7 & 61.4 & 61.1 & 64.8 & 69.3 \\
  \rowcolor[HTML]{EFEFEF}
  \oursfull              & \gfl         & \xmark & 71.2 & 73.8 & \textbf{76.0} & 67.0 & 69.9 & 64.9 & 70.9 & 74.5 \\
  \bottomrule
  \end{tabular}}
  \caption{\small \textbf{Referring expression segmentation} on RefCOCO/+/g
  \citep{kazemzadeh2014referitgame,mao2016generation}. The \emph{top block}
  reports published full-split cIoU results quoted from
  \citet{carion2025sam}, separated into methods fine-tuned on the RES task
  (RES~\cmark) and zero-shot methods (RES~\xmark). Note that nearly all of the
  MLLM-based methods were trained on the RefCOCO \emph{box} annotations. The
  \emph{bottom blocks} report our runs on matched 500-item random subsets per
  split (shuffled, seed 0), with gIoU in the upper block and cIoU in the lower
  block. All methods in the bottom blocks are zero-shot for RES and use the same
  selected items.
  \samthree and \samagent are baselines from \citet{carion2025sam},
  reproduced by us on these items. Only the shaded rows show our methods. Best
  results within each block are in \textbf{bold}. The best zero-shot result in
  each published-block column is \underline{underlined}. RefCOCOg val~(U) and
  test~(U) are the splits used in our evaluation.}
  \label{tab:refcoco_full}
  \end{table}
  
  \subsection{Complete open-vocabulary comparison}
  \label{app:other_full_comparisons}
  
  \tblLabel\ref{tab:openvocab} reports results for LVIS and COCO only. The
  complete comparison, including \pmf, SA-Co/Gold results, and the human
  reference, follows.
  
  \begin{table}[!htbp]
  \centering
  \resizebox{\linewidth}{!}{
  \begin{tabular}{ll cccc cccc cc}
  \toprule
  & & \multicolumn{4}{c}{\textbf{LVIS}} & \multicolumn{4}{c}{\textbf{COCO}} & \multicolumn{2}{c}{\textbf{SA-Co/Gold}} \\
  \cmidrule(lr){3-6}\cmidrule(lr){7-10}\cmidrule(lr){11-12}
  \textbf{Method} & \textbf{MLLM} & \cgf & \pmf & box AP & mask AP & \cgf & \pmf & box AP & mask AP & \cgf & \pmf \\
  \midrule
  Human                  & \na & \na & \na & \na & \na & \na & \na & \na & \na & 72.8 & \na \\
  OWLv2~\citep{minderer2024scalingopenvocabularyobjectdetection}
                         & \na & 20.1 & \na & 35.2 & \na  & \na & \na & 38.2 & \na & 17.3 & \na \\
  OWLv2 (LVIS-trained)   & \na & 29.3 & \na & 45.5 & 43.4 & \na & \na & 46.1 & \na & 24.6 & \na \\
  gDino-T~\citep{groundingdino}
                         & \na & 14.7 & \na & 20.5 & \na  & \na & \na & 45.7 & \na & ~3.3 & \na \\
  LLMDet-L~\citep{fu2025llmdet}
                         & \na & 35.1 & \na & 42.0 & 36.3 & \na & \na & 55.6 & \na & ~6.5 & \na \\
  APE-D (LVIS-trained)~\citep{Shen_2024_CVPR}
                         & \na & \na  & \na & 59.6 & 53.0 & \na & \na & 58.3 & \na & 16.4 & \na \\
  DINO-X~\citep{ren2025dinoxunifiedvisionmodel}
                         & \na & \na  & \na & 52.4 & 38.5 & \na & \na & 56.0 & \na & 21.3 & \na \\
  Gemini 2.5 Flash~\citep{comanici2025gemini}
                         & \na & 13.4 & \na & \na & \na  & \na & \na & \na  & \na & 13.0 & \na \\
  \samthree~\citep{carion2025sam}
                         & \na & 37.2 & \na & 53.6 & 48.5 & \na & \na & 56.4 & \na & 54.1 & \na \\
  SAM~3$^{*}$            & \na         & 50.1 & \underline{58.0} & \textbf{69.1} & \textbf{62.2} & \underline{52.4} & 57.6 & \textbf{68.0} & \textbf{59.2} & \textbf{45.4} & \underline{55.2} \\
  SAM~3 Agent$^{*}$      & Gemma-4-31B & \underline{51.0} & \textbf{62.4} & 56.2 & \underline{50.9} & 49.9 & \underline{58.6} & 48.9 & 42.0 & \na & \na \\
  \rowcolor[HTML]{EFEFEF}
  \oursnr                & \gfl        & 50.4 & 53.0 & 57.1 & 36.9 & 50.4 & 54.0 & 59.6 & 38.0 & 31.9 & 48.6 \\
  \rowcolor[HTML]{EFEFEF}
  \oursfull              & \gfl        & \textbf{53.9} & 57.3 & \underline{64.2} & 47.2 & \textbf{56.1} & \textbf{60.6} & \underline{66.6} & \underline{50.5} & \underline{36.4} & \textbf{55.4} \\
  \bottomrule
  \end{tabular}}
  \caption{\small \textbf{Open-vocabulary instance segmentation.} Rows without
  asterisks or shading report published full-split results quoted from
  \citet{carion2025sam}. Rows marked with $^{*}$ show our evaluations of the
  SAM~3 baselines on matched random subsets with hard negatives, using the same
  items as our methods (shaded rows). \cgf is presence-gated F1, and \pmf is
  its positive-only counterpart. Within the starred and shaded rows
  (our matched block), the best and second-best result per column are shown in
  \textbf{bold} and \underline{underlined}, respectively. Published rows and
  the human reference are not ranked.}
  \label{tab:openvocab_full}
  \end{table}

  \subsection{Qualitative results on the main benchmarks} \label{app:moved_qualitative}
  \vspace{-2pt}
  
\begin{figure}[p]
    \centering
    \setlength{\tabcolsep}{1pt}
    \setlength{\qgw}{0.175\linewidth}
    \renewcommand{\arraystretch}{1.0}
    \begin{tabular}{@{}m{0.23\linewidth}@{\hspace{3pt}}cccc@{}}
      & \qhd{Input} & \qhd{Ground truth} & \qhd{Initial trace} & \qhd{\ours} \\[1pt]
      \qlab{LVIS}{lampshade} & \qim{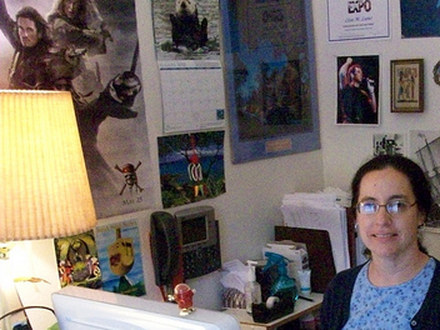} & \qim{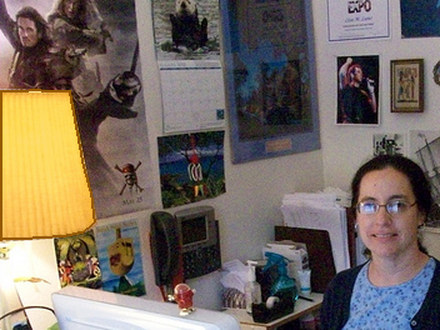} & \qim{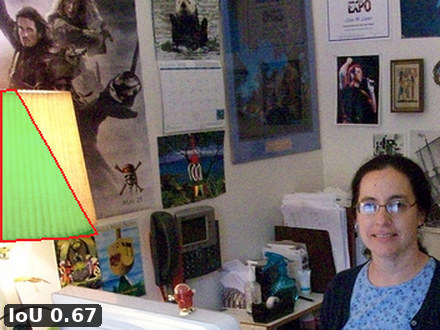} & \qim{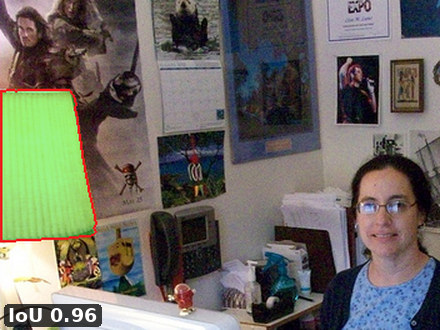} \\
      \qlab{COCO}{person} & \qim{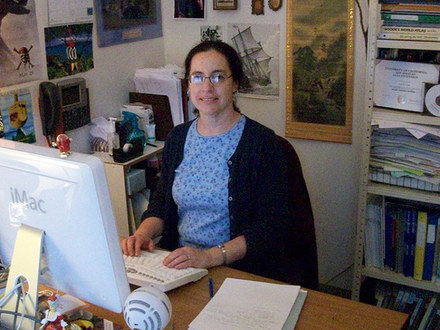} & \qim{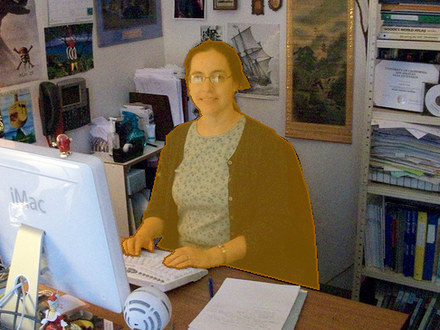} & \qim{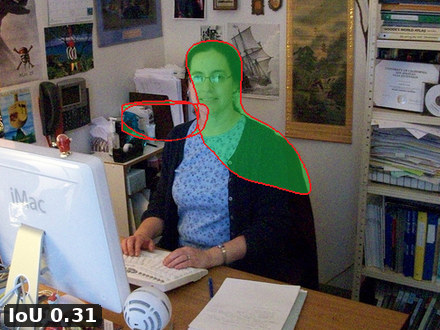} & \qim{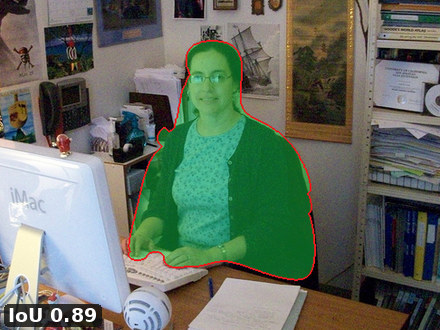} \\
      \qlab{SA-Co/Gold}{suede shoe} & \qim{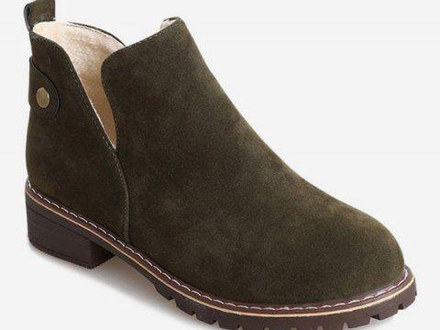} & \qim{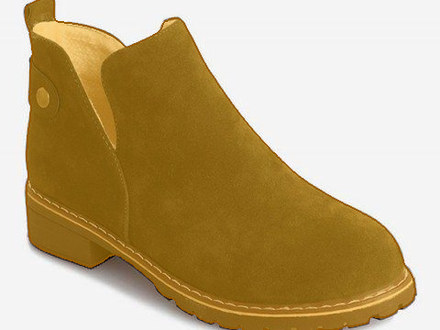} & \qim{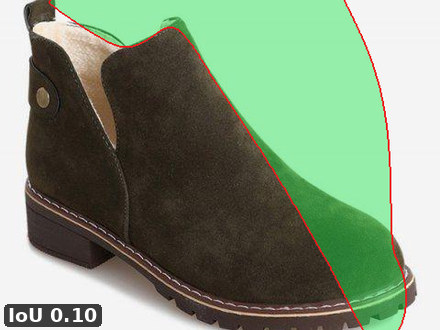} & \qim{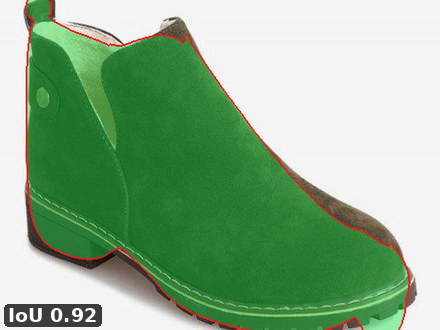} \\
      \qlab{SA-Co/Gold}{metal pipe} & \qim{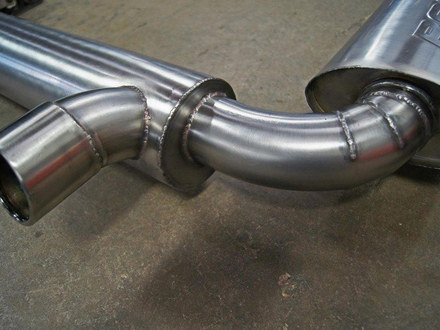} & \qim{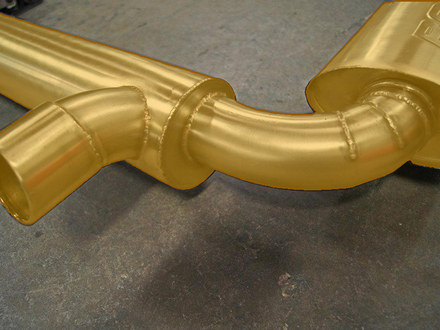} & \qim{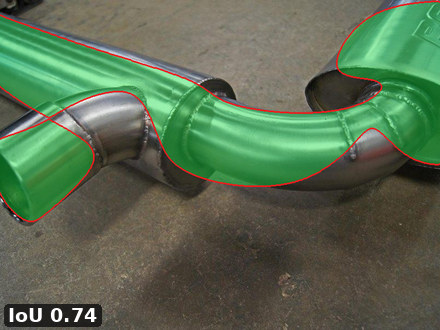} & \qim{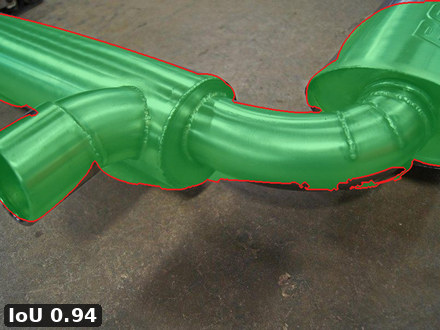} \\
      \qlab{RefCOCOg}{a pink and white striped sofa with soft pastel pillow\ldots} & \qim{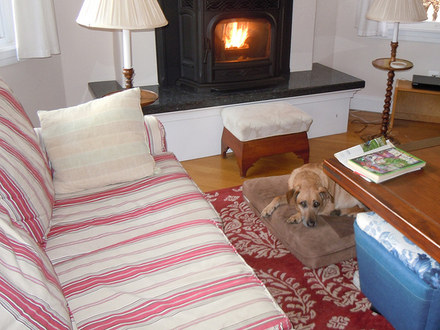} & \qim{qual_bench/refcocog_sofa_gt.jpg} & \qim{qual_bench/refcocog_sofa_seg.jpg} & \qim{qual_bench/refcocog_sofa_rec.jpg} \\
      \qlab{RefCOCO}{left sandwich} & \qim{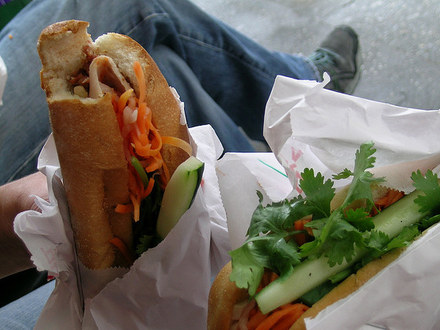} & \qim{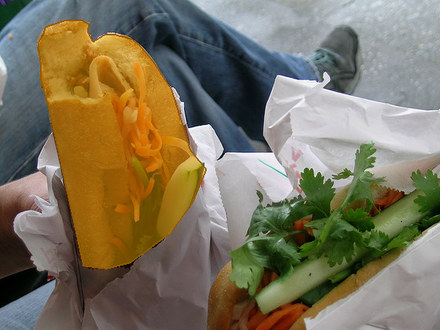} & \qim{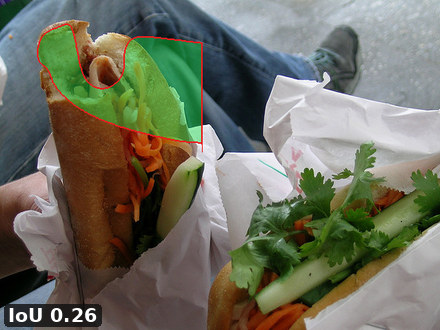} & \qim{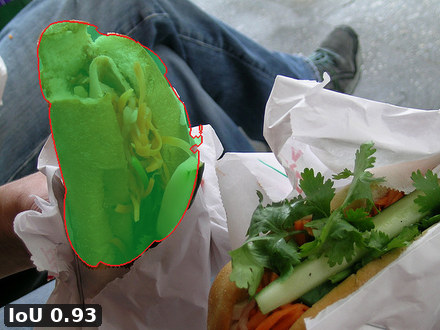} \\
      \qlab{RefCOCO+}{her} & \qim{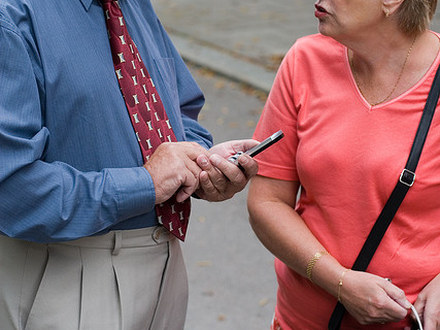} & \qim{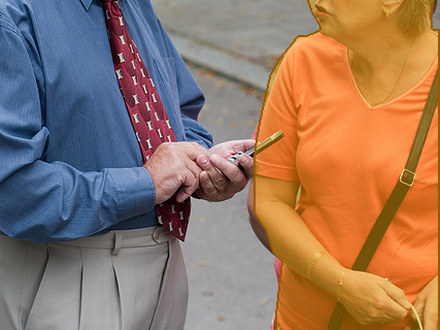} & \qim{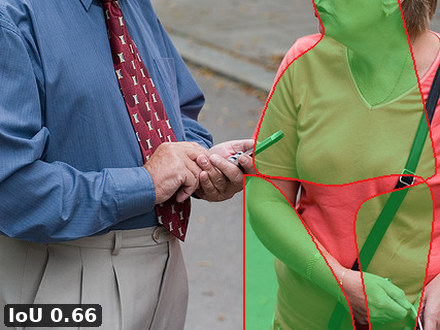} & \qim{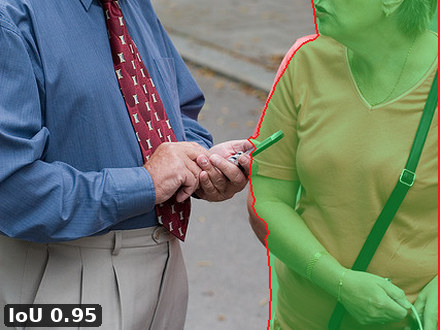} \\
      \qlab{ADE20K}{arcade machine} & \qim{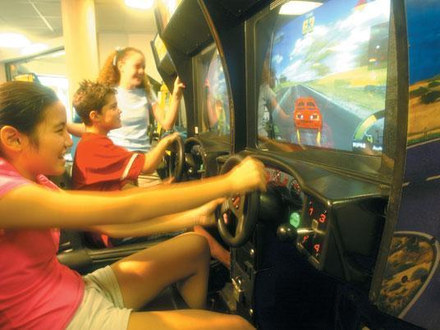} & \qim{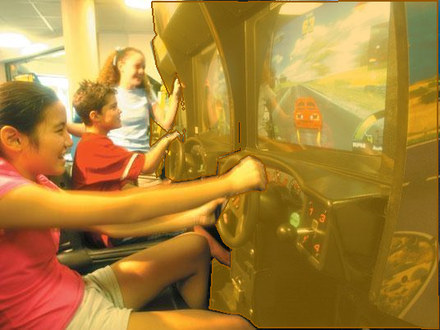} & \qim{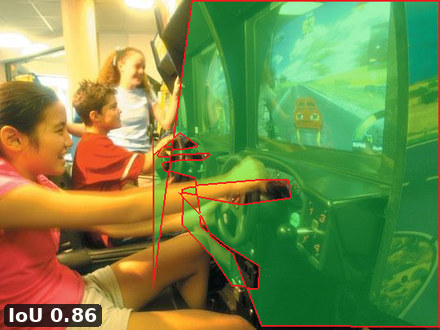} & \qim{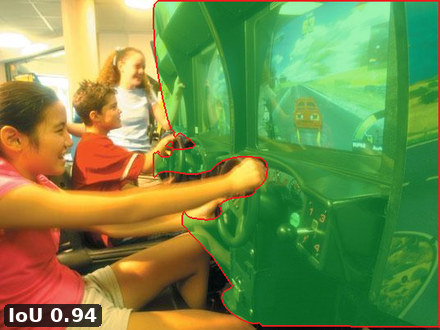} \\
    \end{tabular}
    \vspace{-6pt}
    \caption{\small \textbf{\ours across benchmarks and task formats}. The same
    \gfl backend and settings are used throughout. The prompt appears at left. Ground truth
    is shaded amber, and predictions are shown in green with a red outline. Each
    chip reports the panel's IoU. Column~3 shows the stage-2 mask, and column~4
    shows the final mask. Together, they isolate the effect of refinement.}
    \label{fig:qual_bench}
  \end{figure}

\begin{figure}[t]
    \centering
    \setlength{\tabcolsep}{1pt}
    \setlength{\qgw}{0.244\linewidth}
    \renewcommand{\arraystretch}{1.1}
    \begin{tabular}{@{}cccc@{}}
      \qhd{Input} & \qhd{Ground truth} & \qhd{Initial trace} & \qhd{\ours} \\[1pt]
      \qim{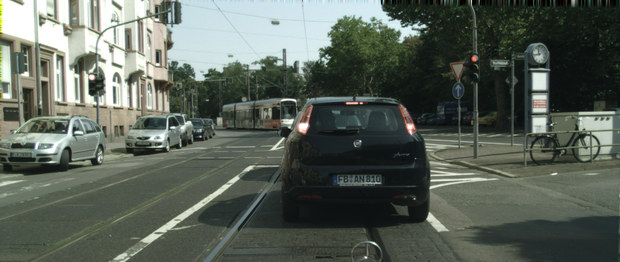} & \qim{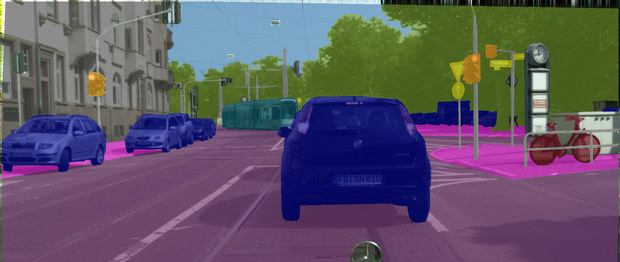} & \qim{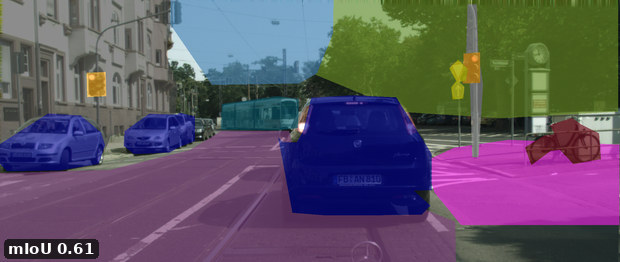} & \qim{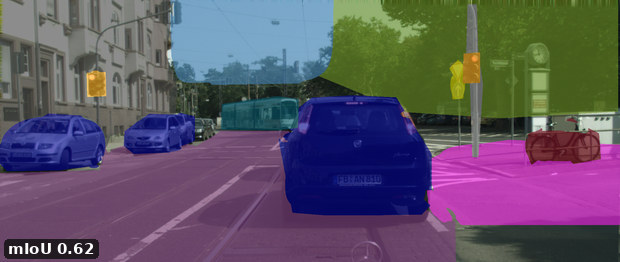} \\
      \qim{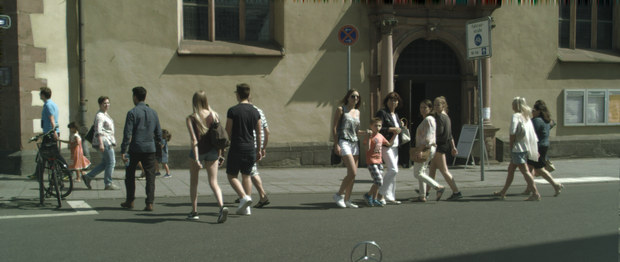} & \qim{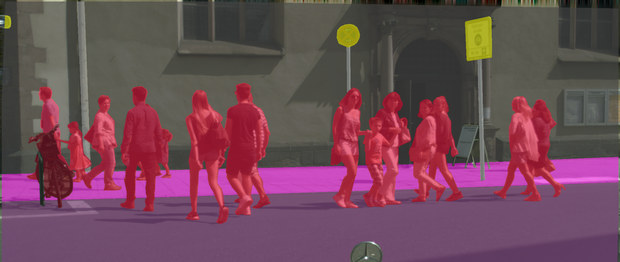} & \qim{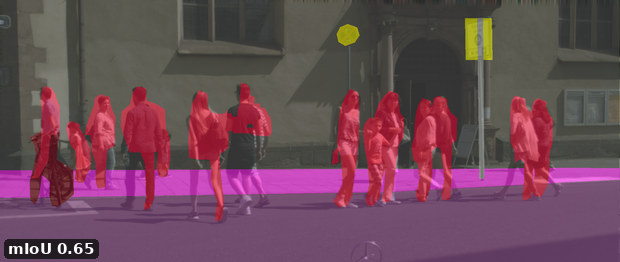} & \qim{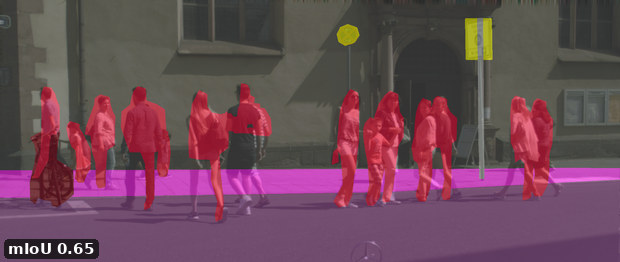} \\
      \qim{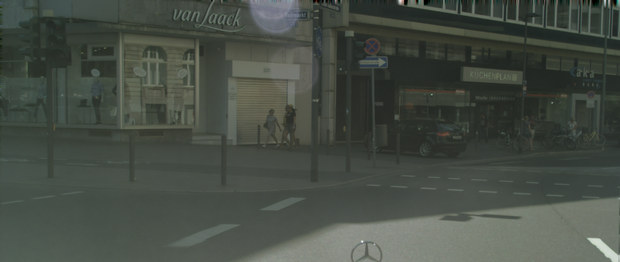} & \qim{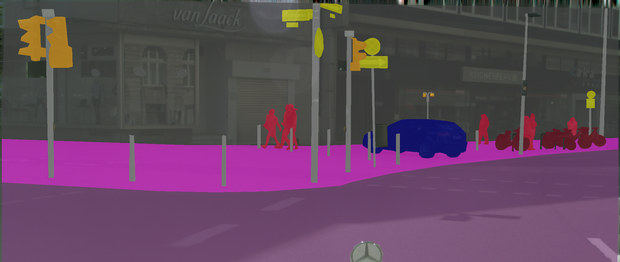} & \qim{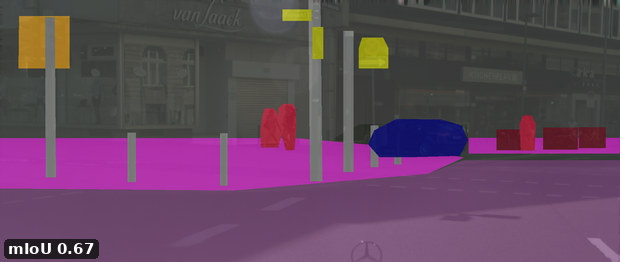} & \qim{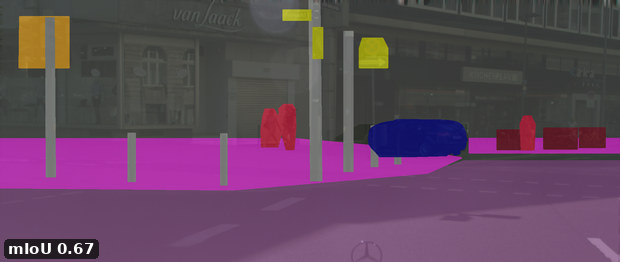} \\
      \qim{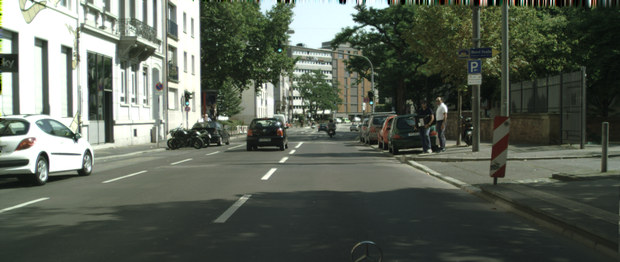} & \qim{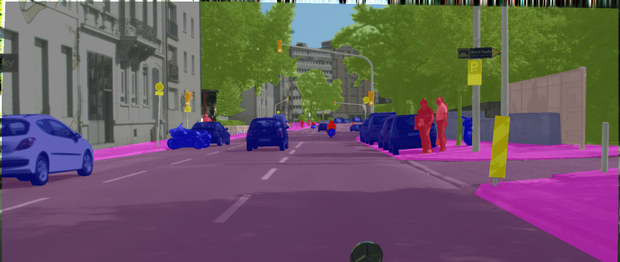} & \qim{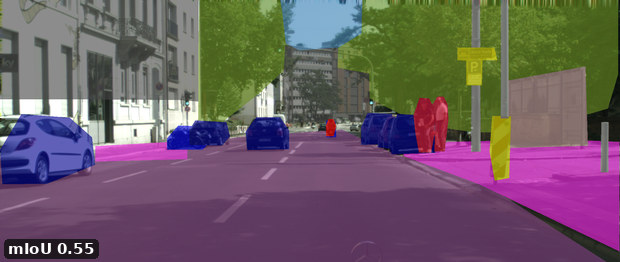} & \qim{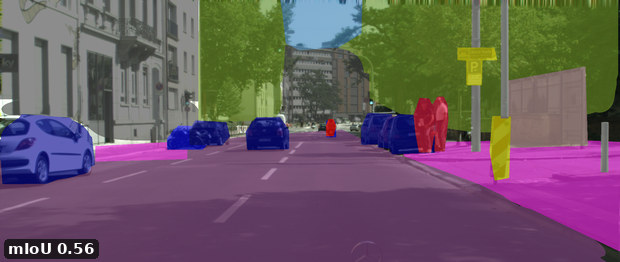} \\
      \qim{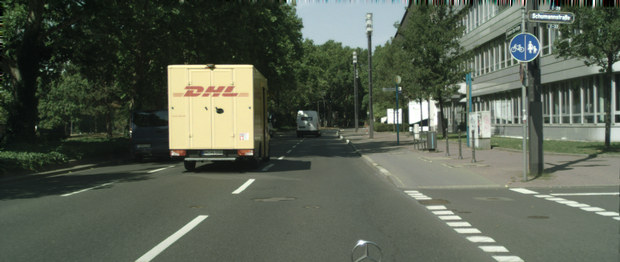} & \qim{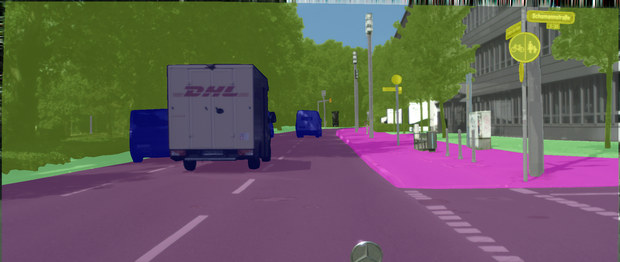} & \qim{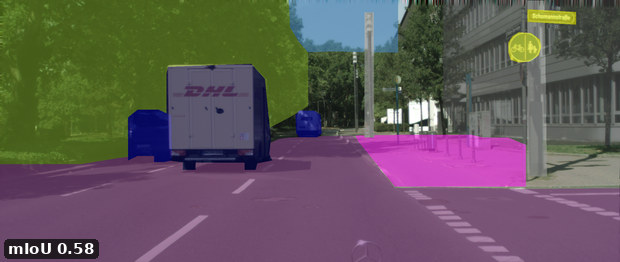} & \qim{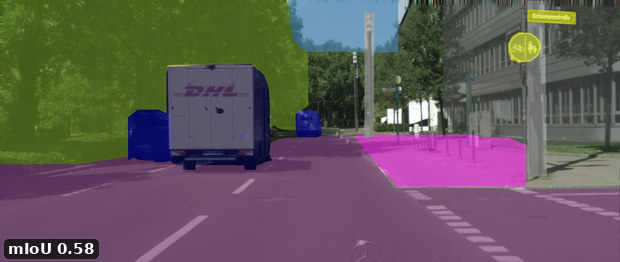} \\
    \end{tabular}
    \vspace{-6pt}
    \caption{\small \textbf{Dense semantic labeling on Cityscapes}
    \citep{cordts2016cityscapes}, \gfl backend, official 19-class palette. \ours
    labels every class in an image from a single query and resolves overlaps by
    deterministic mask compositing. Each panel is therefore a full label map,
    rather than a single mask, and its chip reports the image's mIoU. The losses
    occur on amorphous ground classes, which lack clear visual boundaries.}
    \label{fig:qual_semantic}
  \end{figure}

\begin{figure}[p]
    \centering
    \setlength{\tabcolsep}{1pt}
    \setlength{\qgw}{0.142\linewidth}
    \renewcommand{\arraystretch}{1.0}
    \begin{tabular}{@{}m{0.21\linewidth}@{\hspace{3pt}}ccccc@{}}
      & \qhd{Input} & \qhd{Ground truth} & \qhd{\samthree} & \qhd{\samagent} & \qhd{\ours} \\[1pt]
      \qlab{ReasonSeg}{To keep bread fresh and protected, it is often placed in\ldots} & \qim{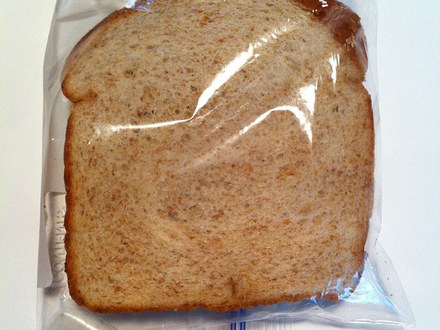} & \qim{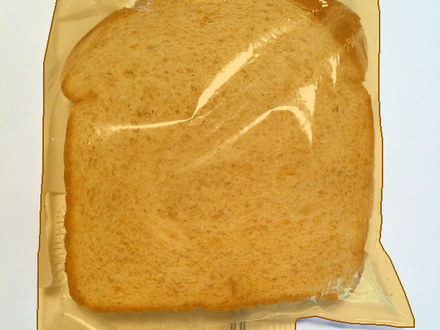} & \qim{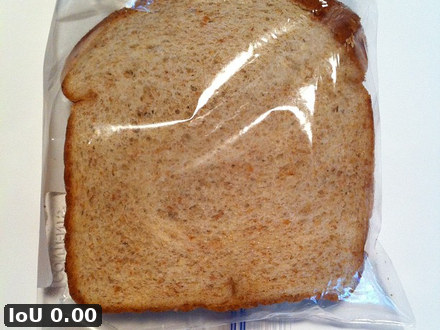} & \qim{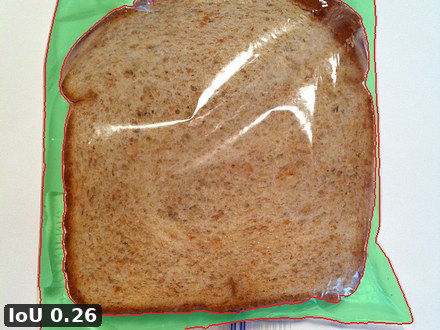} & \qim{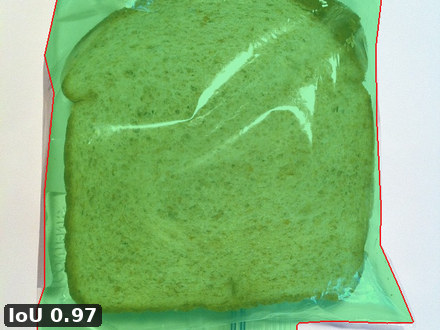} \\
      \qlab{ReasonSeg}{something that can control the fan speed} & \qim{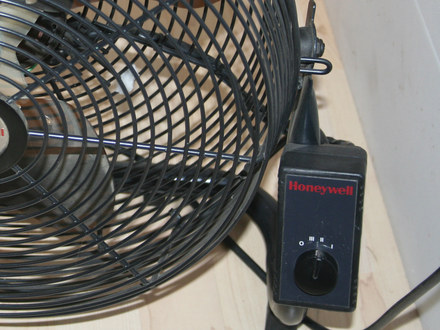} & \qim{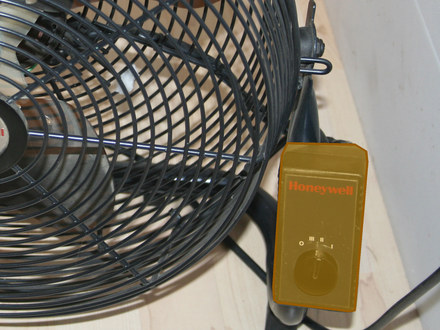} & \qim{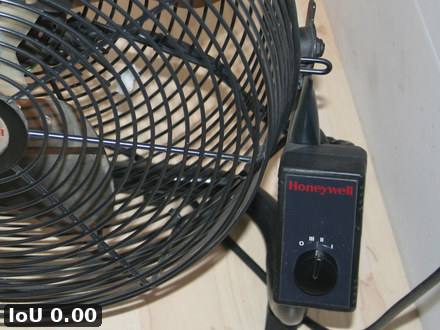} & \qim{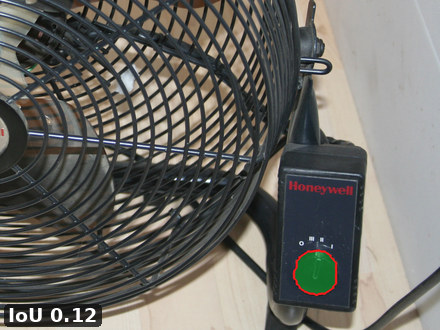} & \qim{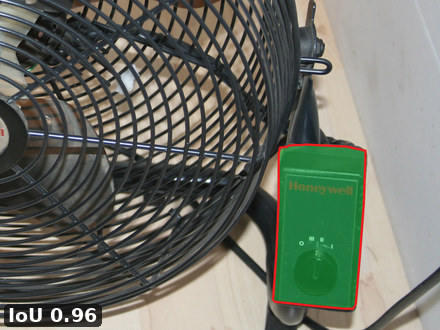} \\
      \qlab{ReasonSeg}{the person who is most likely to be the girl's trainer} & \qim{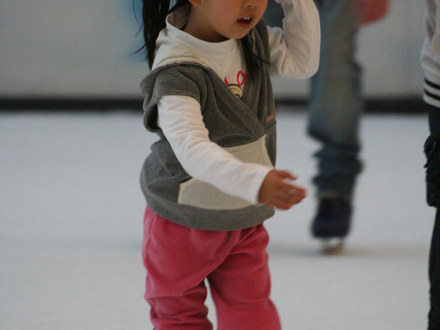} & \qim{qual_compare/reasonseg_trainer_gt.jpg} & \qim{qual_compare/reasonseg_trainer_sam3.jpg} & \qim{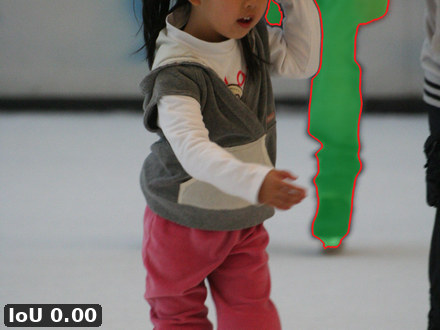} & \qim{qual_compare/reasonseg_trainer_ours.jpg} \\
      \qlab{RefCOCOg}{the gray keyboard with the touch pad} & \qim{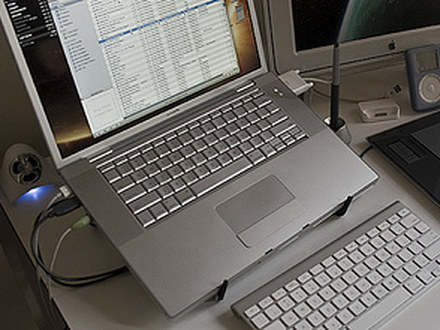} & \qim{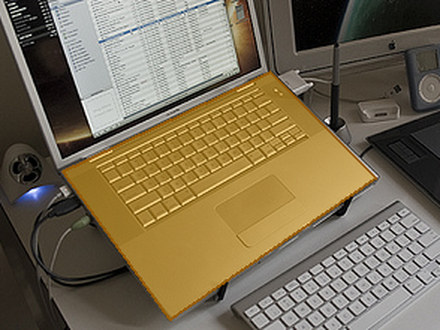} & \qim{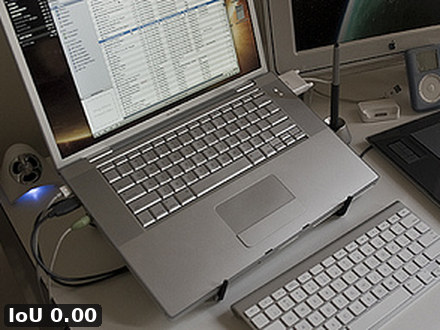} & \qim{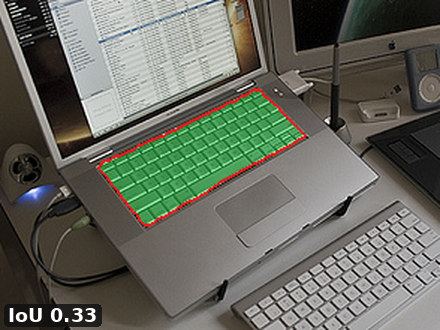} & \qim{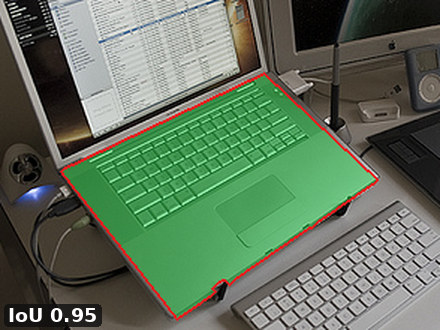} \\
      \qlab{RefCOCOg}{bowl on right} & \qim{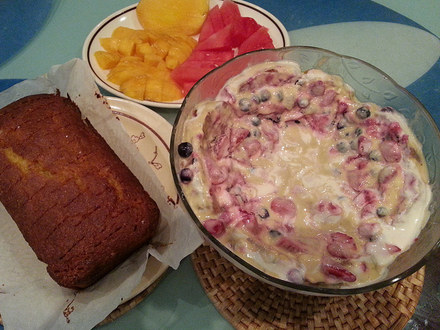} & \qim{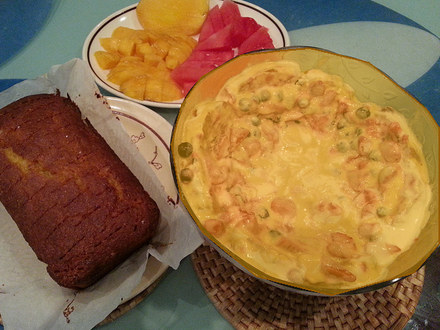} & \qim{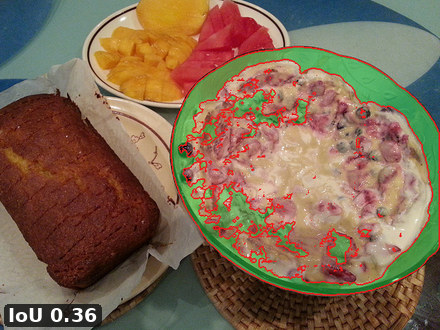} & \qim{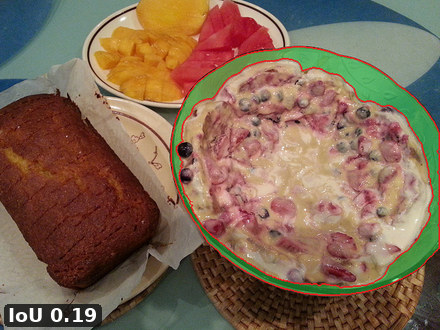} & \qim{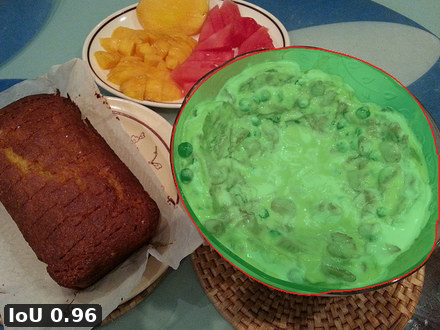} \\
      \qlab{RefCOCO}{gurl} & \qim{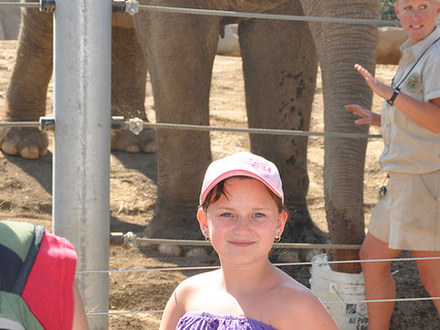} & \qim{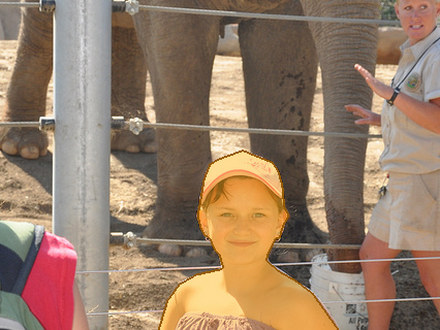} & \qim{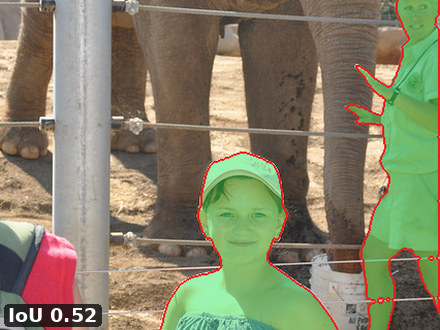} & \qim{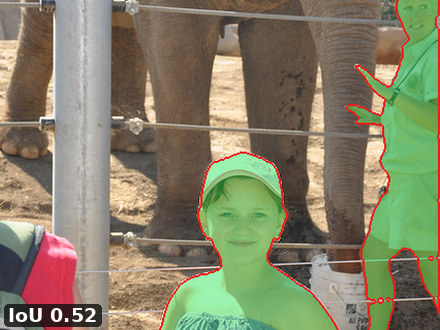} & \qim{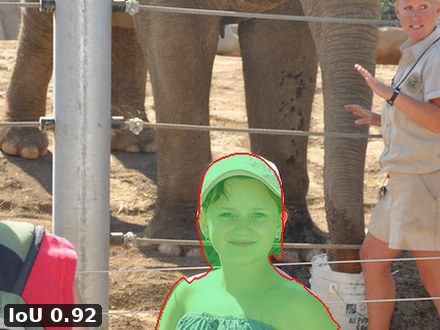} \\
      \qlab{RefCOCO+}{toasted darker larger half} & \qim{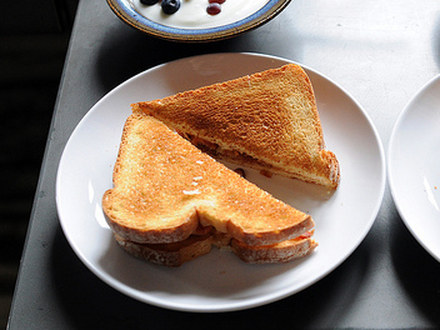} & \qim{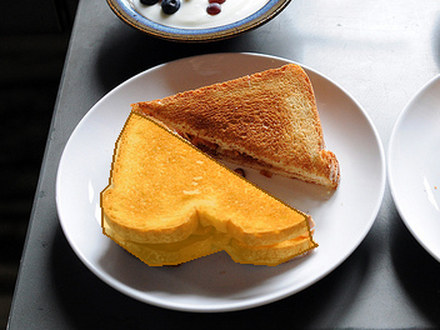} & \qim{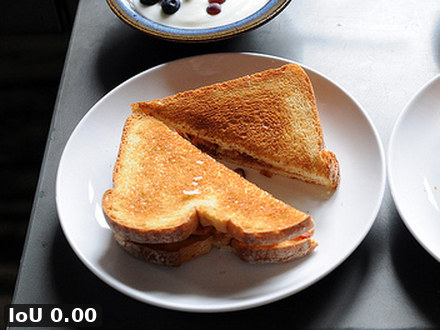} & \qim{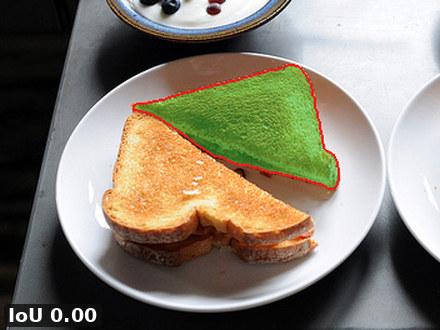} & \qim{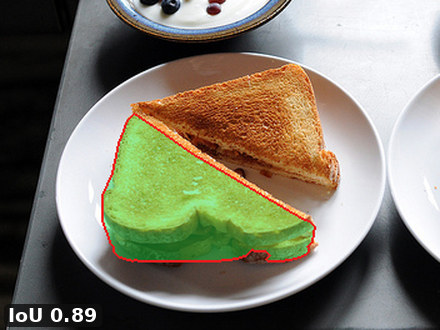} \\
      \qlab{SA-Co/Gold}{fixed ladder} & \qim{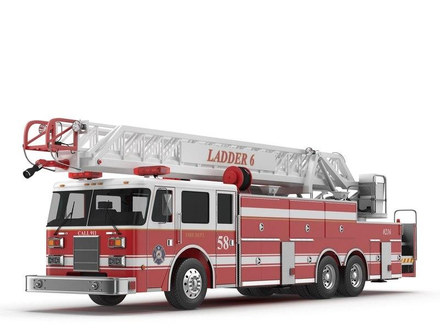} & \qim{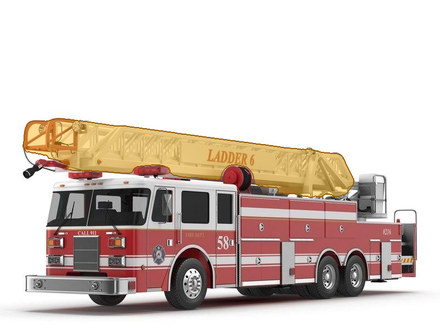} & \qim{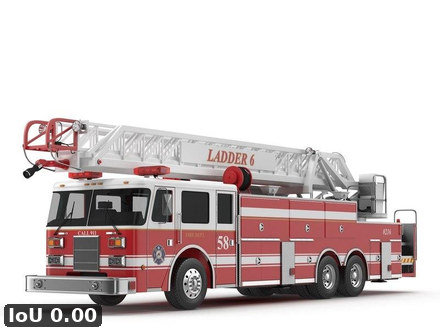} & \qim{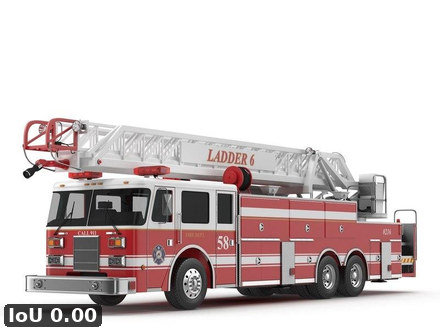} & \qim{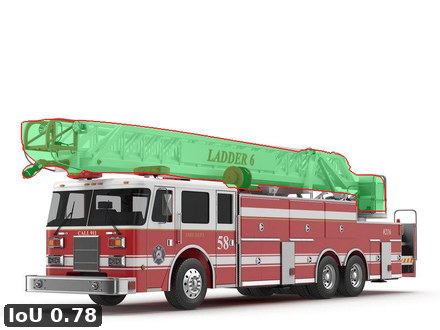} \\
      \qlab{ADE20K}{grandstand} & \qim{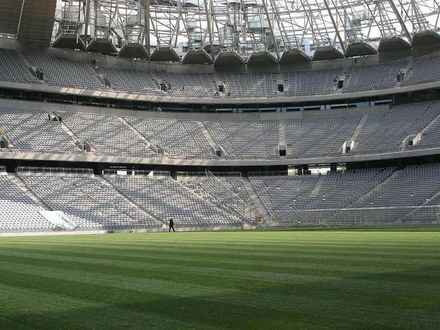} & \qim{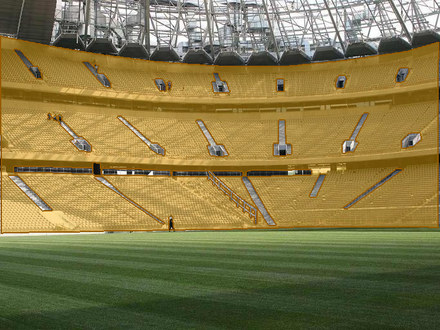} & \qim{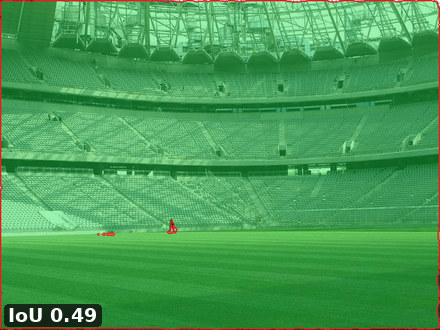} & \qim{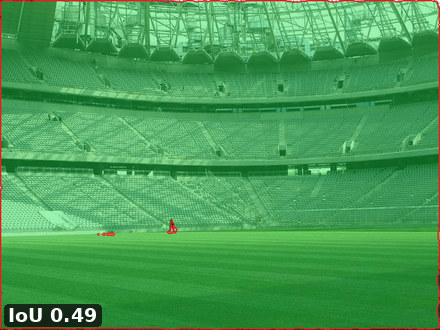} & \qim{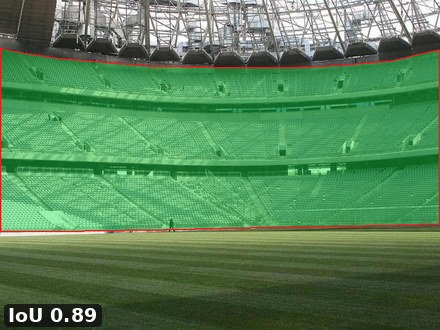} \\
    \end{tabular}
    \vspace{-6pt}
    \caption{\small \textbf{Where \ours separates from both \samthree baselines.}
    Selected items from ReasonSeg, RefCOCO/+/g, SA-Co/Gold, and ADE20K.
    \samthree \citep{carion2025sam} receives the same text query as \ours, and
    each chip reports that system's recorded IoU.}
    \label{fig:qual_compare}
  \end{figure}

  The grids below extend the qualitative evidence of \secLabel\ref{sec:results}.
  They use the same \gfl runs as the tables, and each chip reports the IoU recorded
  by the harness for that panel. \figLabel\ref{fig:qual_bench} isolates the effect
  of refinement across benchmark families. \figLabel\ref{fig:qual_semantic} shows
  full Cityscapes label maps. \figLabel\ref{fig:qual_compare} compares \ours
  with both \samthree baselines.
  
  \subsection{Crowded scenes, negatives, rare domains, and single passes} \label{app:qual_extra}
  \vspace{-2pt}
  
\begin{figure}[p]
  \centering
  \setlength{\tabcolsep}{1pt}
  \setlength{\qgw}{0.129\linewidth}
  \renewcommand{\arraystretch}{1.0}
  \begin{tabular}{@{}m{0.21\linewidth}@{\hspace{3pt}}ccccc@{}}
    & \qhd{Input} & \qhd{Ground truth} & \qhd{\samthree} & \qhd{\samagent} & \qhd{\ours} \\[1pt]
    \qlab{COCO, 3 objects}{bird} & \qim{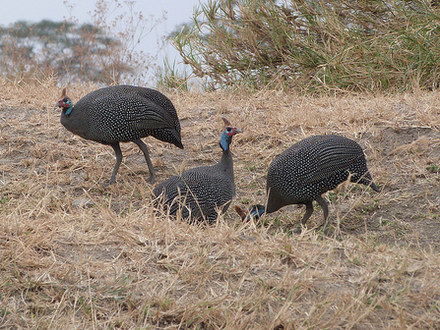} & \qim{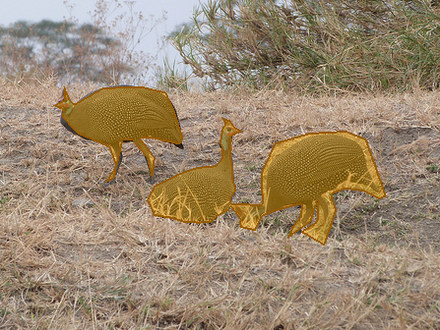} & \qim{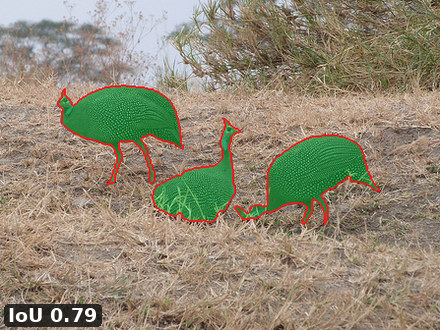} & \qim{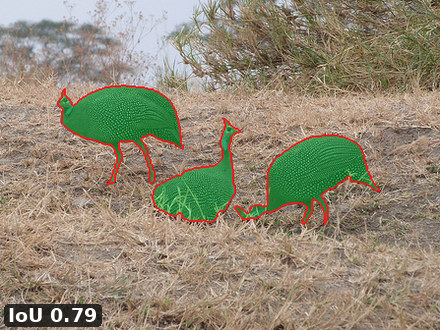} & \qim{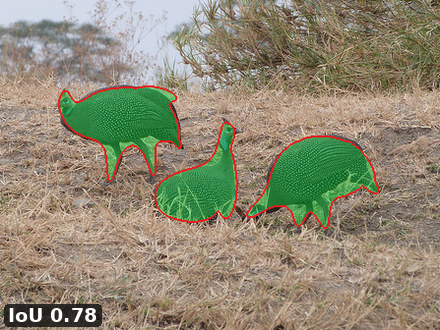} \\
    \qlab{LVIS, 3 objects}{pie} & \qim{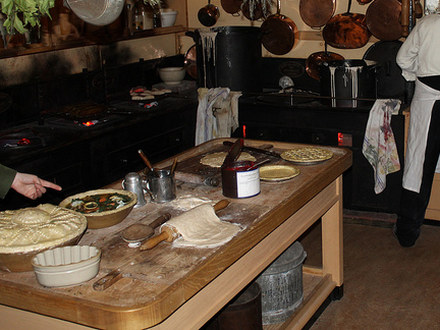} & \qim{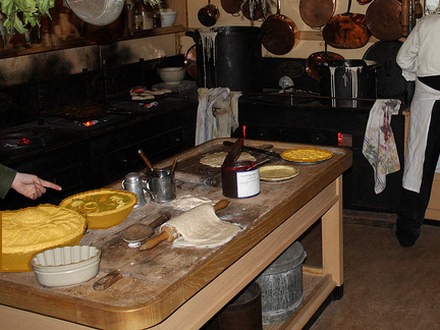} & \qim{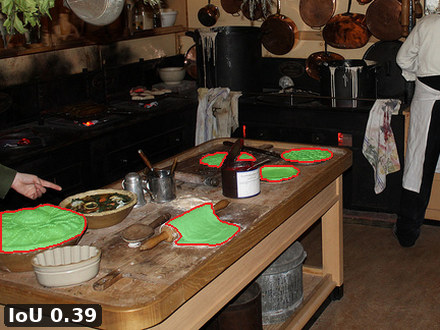} & \qim{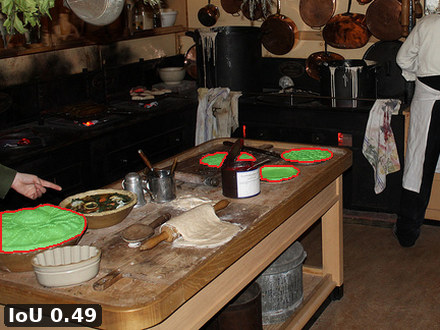} & \qim{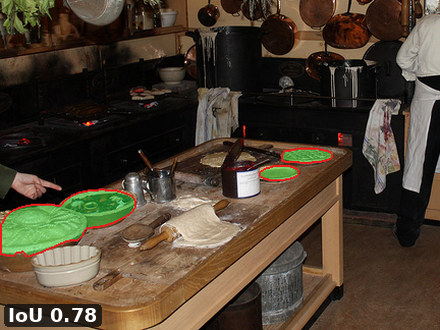} \\
    \qlab{LVIS, 4 objects}{orange (fruit)} & \qim{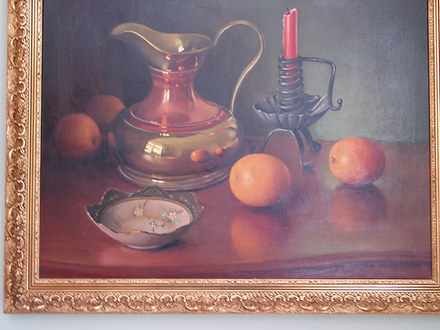} & \qim{qual_supp/multi_orange_gt.jpg} & \qim{qual_supp/multi_orange_sam3.jpg} & \qim{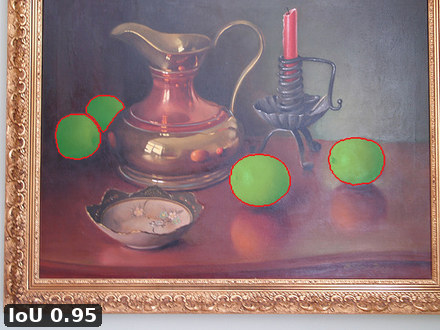} & \qim{qual_supp/multi_orange_ours.jpg} \\
    \qlab{SA-Co/Gold, 4 objects}{metal lattice} & \qim{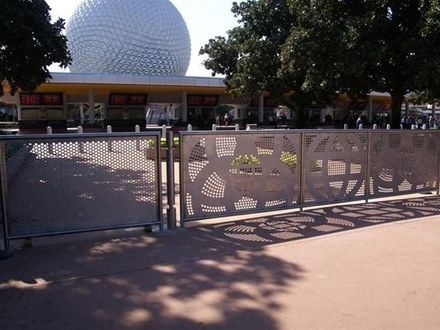} & \qim{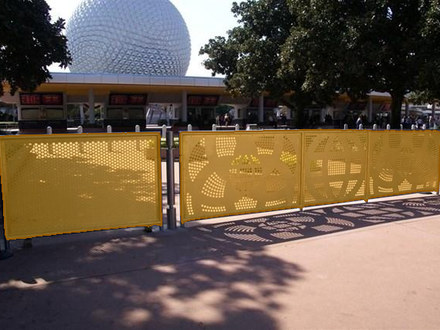} & \qim{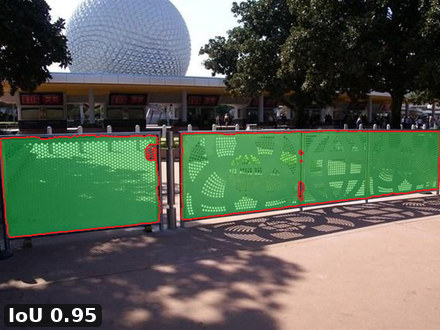} & \qim{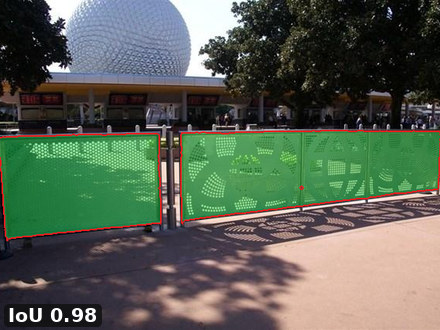} & \qim{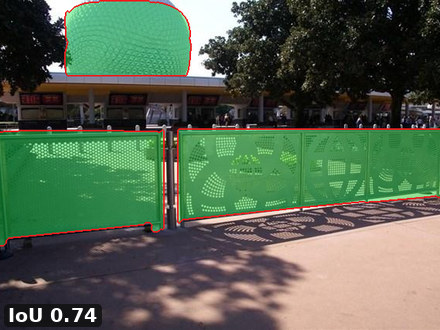} \\
    \qlab{COCO, 5 objects}{car} & \qim{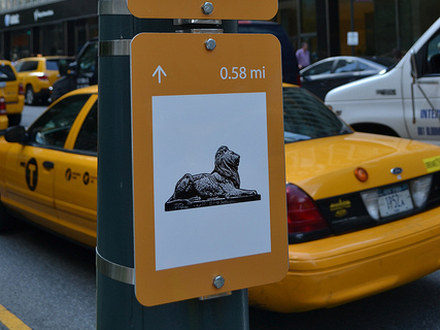} & \qim{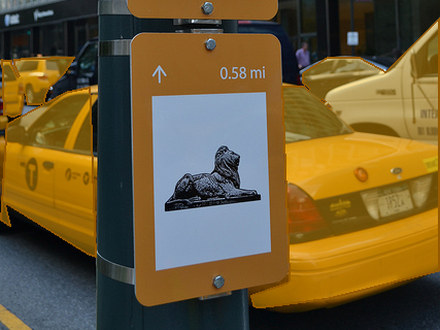} & \qim{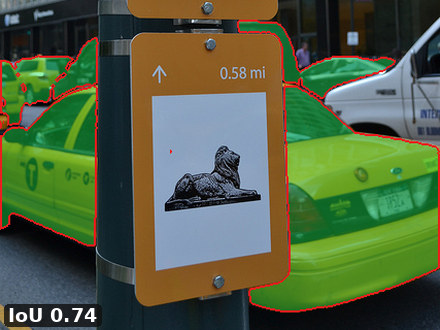} & \qim{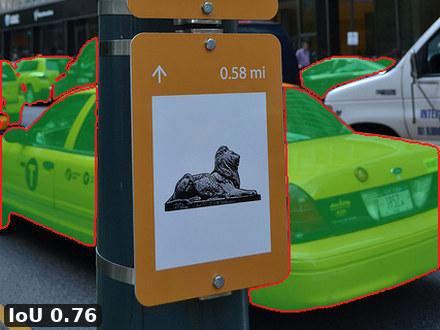} & \qim{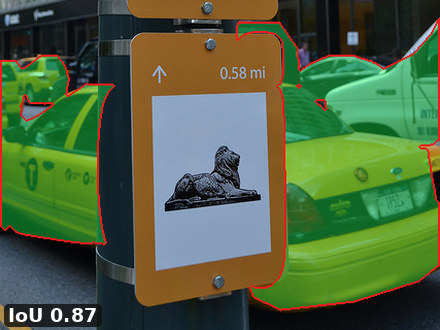} \\
    \qlab{COCO, 8 objects}{person} & \qim{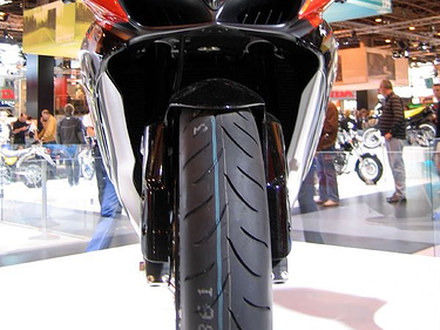} & \qim{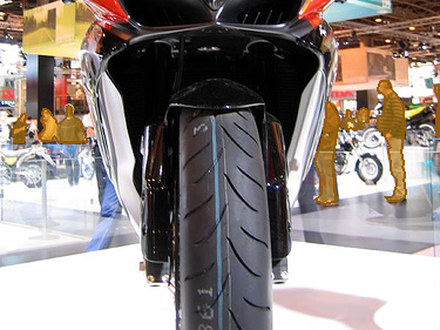} & \qim{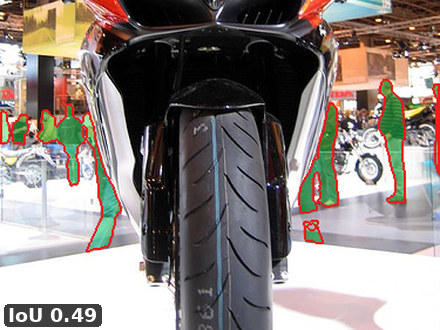} & \qim{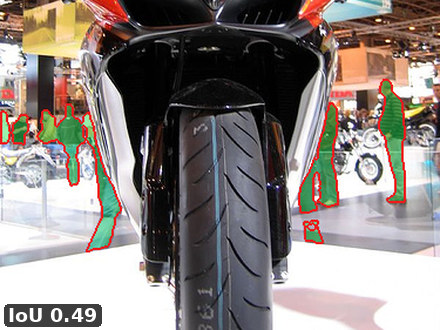} & \qim{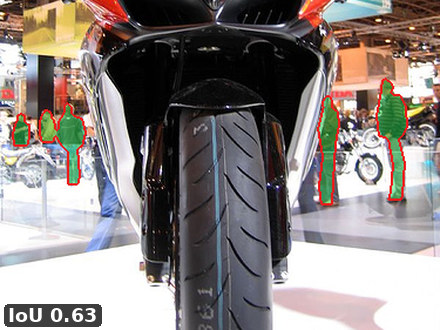} \\
    \qlab{COCO, 10 objects}{toilet} & \qim{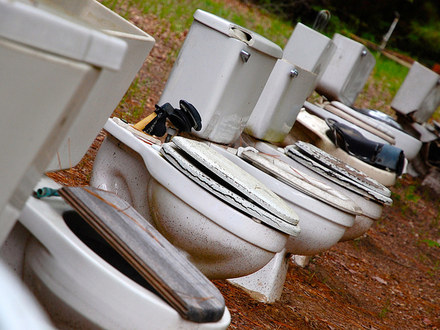} & \qim{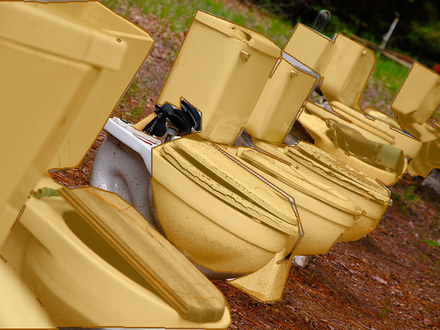} & \qim{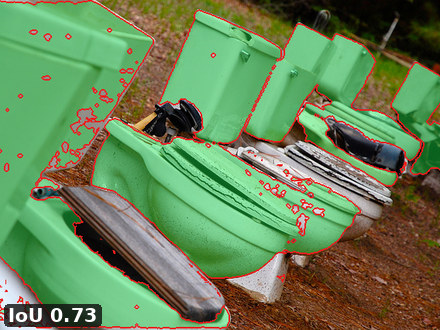} & \qim{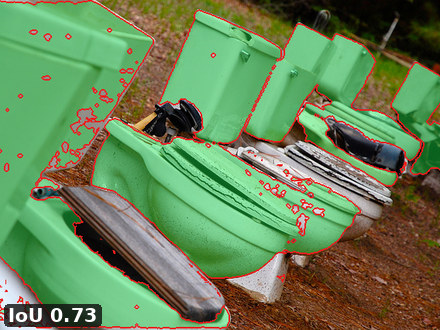} & \qim{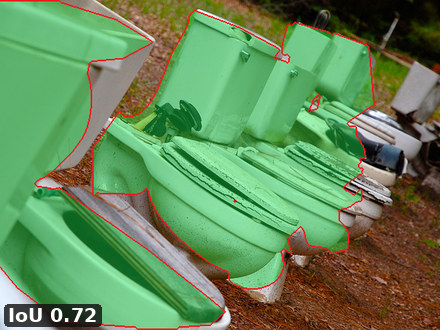} \\
    \qlab{COCO, 13 objects}{sheep} & \qim{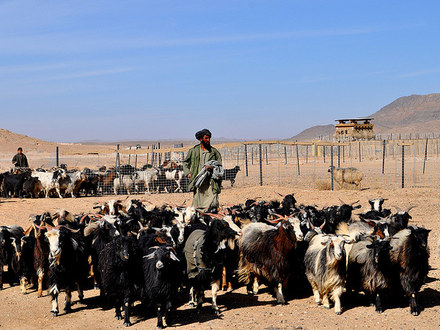} & \qim{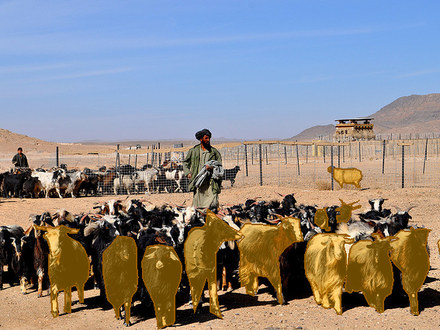} & \qim{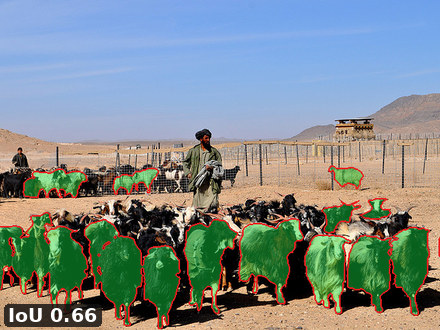} & \qim{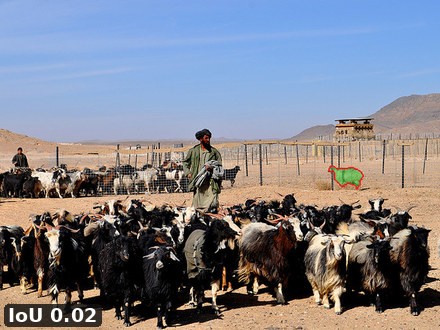} & \qim{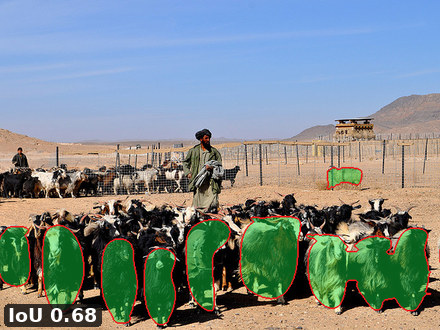} \\
    \qlab{LVIS, 47 objects}{book} & \qim{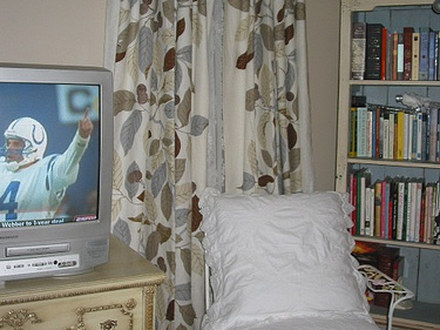} & \qim{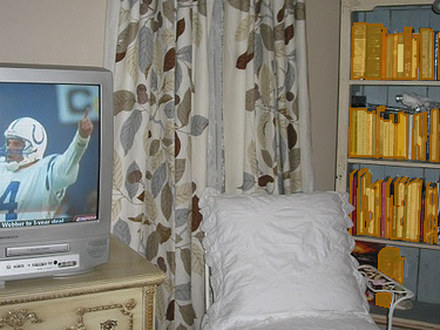} & \qim{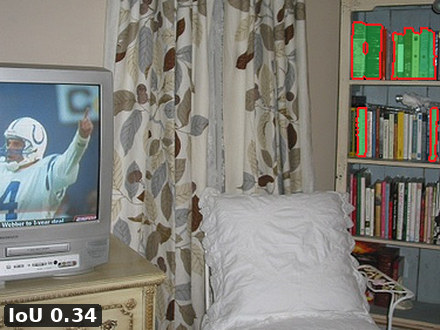} & \qim{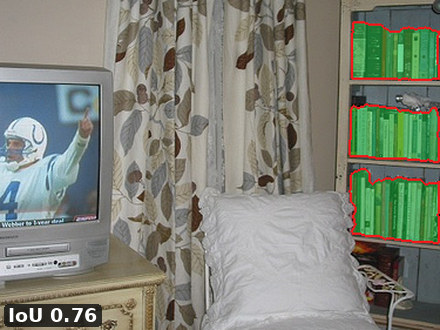} & \qim{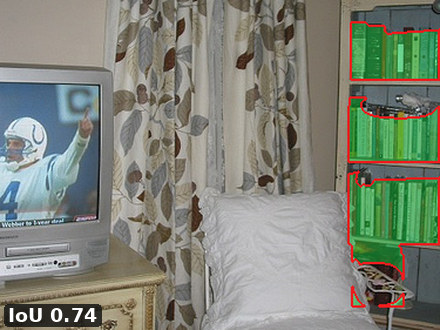} \\
  \end{tabular}
  \vspace{-6pt}
  \caption{\small \textbf{Images containing multiple objects of the queried class.}
  LVIS, COCO, and SA-Co/Gold examples whose ground truth contains at least three
  objects. Each label gives the benchmark and number of ground-truth objects. Each chip reports the
  corresponding system's recorded union IoU, which does not separate touching
  objects that a system merges into one region.}
  \label{fig:supp_multi}
\end{figure}

\begin{figure}[t]
  \centering
  \setlength{\tabcolsep}{1pt}
  \setlength{\qgw}{0.185\linewidth}
  \renewcommand{\arraystretch}{1.0}
  \begin{tabular}{@{}m{0.22\linewidth}@{\hspace{3pt}}cccc@{}}
    & \qhd{Input} & \qhd{\samthree} & \qhd{\samagent} & \qhd{\ours} \\[1pt]
    \qlab{SA-Co/Gold}{inflatable canoe} & \qim{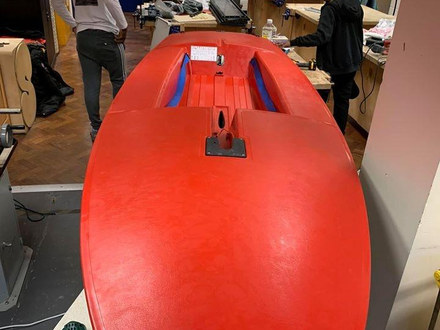} & \qim{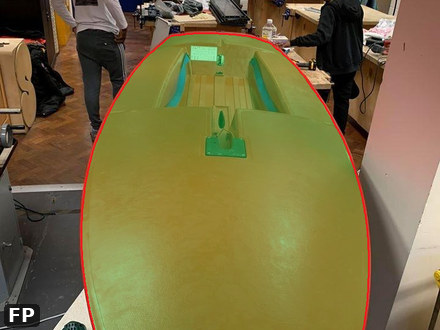} & \qim{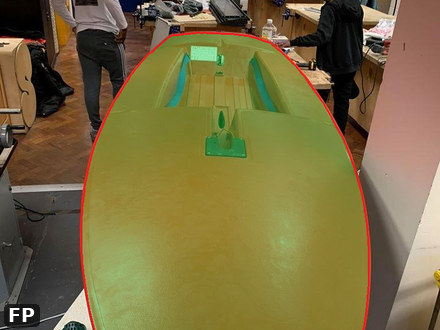} & \qim{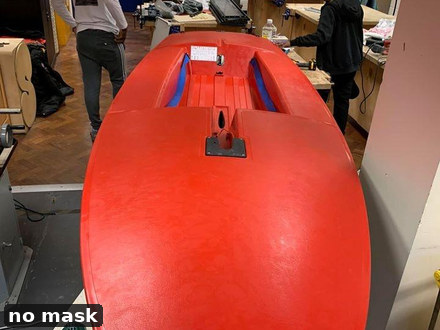} \\
    \qlab{SA-Co/Gold}{roller conveyor} & \qim{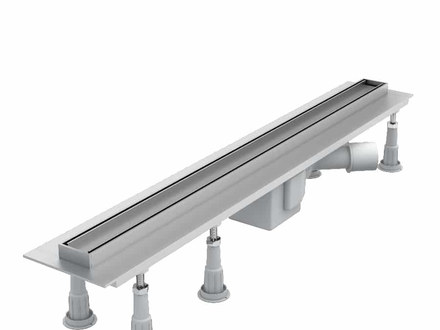} & \qim{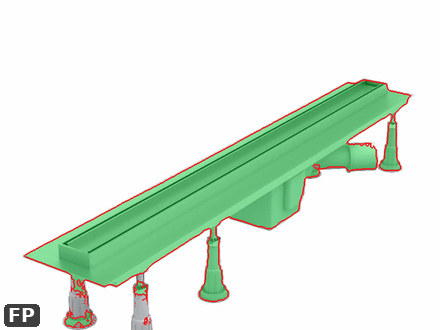} & \qim{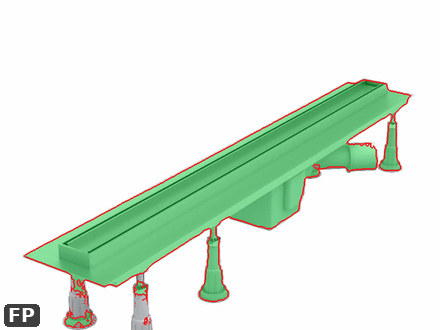} & \qim{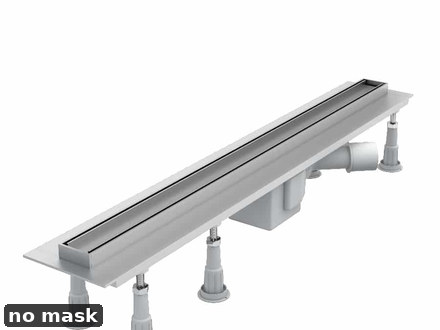} \\
    \qlab{SA-Co/Gold}{metal drum} & \qim{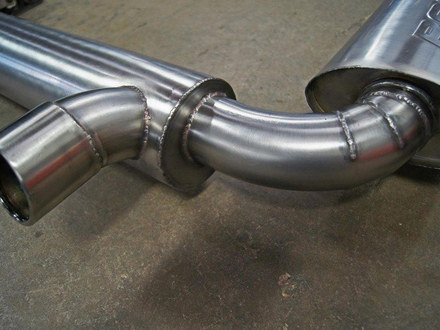} & \qim{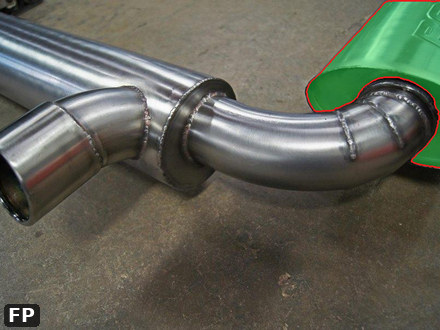} & \qim{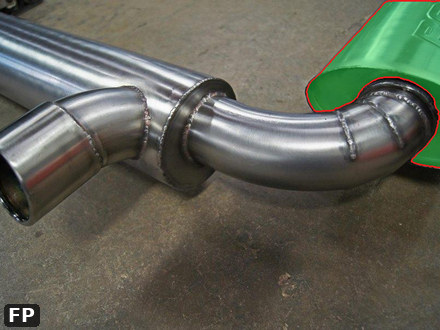} & \qim{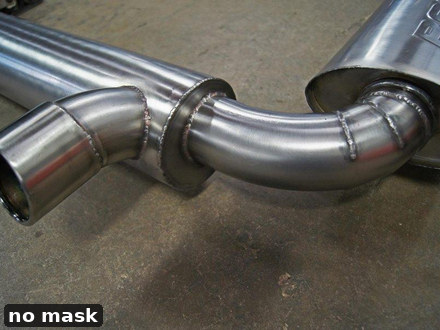} \\
    \qlab{SA-Co/Gold}{built-in bidet} & \qim{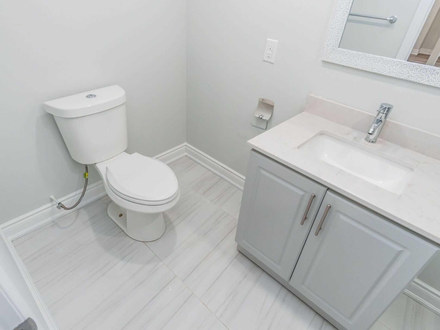} & \qim{qual_supp/neg_bidet_sam3.jpg} & \qim{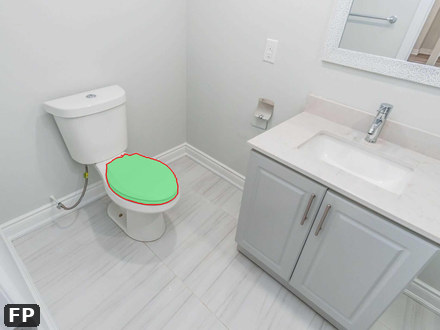} & \qim{qual_supp/neg_bidet_ours.jpg} \\
    \qlab{SA-Co/Gold}{striped shirt} & \qim{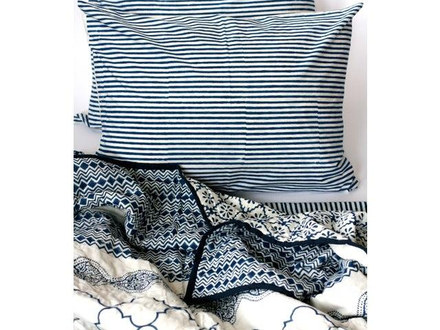} & \qim{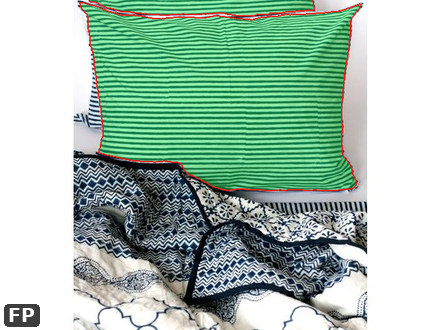} & \qim{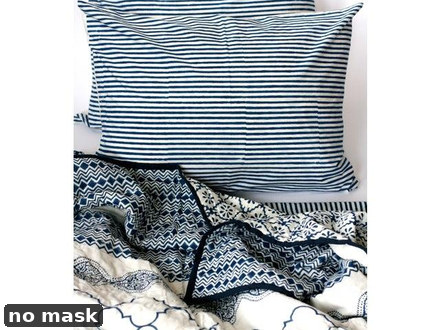} & \qim{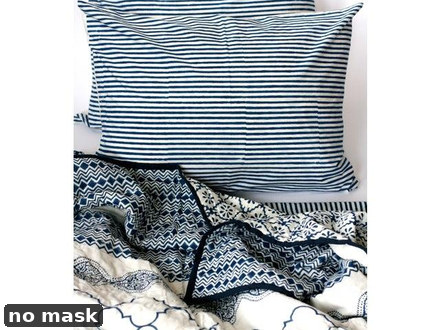} \\
    \qlab{SA-Co/Gold}{V-neck t-shirt} & \qim{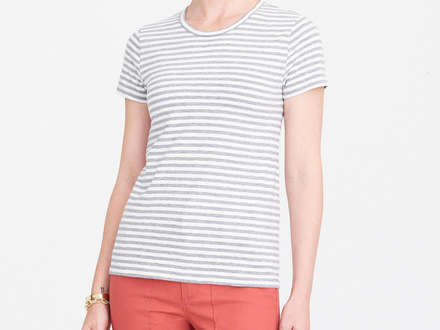} & \qim{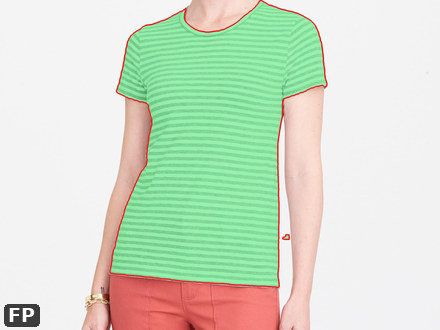} & \qim{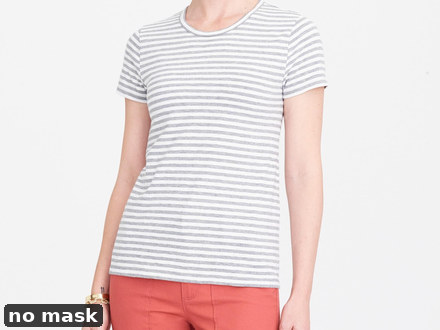} & \qim{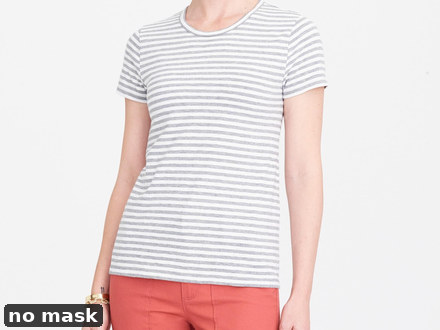} \\
  \end{tabular}
  \vspace{-6pt}
  \caption{\small \textbf{Hard negatives: the prompted concept is absent.} Six
  SA-Co/Gold examples whose phrases have no matching annotation in the image.
  Because the ground truth is empty, no ground-truth panel is shown. Each chip
  indicates only whether the system produced a mask. Each scene contains the
  broader \emph{category}, but not the qualified concept.}
  \label{fig:supp_neg}
\end{figure}

\begin{figure}[p]
  \centering
  \setlength{\tabcolsep}{1pt}
  \setlength{\qgw}{0.130\linewidth}
  \renewcommand{\arraystretch}{1.0}
  \begin{tabular}{@{}m{0.22\linewidth}@{\hspace{3pt}}cccc@{}}
    & \qhd{Input} & \qhd{Ground truth} & \qhd{\samthree} & \qhd{\ours} \\[1pt]
    \qlab{Documents: paper-parts}{figure} & \qim{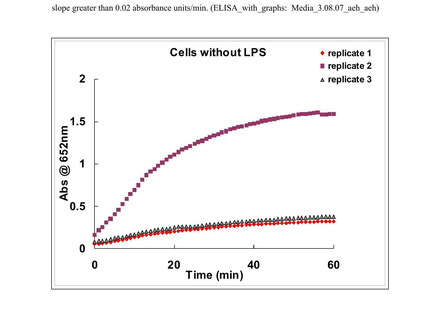} & \qim{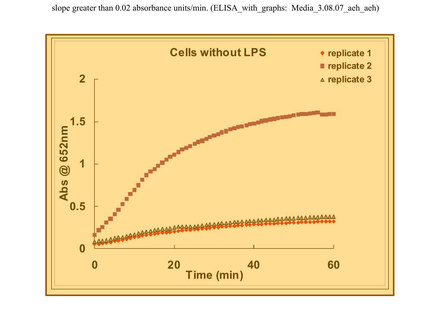} & \qim{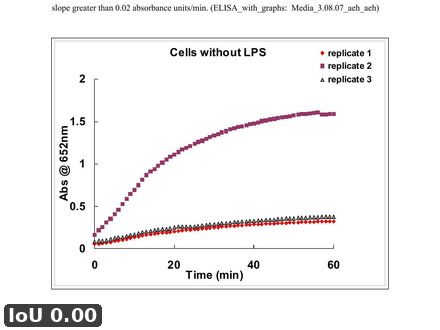} & \qim{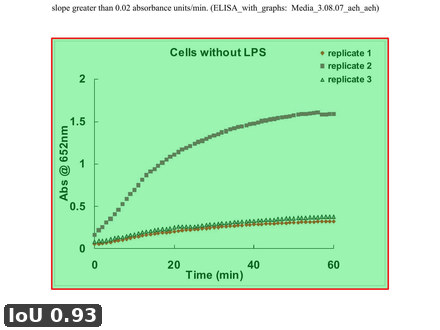} \\
    \qlab{Documents: invoice-processing}{table} & \qim{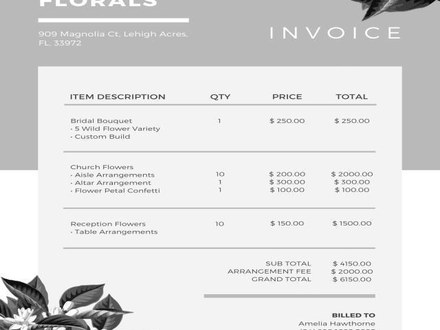} & \qim{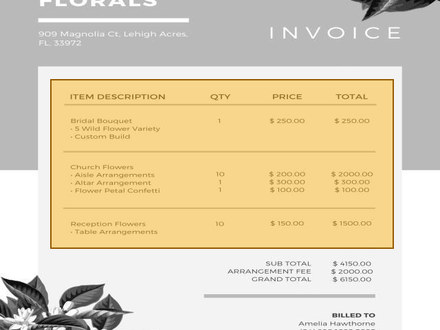} & \qim{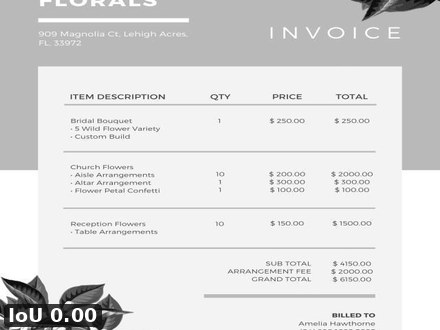} & \qim{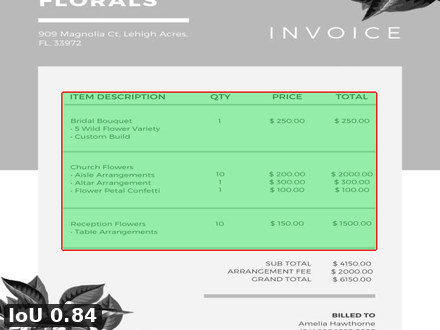} \\
    \qlab{Documents: activity-diagrams}{action} & \qim{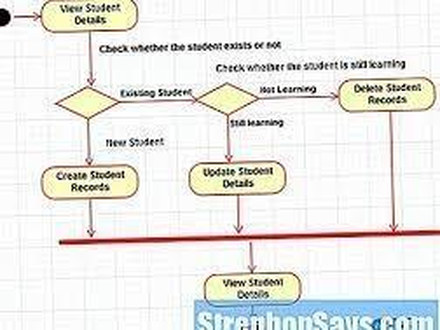} & \qim{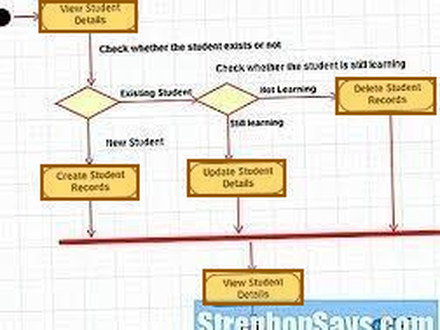} & \qim{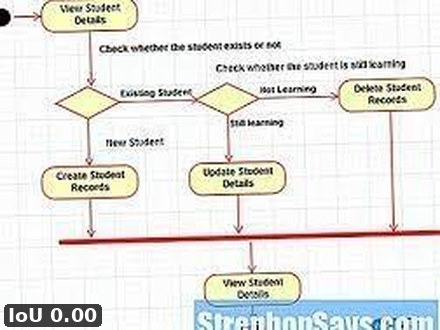} & \qim{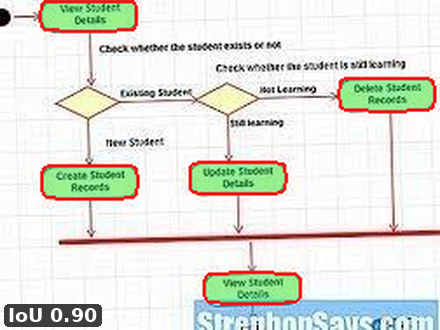} \\
    \qlab{Aerial: electric-pylon-detection}{pylon} & \qim{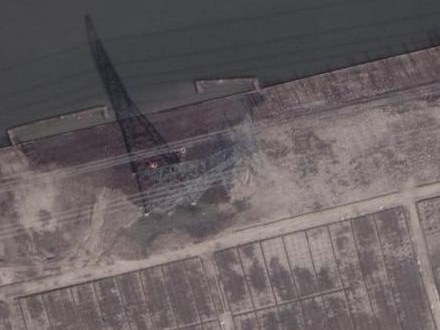} & \qim{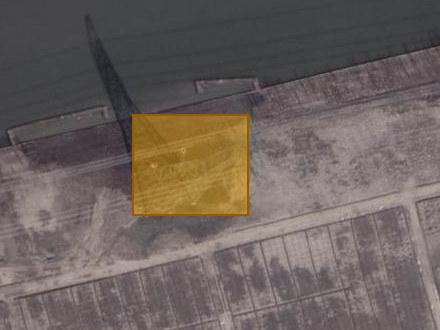} & \qim{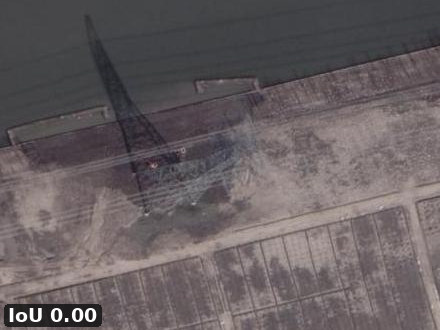} & \qim{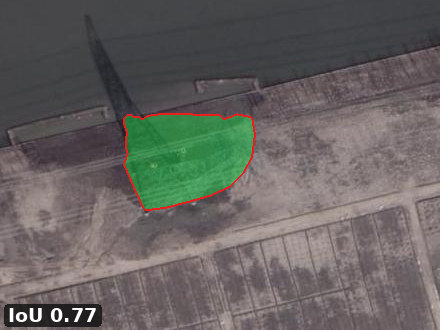} \\
    \qlab{Aerial: construction equipment}{dump truck} & \qim{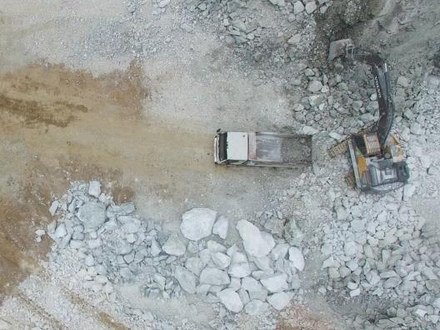} & \qim{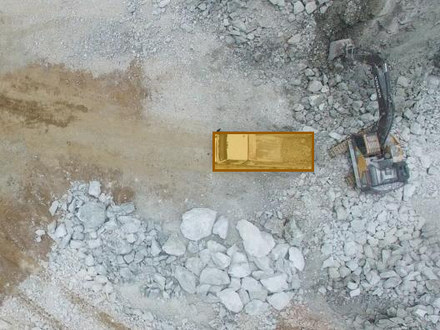} & \qim{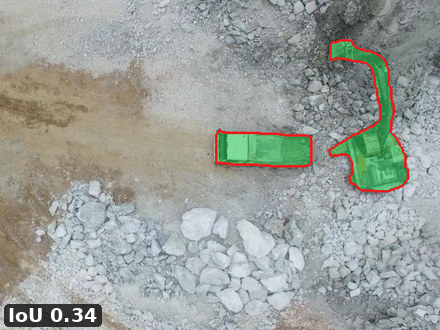} & \qim{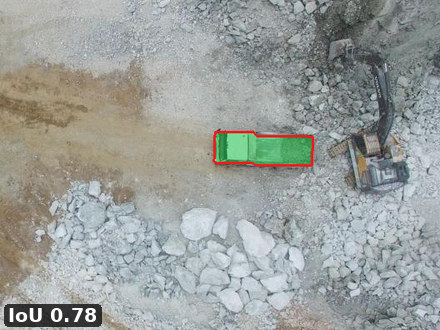} \\
    \qlab{Medical: brain-tumors-MRI}{pitiutary [sic]} & \qim{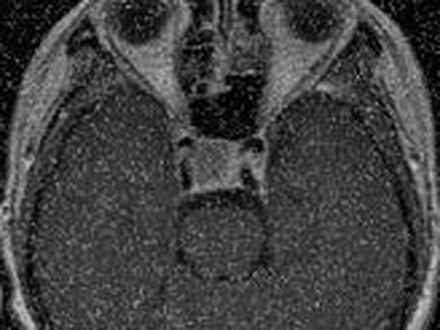} & \qim{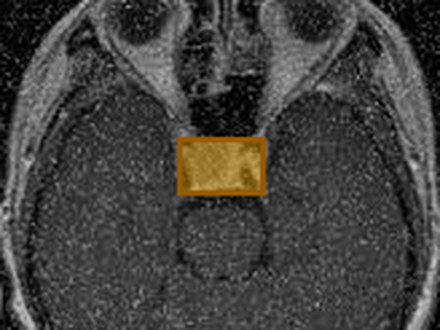} & \qim{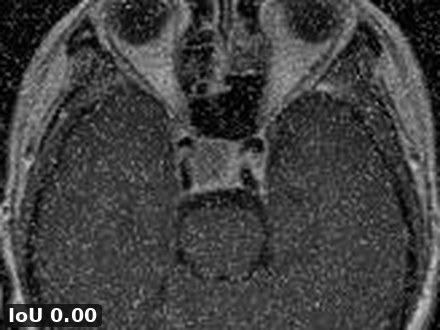} & \qim{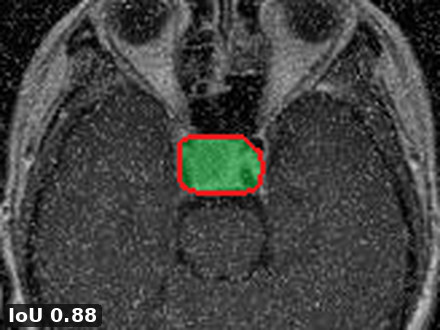} \\
    \qlab{Medical: nih-xray}{Pneumonia} & \qim{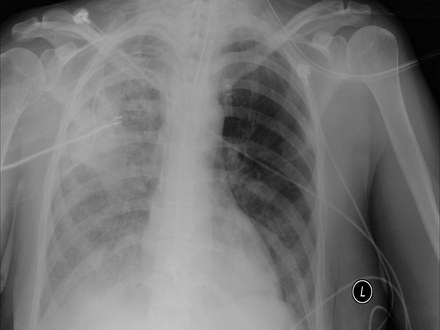} & \qim{qual_supp/dom_xray_gt.jpg} & \qim{qual_supp/dom_xray_sam3.jpg} & \qim{qual_supp/dom_xray_ours.jpg} \\
    \qlab{Medical: dentalai}{Impacted Tooth} & \qim{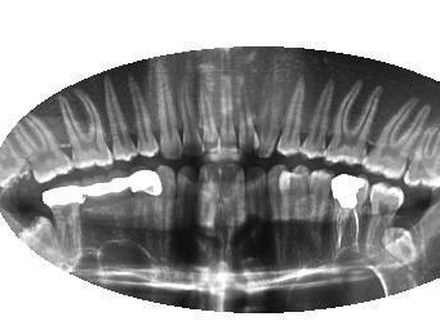} & \qim{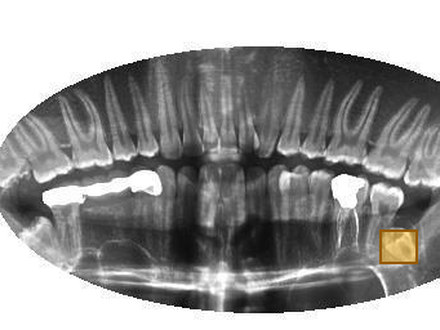} & \qim{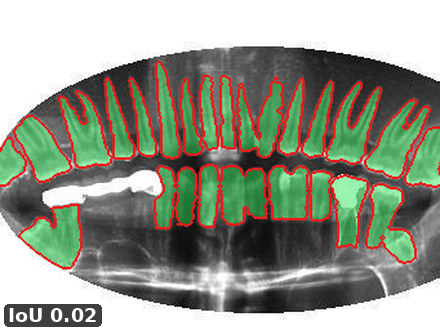} & \qim{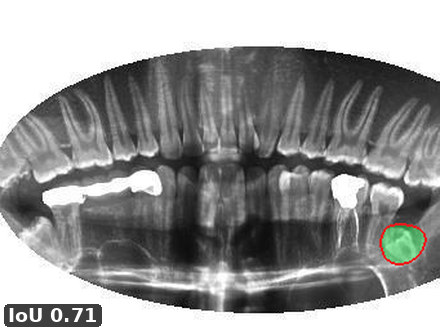} \\
    \qlab{Industrial: wheel-defect-detection}{Wheel} & \qim{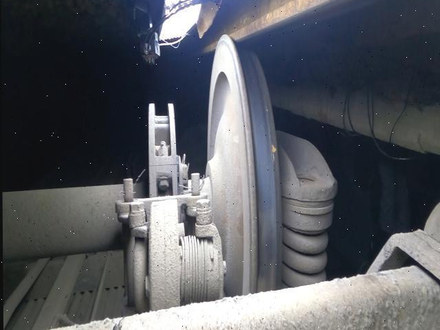} & \qim{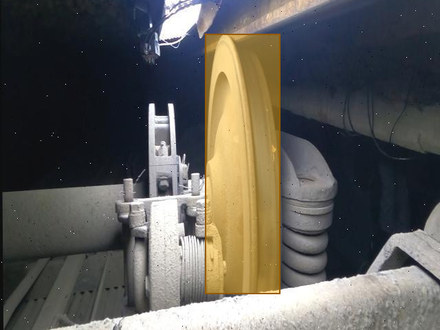} & \qim{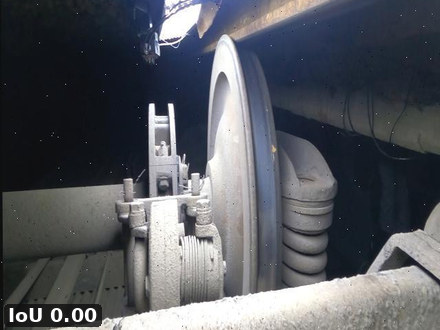} & \qim{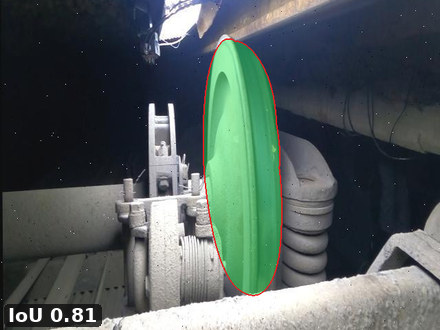} \\
  \end{tabular}
  \vspace{-6pt}
  \caption{\small \textbf{Rare domains segmentation on RF-100VL subsets.} The rows span documents, aerial imagery, medical imaging, and
  industrial inspection, with one subset per row. Each prompt uses that
  subset's category name verbatim, and each ground-truth panel shows the
  annotated detection box filled in. Some category names are domain jargon, so a failure here may reflect vocabulary rather than localization.}
  \label{fig:supp_domain}
\end{figure}

\begin{figure}[t]
  \centering
  \setlength{\tabcolsep}{0.5pt}
  \setlength{\qgw}{0.163\linewidth}
  \renewcommand{\arraystretch}{1.0}
  \begin{tabular}{@{}cccccc@{}}
    \qhd{Input} & \qhd{Ground truth} & \qhd{Stage 2} & \qhd{Pass 1} & \qhd{Pass 2} & \qhd{Last pass} \\[2pt]
    \multicolumn{6}{@{}p{\linewidth}@{}}{\bqry{man on left} \hfill {\scriptsize RefCOCOg, 12 passes}} \\[1pt]
    \qim{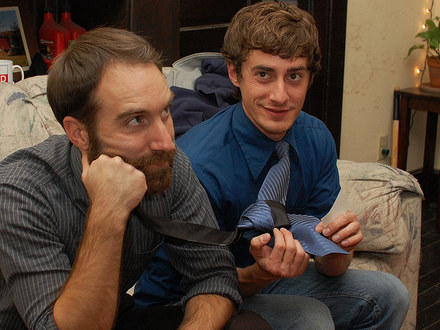} & \qim{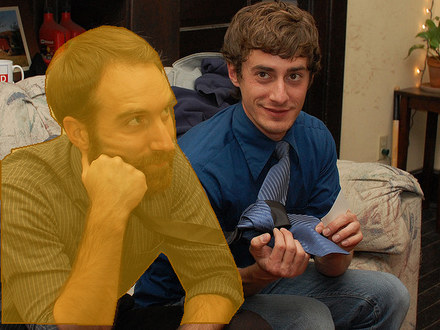} & \qim{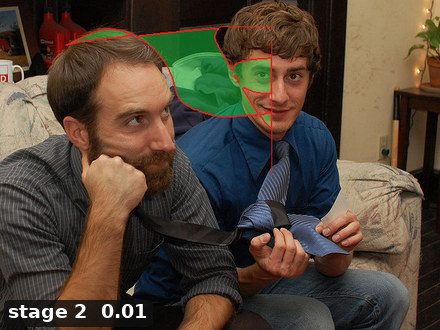} & \qim{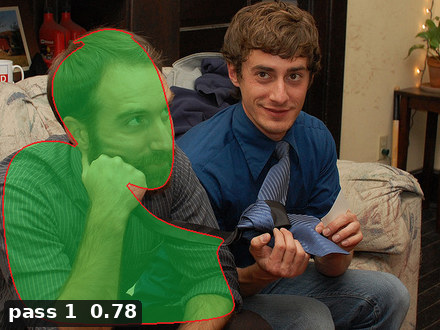} & \qim{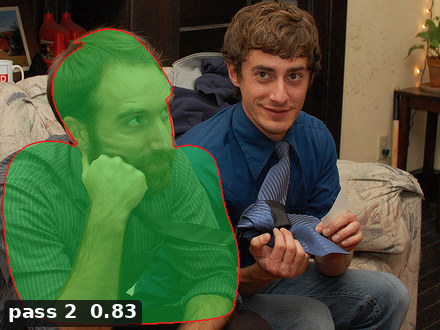} & \qim{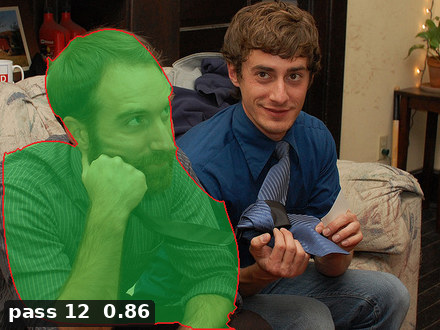} \\[3pt]
    \multicolumn{6}{@{}p{\linewidth}@{}}{\bqry{the goat nearest to the bottom stone} \hfill {\scriptsize ReasonSeg, 11 passes}} \\[1pt]
    \qim{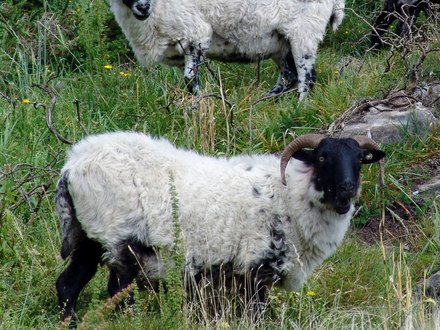} & \qim{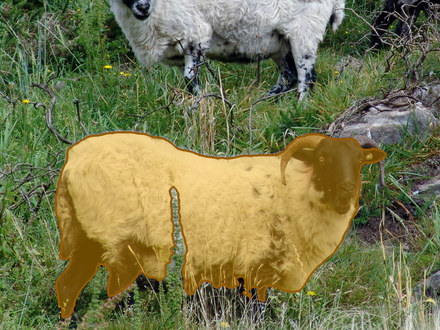} & \qim{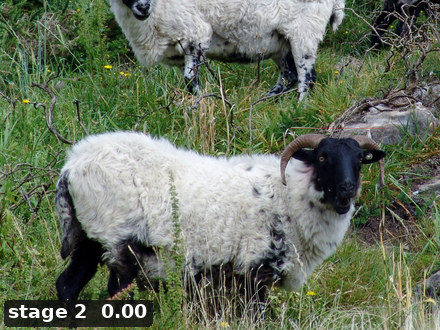} & \qim{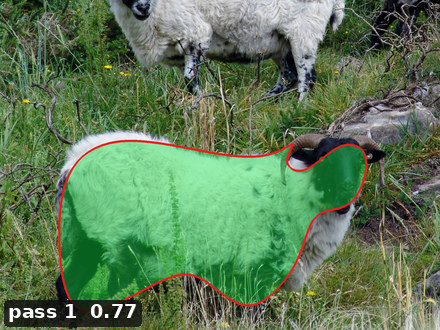} & \qim{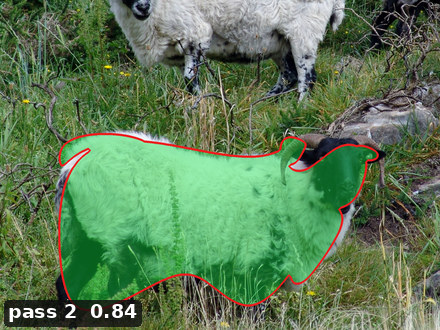} & \qim{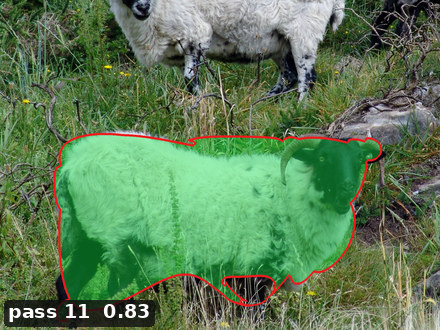} \\[3pt]
    \multicolumn{6}{@{}p{\linewidth}@{}}{\bqry{bicycle} \hfill {\scriptsize LVIS, 11 passes}} \\[1pt]
    \qim{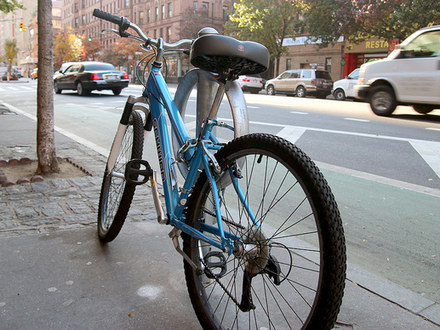} & \qim{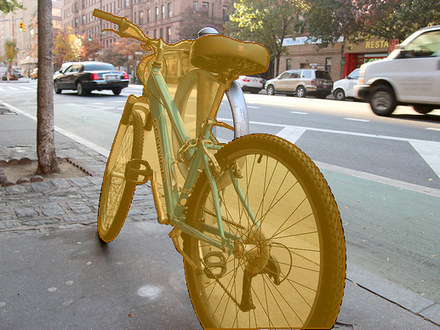} & \qim{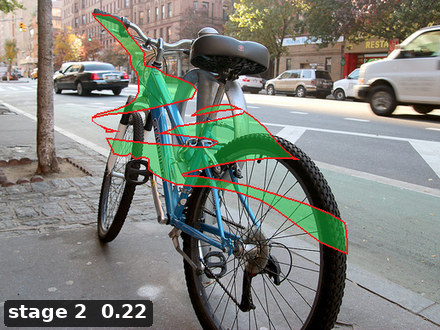} & \qim{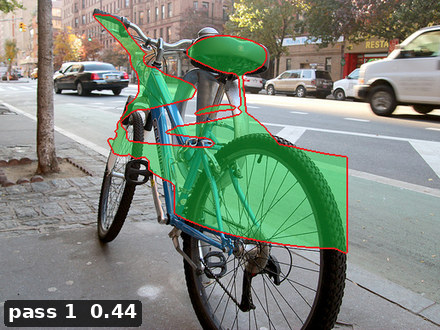} & \qim{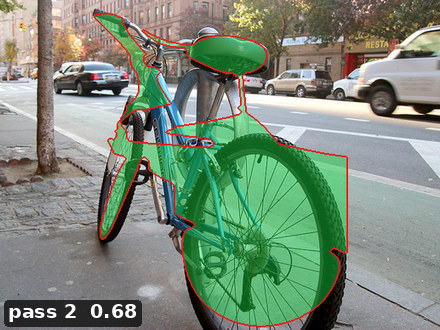} & \qim{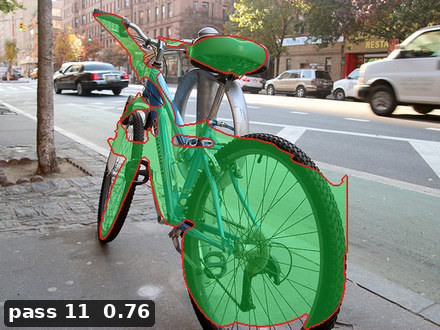} \\[3pt]
    \multicolumn{6}{@{}p{\linewidth}@{}}{\bqry{synthetic rug} \hfill {\scriptsize SA-Co/Gold, 6 passes}} \\[1pt]
    \qim{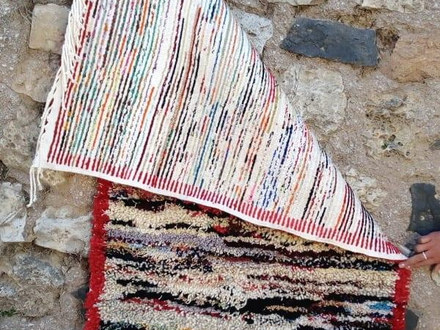} & \qim{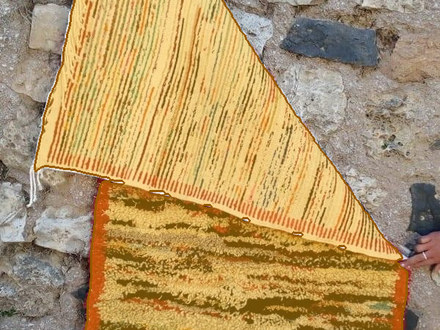} & \qim{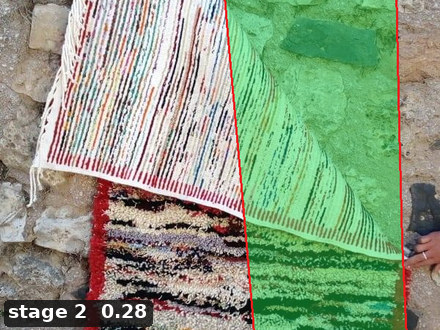} & \qim{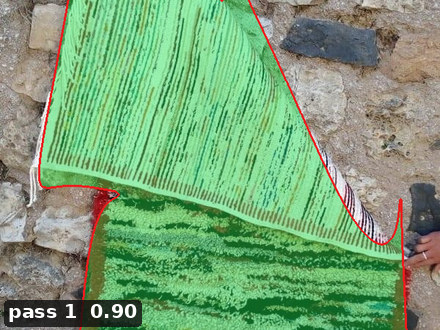} & \qim{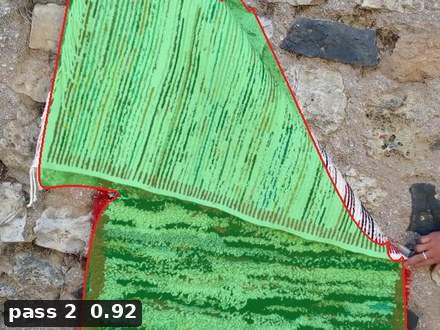} & \qim{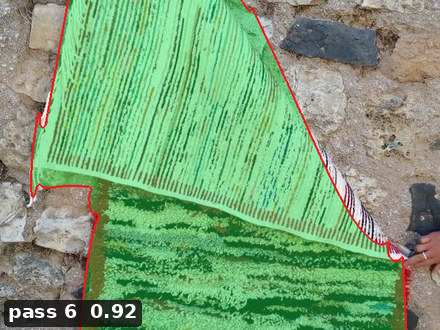} \\
  \end{tabular}
  \vspace{-6pt}
  \caption{\small \textbf{What one refinement pass does.} Four items, one per
  task format, for which the final mask substantially outperforms the stage-2
  mask. Each is shown at stage~2, after the first two passes, and after the
  final pass. Each chip identifies the state and reports its IoU. The
  intermediate masks are replayed from each item's refinement record using
  the pipeline's own composition code.}
  \label{fig:supp_passes}
\end{figure}

  Four grids extend the qualitative evidence of \secLabel\ref{sec:results} to
  cases discussed only in aggregate in the main paper: images containing many
  instances of the queried class (\figLabel\ref{fig:supp_multi}), SA-Co prompts
  whose concept is absent from the image (\figLabel\ref{fig:supp_neg}), the rare
  visual domains of RF-100VL (\figLabel\ref{fig:supp_domain}), and the effect of
  an individual refinement pass (\figLabel\ref{fig:supp_passes}). Every panel
  comes from the same \gfl runs at $T{=}12$ used for the tables. Each baseline
  panel comes from the corresponding \samthree or \samagent run scored on the
  same item. Each chip reports that system's recorded IoU, so its value is
  consistent with the reported results.


\clearpage
\section{Extended Analysis} \label{app:analysis_extended}
\vspace{-4pt}

This section expands the findings of \secLabel\ref{sec:analysis}. The
complete ablation results are in \tblLabel\ref{tab:ablations} and
\tblLabel\ref{tab:ablations_refcoco}. The backend study and ablations use the
study configuration described in \secLabel\ref{app:config}.

\subsection{Refinement and convergence} \label{app:convergence}
\vspace{-2pt}

\mysection{Does the loop actually converge?}
This matters more than a typical hyperparameter question. \eqn{eq:refine} is
an open loop driven by a stochastic model that can propose a fresh set of focus
boxes at every pass, with no contraction guarantee. Left unchecked, such a loop
can cycle indefinitely, repainting a boundary one way and then the other. A
reader is right to ask whether the pass budget $T$ provides a genuine
early stop or merely truncates an unstable process.

\begin{figure}[t]
  \centering
  \includegraphics[width=\textwidth]{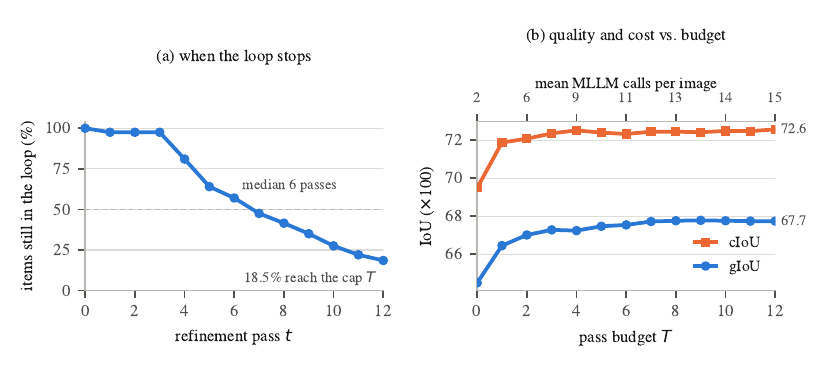}
  \vspace{-6pt}
  \caption{\small \textbf{The refinement loop stops on its own.} ReasonSeg val, all
  200 items, \gfl. \textbf{(a)} Share of items still inside the loop at each
  pass. A run stops once the mask changes by less than $0.5\%$ of its box
  area for three consecutive passes.
  \textbf{(b)} The same run's mask after the first $T$ passes, scored at every
  budget, with the mean MLLM calls on the top axis.}
  \label{fig:residual}
\end{figure}

\figLabel\ref{fig:residual}(a) answers this empirically on ReasonSeg val. The
stopping criterion of \secLabel\ref{sec:experiments} ends the loop for $79\%$ of
items before they reach the budget ($81\%$ of the items that enter the loop).
The median item uses $6$ passes, $36\%$ stop within $4$, and only $18.5\%$ are
still running when $T$ cuts them off. This behavior is consistent with the
intended damping mechanism: when successive passes agree on a boundary, the
commitment field makes those pixels progressively harder to overturn. However,
the study does not isolate the soft lock's contribution. The minority that
reaches the cap is a harder set to begin with: its stage-2 masks are already
$5.7$ gIoU behind ($60.0$ vs.\ $65.7$), and its final masks remain $7.0$ behind
($62.3$ vs.\ $69.3$).
Normalizing the criterion by box area rather than image area makes its
threshold relative to the region the loop can alter.

\figLabel\ref{fig:residual}(b) ties this to accuracy rather than to mask
stability alone. Quality saturates well inside the budget: the first pass buys
$+2.0$ gIoU, four passes buy $+2.7$ of the eventual $+3.2$, and passes $8$ to
$12$ move it by less than $0.1$. These figures come from this one run. Other
runs of \gfl give different totals ($+3.7$ in \tblLabel\ref{tab:backends}, and
$5.4$ gIoU between the ablation reference and its no-refinement row), because
the pipeline is stochastic and the runs differ in configuration. Because most items have already stopped, the
measured cost is far below the worst-case cost of $2T{+}2 = 26$ calls: $14.7$
calls per image on average at $T{=}12$ in this run (other $T{=}12$ runs give
$14.3$ to $15.0$), and about $9$ at the $T{=}4$ knee. $T$ is
a safety net rather than an operating point.

\mysection{The residual as a stopping signal} \label{app:stopping}

\begin{figure}[h]
  \centering
  \includegraphics[width=\textwidth]{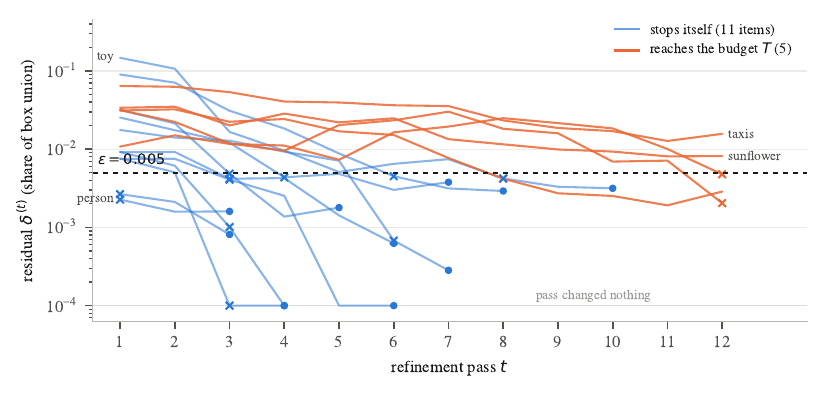}
  \vspace{-6pt}
  \caption{\small \textbf{Per-item refinement residuals on the development set.}
  Each curve is one development image, run with \gfl at budget $T{=}12$.
  The curve is the $3$-pass moving average of the mask residuals $\delta^{(t)}$ on a
  logarithmic axis, and the dashed line is the threshold $\varepsilon=0.005$.
  Blue curves are images that stop on their own with stopping patience $s=3$, orange curves are images that
  exhaust the budget, and $\times$ marks the first smoothed pass below
  $\varepsilon$. The residuals decay as the masks stabilize, which is the
  signal the stopping criterion uses in \ours.}
  \label{fig:residual_traces}
\end{figure}

The stopping criterion of \secLabel\ref{sec:experiments} acts on the residual
$\delta^{(t)} = |M^{(t)} \triangle M^{(t-1)}| \,/\, |\bigcup_{k=1}^{N} b_k|$,
the fraction of the instance-box union that a pass repaints, rather than on
quality: the pipeline sees no ground truth at run time, so the amount of mask
change is its only observable. \figLabel\ref{fig:residual_traces} shows
that quantity pass by pass on a $16$-image development set (free text
queries on images that belong to no evaluation benchmark), which is where
$\varepsilon$ and $s$ were set before any benchmark was run. The $16$ images
and their queries \ifdefined\arxivversion are released with the code.\else are attached to the supplementary material with the code.\fi

Three observations stand out. First, the residual decays. It is not monotone,
since a pass is free to propose a large new edit at any point, but every trace
ends below where it started, most of them by an order of magnitude or more,
and several reach the axis floor, that is, passes that changed no pixel at
all. Second, the rate is a property of the image, not a constant. The first
pass repaints between $0.2\%$ (\emph{person}) and $27\%$ (\emph{toy}) of the
box union, a $175\times$ spread (based on unsmoothed values, whereas the figure plots
the $3$-pass average), and the starting value does not predict the decay:
\emph{toy} starts highest and stops itself at pass $10$, while \emph{taxis},
with six crowded instances of the same class, starts at $6\%$ and is still
moving $0.7\%$ of its box union at pass $12$ ($1.3\%$ after the $3$-pass
average). Boundary length, occlusion, and instance count may affect how quickly
an item stabilizes, but the first pass does not reveal it. This is why the
budget must be allocated per image rather than fixed globally.
Third, $\varepsilon{=}0.005$ with $s{=}3$ sits where those two regimes
separate: $11$ of the $16$ items stop on their own, the earliest at pass $3$
(the patience floor, which no item can beat), the median at $6$, and $5$ run
to the cap.

A small residual indicates stability, not correctness. A loop could in
principle settle on a wrong mask. Its use as a surrogate is supported by the
benchmark-scale behavior above: quality saturates over the same passes in which
these residuals collapse, and the items that never stop are the ones that end
furthest behind.

\subsection{Backend study} \label{app:backend_study}
\vspace{-2pt}

\begin{figure}[t]
  \centering
  \includegraphics[width=\textwidth]{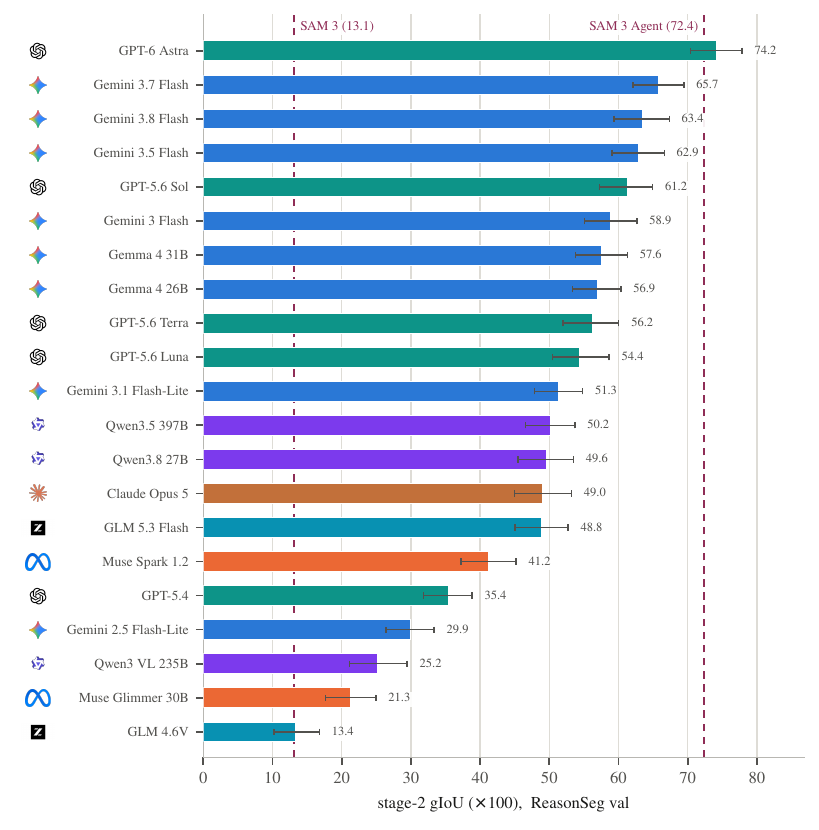}
  \vspace{-6pt}
  \caption{\small \textbf{Complete backend comparison in \ours.} All 21 backends we ran,
  including Gemini~3.8~Flash, Gemma~4~26B, and Qwen3.8~27B. Stage-2 gIoU, 95\%
  item-bootstrap intervals, and baselines use the same 200 ReasonSeg val items.}
  \label{fig:backends_full}
\end{figure}

Because \ours adds no trained parameters, backend progress is the only thing
that has to improve for segmentation to improve. We test this directly by
running the same pipeline, in the study configuration
(\secLabel\ref{app:config}), on ReasonSeg val with \textbf{twenty-one} MLLM
backends from six developers. The only change is which model answers the three
calls.
\figLabel\ref{fig:backends_full} is the complete ranking, read on the stage-2 mask
so that one bar carries one backend. \tblLabel\ref{tab:backends} repeats every
row with its final mask, both bootstrap intervals and the signed effect of
refinement. First, scores range from $13.4$ to $74.2$ gIoU at stage 2 and from
$14.2$ to $72.9$ after refinement, a spread far wider than that of any design
choice in \tblLabel\ref{tab:ablations}.
Second, refinement lowers gIoU on eleven of the twenty-one backends,
especially the weaker models, which cannot refine their own masks well.
Its gains therefore depend on the model's ability to inspect and correct
its own masks.
Every backend scores at least as high
as the \samthree baseline score ($13.1$) on this benchmark, although the weakest
is not clearly above it: GLM~4.6V scores $13.4$ ($[10.2, 16.8]$) at stage 2 and
$14.2$ ($[11.0, 17.6]$) after refinement. The other twenty backends score clearly above \samthree. At $72.4$, the \samagent baseline uses the
same \gfl as our pipeline and a segmentation specialist as its tool. It is
close to our best backend, GPT-6~Astra, which scores $74.2$ ($[70.4, 77.9]$) at
stage 2 and $72.9$ ($[69.2, 76.6]$) after refinement. The gaps are $1.8$ and
$0.5$ gIoU, respectively, and the \samagent score lies within both corresponding
GPT-6~Astra item-bootstrap intervals. By contrast, \samthree alone trails
GPT-6~Astra by $59.8$.
The two dashed lines also respond very differently to the MLLM. Swapping the
agent's MLLM among Gemma-4-31B, Gemini-3.5-Flash, and \gfl changes its score by
just $0.7$ gIoU ($72.4$ to $73.1$), because its masks come from \samthree
either way. Swapping our MLLM across the same Gemini generations changes our
score by $10.8$. Our score is therefore far more sensitive to the backend than the agent's is,
and the backend is the part of the stack that continues to improve.
\figLabel\ref{fig:qual_backend} shows what that spread looks like on single
items: the same five ReasonSeg queries traced by seven of the backends, where
the object being selected barely changes down a row and the outline drawn
around it does.

\begin{figure}[h]
  \centering
  \includegraphics[width=\textwidth]{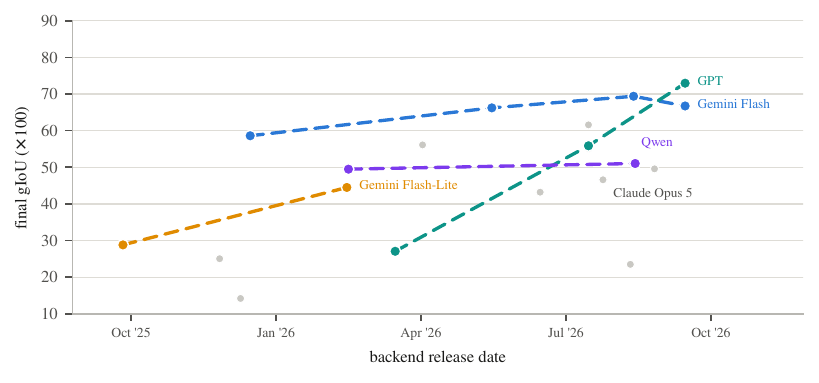}
  \vspace{-6pt}
  \caption{\small \textbf{Backend progress is \ours progress.} \ours final segmentation gIoU
  on ReasonSeg val against backend release date, one coloured line per product
  line with two or more releases. All other backends are gray. Nothing in the
  pipeline changes between the points of a line.}
  \label{fig:scaling}
\end{figure}

\begin{figure}[h]
  \centering
  \includegraphics[width=\textwidth]{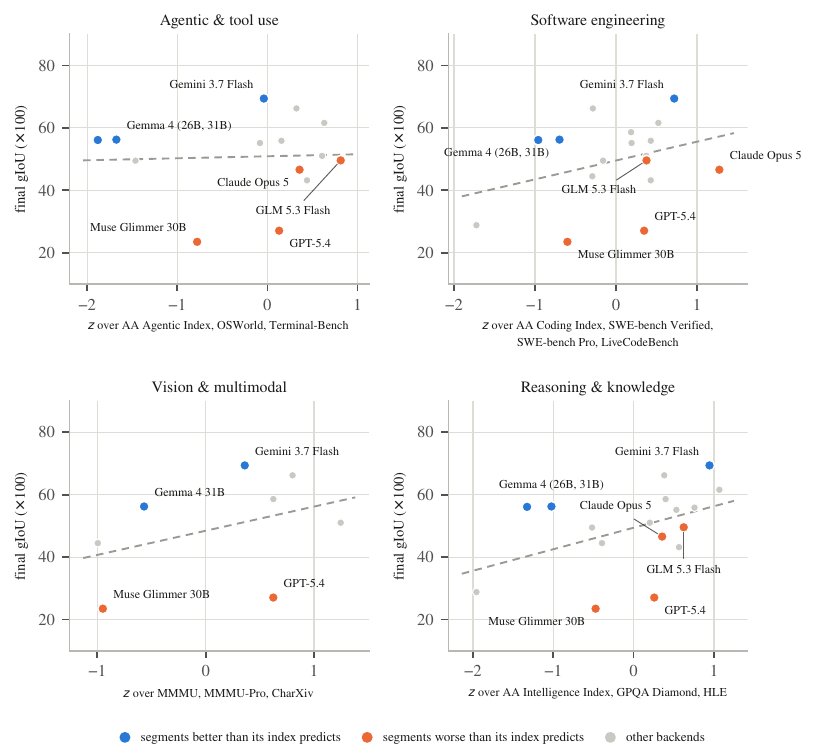}
  \vspace{-6pt}
  \caption{\small \textbf{Which kind of capability predicts \ours quality?} Final
  gIoU on ReasonSeg val is plotted against four public-benchmark capability axes,
  with dashed OLS fits. Slopes (gIoU per unit $z$) are \emph{7.7 for vision}
  and \emph{6.9 for reasoning}, followed by \emph{6.0 for software} and 0.6 for
  agentic skill. Reasoning and vision track segmentation quality most closely.
  Agentic skill shows almost no association.}
  \label{fig:capability}
\end{figure}

\mysection{Public capability and segmentation quality}
\figLabel\ref{fig:capability} plots that quality against four capability axes
built from public benchmark scores (\S\ref{app:backends}), one panel per axis.
Two things follow. First, no published capability is a reliable predictor of
segmentation quality: no axis reaches significance. The reasoning-and-knowledge
axis comes closest ($\rho = 0.43$, $p = 0.09$, $n{=}17$), followed by the
vision-and-multimodal axis ($\rho = 0.36$, $n{=}8$) and software engineering
($\rho = 0.21$, $p = 0.42$). Agentic skill carries no signal ($\rho = -0.05$,
$n{=}14$). The exceptions run in both directions.
Claude~Opus~5 ($96.0$ on SWE-bench Verified) and GLM~5.3~Flash score high on
software engineering and reasoning yet segment at $46.6$ and $49.6$,
fourteenth and twelfth of twenty-one. GPT-5.4 sits on the right of every panel
but near the bottom in segmentation quality ($27.1$). Muse~Glimmer~30B performs
similarly ($23.5$). We did not diagnose where these backends lose accuracy.
The reverse holds as well. Gemma~4~31B and Gemma~4~26B sit at the low end of
the agentic, software and reasoning axes and still segment seventh and eighth
at $56.2$ and $56.1$, and \gfl lies further above the fit than any other
backend on two of the four axes (agentic and vision). The practical reading rests on those
exceptions rather than on the fit. No published headline score, whether for
vision or coding, is a reliable way to pick a backend for \ours, and the only dependable
test is to run the pipeline on the model.

\begin{figure}[h]
  \centering
  \includegraphics[width=0.55\textwidth]{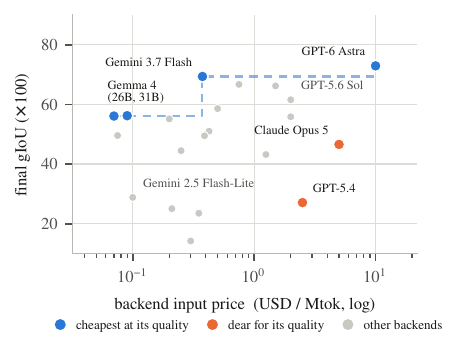}
  \vspace{-6pt}
  \caption{\small \textbf{Segmentation quality is not bought.} Backend input price is plotted
  against \ours final gIoU on ReasonSeg val. The dashed staircase marks the
  cheapest-so-far frontier, whose backends are shown in blue. Price does not
  order quality. Gemini 3.7 Flash is the cheapest backend that reaches $72.9$ final gIoU, which motivates our \ours choice of backend for the scalable main results.}
  \label{fig:costquality}
\end{figure}

Second, and more directly useful, \figLabel\ref{fig:scaling} re-plots the
same runs against backend release date, joining models within each product
line. All four product lines with multiple releases improve over the period
shown while the pipeline remains unchanged. Gemini Flash goes
$58.6 \to 66.2 \to 69.4 \to 66.7$ over four releases in nine months, Gemini
Flash-Lite $28.8 \to 44.5$, GPT $27.1 \to 55.9 \to 72.9$, and Qwen $49.5 \to 51.0$.
The trend is not monotone at the release level: Gemini~3.8~Flash scores $2.7$
below Gemini~3.7~Flash, a reminder that a
newer model is not automatically a better backend. The Flash-Lite line is the
most economical evidence: a $+15.7$ gIoU gain without retraining anything.
Size within a generation buys little: Gemma~4~31B and its
mixture-of-experts sibling Gemma~4~26B, released the same day, land $0.1$
apart ($56.2$ and $56.1$). The two Z.ai models are not one product line and the
generation plot joins no arrow between them, but they bracket the same eight months as
the Gemini Flash line: GLM~4.6V, of December 2025, is the weakest backend in
the study at $14.2$, and GLM~5.3~Flash, of August 2026, reaches $49.6$. Price
buys much less than generation does ($\rho = 0.25$ against input price,
\figLabel\ref{fig:costquality}). The second-best backend here costs
$21\times$ less than the best.

One caveat belongs with these numbers. Refinement is not uniformly positive
across backends: it hurts on eleven of the twenty-one. It adds $+3.7$ gIoU on
\gfl but \emph{costs} $8.3$ on GPT-5.4, $6.8$ on
\gfli, $2.4$ on Claude~Opus~5, $1.3$ on Gemma~4~31B and $1.2$ on GPT-6~Astra,
our strongest backend. The loop can only re-trace what the backend proposes, so a
backend that mislocates its focus boxes spends its passes moving the mask off
the object. The refinement gains reported elsewhere in this paper are \gfl's,
and the appropriate refinement budget must be checked for each backend. This is also why
\figLabel\ref{fig:backends} ranks the backends on what they trace: a bar pair
would read as a refinement result on rows where refinement is not one.

\begin{table}[h]
  \centering
  \small
  \setlength{\tabcolsep}{4pt}
  \begin{tabular}{llcccr}
\toprule
Backend & Developer & Release & Stage-2 gIoU & Final gIoU & $\Delta$ \\
\midrule
GPT-6 Astra$^{\dagger}$ & OpenAI & 2026-09 & \textbf{74.2} \ci{70.4}{77.9} & \textbf{72.9} \ci{69.2}{76.6} & -1.2 \\
Gemini 3.7 Flash & Google & 2026-08-13 & 65.7 \ci{62.1}{69.5} & 69.4 \ci{65.9}{72.8} & +3.7 \\
Gemini 3.8 Flash$^{\dagger}$ & Google & 2026-09 & 63.4 \ci{59.4}{67.3} & 66.7 \ci{63.0}{70.4} & +3.3 \\
Gemini 3.5 Flash & Google & 2026-05 & 62.9 \ci{59.1}{66.6} & 66.2 \ci{62.4}{70.0} & +3.3 \\
GPT-5.6 Sol$^{\dagger}$ & OpenAI & 2026-07 & 61.2 \ci{57.3}{64.9} & 61.6 \ci{57.7}{65.5} & +0.3 \\
Gemini 3 Flash & Google & 2025-12 & 58.9 \ci{55.1}{62.7} & 58.6 \ci{55.1}{61.9} & -0.3 \\
Gemma 4 31B$^{\dagger}$ & Google & 2026-04-02 & 57.6 \ci{53.8}{61.3} & 56.2 \ci{52.7}{59.9} & -1.3 \\
Gemma 4 26B$^{\dagger}$ & Google & 2026-04-02 & 56.9 \ci{53.3}{60.3} & 56.1 \ci{52.3}{59.5} & -0.8 \\
GPT-5.6 Terra & OpenAI & 2026-07 & 56.2 \ci{52.0}{60.0} & 55.9 \ci{51.6}{60.1} & -0.3 \\
GPT-5.6 Luna & OpenAI & 2026-07 & 54.4 \ci{50.5}{58.7} & 55.1 \ci{51.0}{59.0} & +0.7 \\
Gemini 3.1 Flash-Lite & Google & 2026-02 & 51.3 \ci{47.8}{54.8} & 44.5 \ci{41.2}{47.8} & -6.8 \\
Qwen3.5 397B$^{\dagger}$ & Qwen & 2026-02-16 & 50.2 \ci{46.6}{53.7} & 49.5 \ci{45.9}{53.1} & -0.7 \\
Qwen3.8 27B$^{\dagger}$ & Qwen & 2026-08-14 & 49.6 \ci{45.5}{53.5} & 51.0 \ci{47.1}{54.7} & +1.4 \\
Claude Opus 5$^{\dagger}$ & Anthropic & 2026-07-24 & 49.0 \ci{44.9}{53.2} & 46.6 \ci{42.6}{50.5} & -2.4 \\
GLM 5.3 Flash$^{\dagger}$ & Z.ai & 2026-08-26 & 48.8 \ci{45.1}{52.7} & 49.6 \ci{45.7}{53.3} & +0.7 \\
Muse Spark 1.2$^{\dagger}$ & Meta & 2026-06 & 41.2 \ci{37.2}{45.2} & 43.2 \ci{39.1}{46.9} & +2.0 \\
GPT-5.4 & OpenAI & 2026-03 & 35.4 \ci{31.8}{38.8} & 27.1 \ci{23.8}{30.8} & -8.3 \\
Gemini 2.5 Flash-Lite & Google & 2025-09-26 & 29.9 \ci{26.4}{33.4} & 28.8 \ci{25.6}{32.1} & -1.1 \\
Qwen3 VL 235B & Qwen & 2025-11-26 & 25.2 \ci{21.1}{29.5} & 25.1 \ci{20.8}{29.4} & -0.2 \\
Muse Glimmer 30B$^{\dagger}$ & Meta & 2026-08-11 & 21.3 \ci{17.7}{24.9} & 23.5 \ci{20.0}{27.3} & +2.2 \\
GLM 4.6V$^{\dagger}$ & Z.ai & 2025-12-09 & 13.4 \ci{10.2}{16.8} & 14.2 \ci{11.0}{17.6} & +0.9 \\
\bottomrule
\end{tabular}
    \caption{\small \textbf{Twenty-one backends.} \ours on all 200 ReasonSeg val
  items, using the same pipeline and settings. Only the MLLM differs.
  Both masks are shown, with 95\% bootstrap intervals and the signed effect
  $\Delta$ of refinement.
  $\dagger$ marks a longer per-call budget.}
  \label{tab:backends}
\end{table}

\begin{figure}[t]
  \centering
  \setlength{\tabcolsep}{0.5pt}
  \setlength{\qgw}{0.1075\linewidth}
  \renewcommand{\arraystretch}{1.0}
  \begin{tabular}{@{}ccccccccc@{}}
    \qhd{Input} & \qhd{\makecell{Ground\\truth}} & \qhd{\makecell{GPT-6\\Astra}} & \qhd{\makecell{Gemini\\3.7 Flash}} & \qhd{\makecell{Gemma\\4 31B}} & \qhd{\makecell{GPT-5.6\\Terra}} & \qhd{\makecell{Qwen3.8\\27B}} & \qhd{\makecell{Claude\\Opus 5}} & \qhd{\makecell{Muse\\Spark 1.2}} \\[2pt]
    \multicolumn{9}{@{}l@{}}{\bqry{What object is used to cover the entrance of the bathroom and ensure privacy?}} \\[1pt]
    \qim{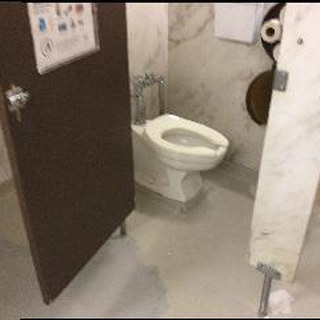} & \qim{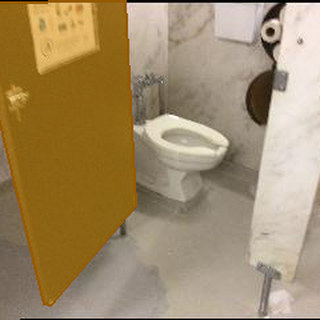} & \qim{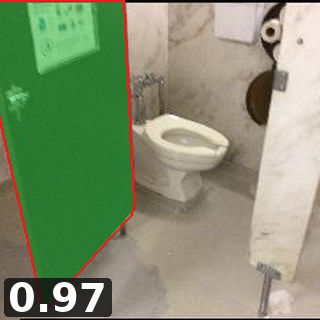} & \qim{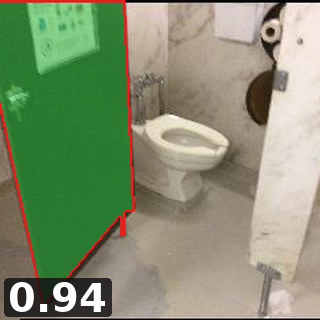} & \qim{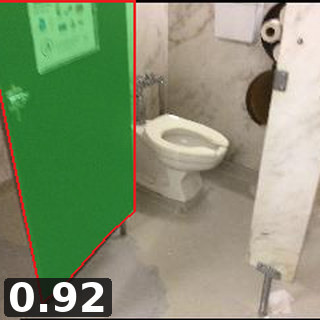} & \qim{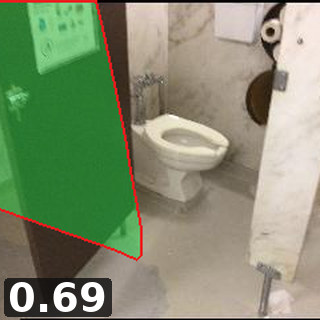} & \qim{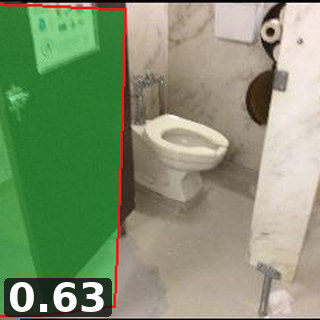} & \qim{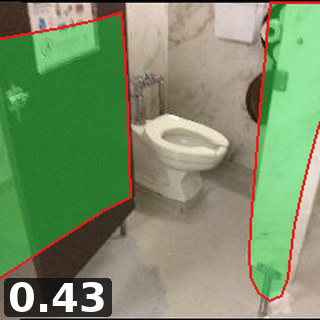} & \qim{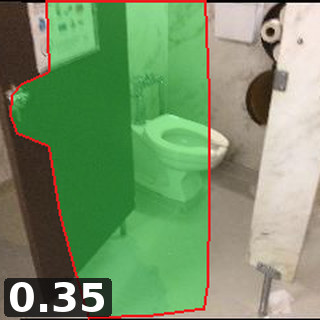} \\[3pt]
    \multicolumn{9}{@{}l@{}}{\bqry{the dog that puts its tongue out}} \\[1pt]
    \qim{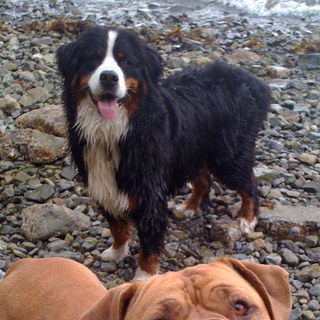} & \qim{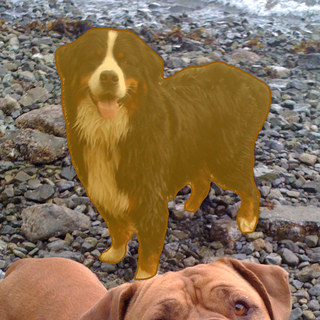} & \qim{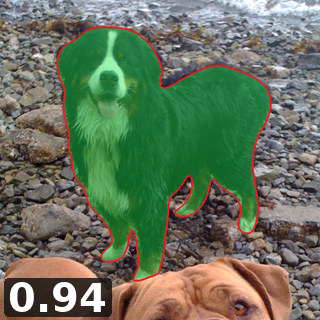} & \qim{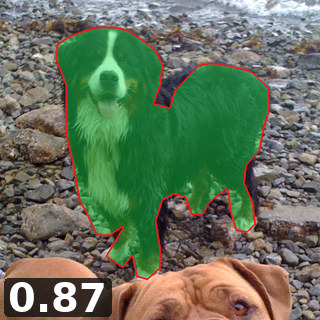} & \qim{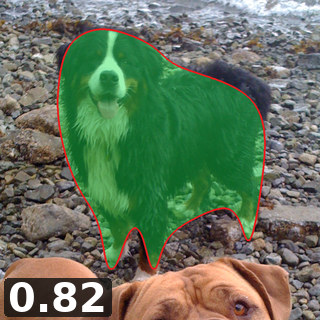} & \qim{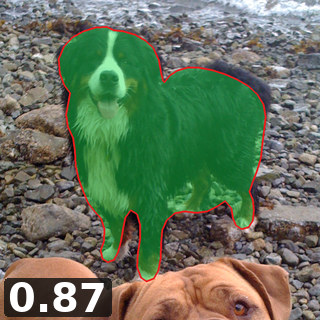} & \qim{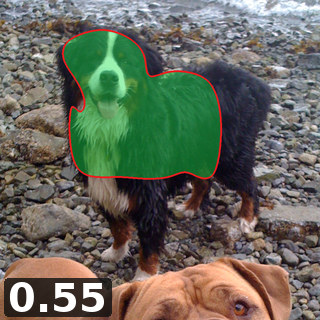} & \qim{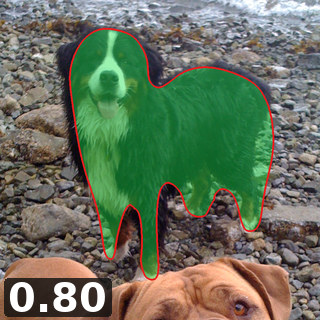} & \qim{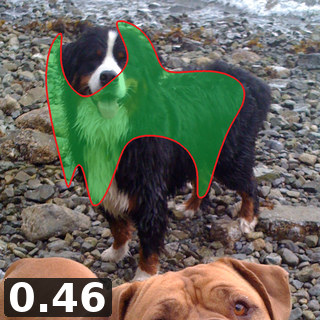} \\[3pt]
    \multicolumn{9}{@{}l@{}}{\bqry{\ldots what area in the picture would a cyclist use to navigate through the city?}} \\[1pt]
    \qim{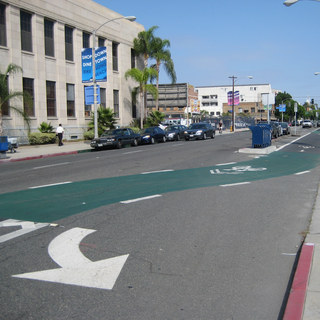} & \qim{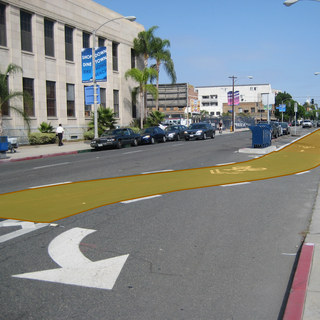} & \qim{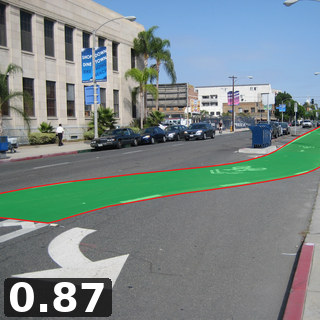} & \qim{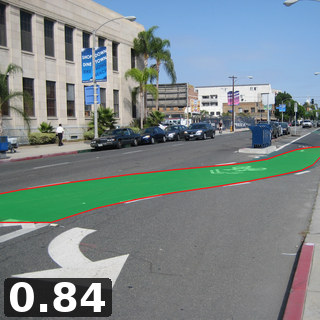} & \qim{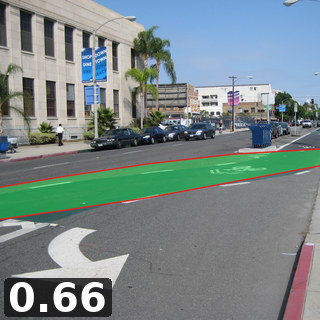} & \qim{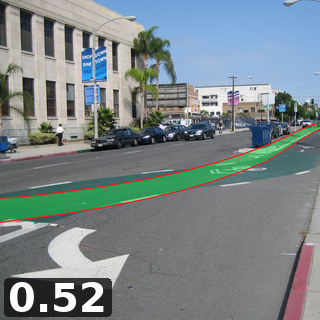} & \qim{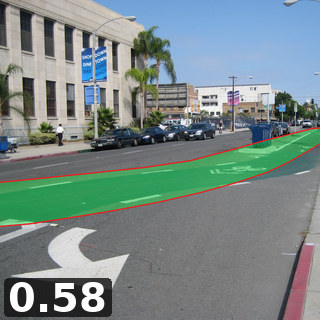} & \qim{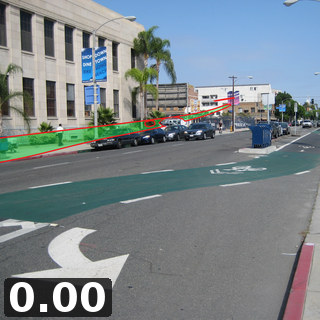} & \qim{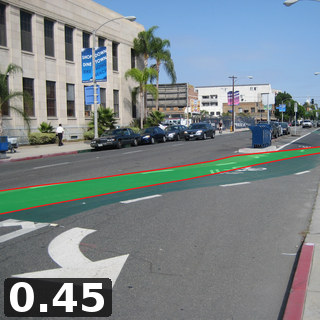} \\[3pt]
    \multicolumn{9}{@{}l@{}}{\bqry{\ldots what in the picture could be used to display such signs or symbols?}} \\[1pt]
    \qim{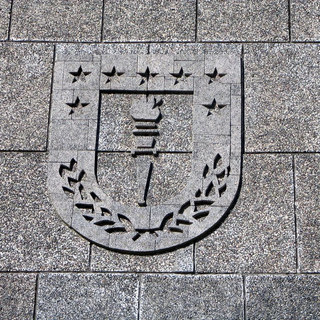} & \qim{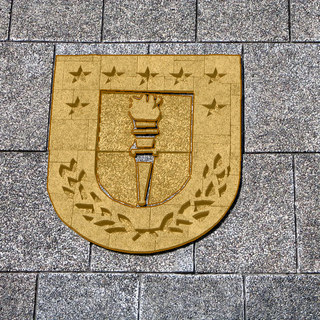} & \qim{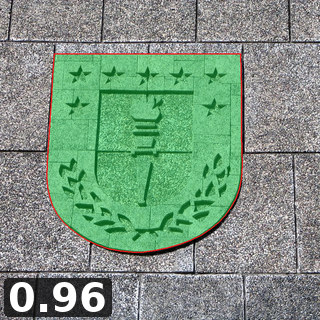} & \qim{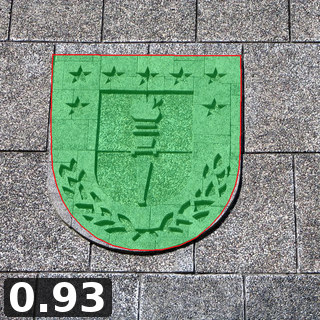} & \qim{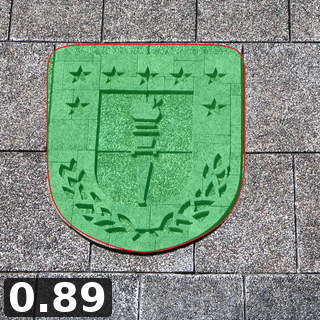} & \qim{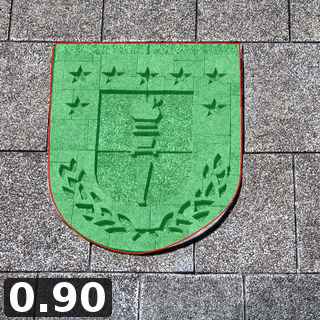} & \qim{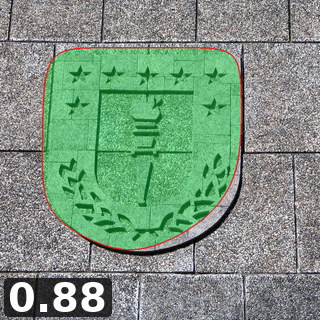} & \qim{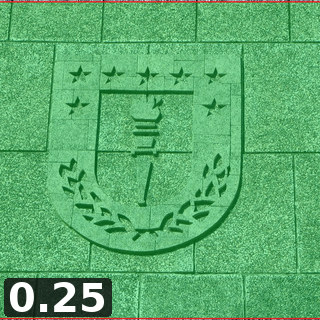} & \qim{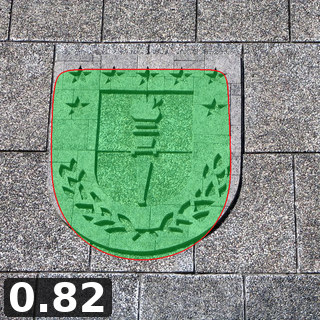} \\[3pt]
    \multicolumn{9}{@{}l@{}}{\bqry{\ldots where can we throw away the rest of the food and scraps?}} \\[1pt]
    \qim{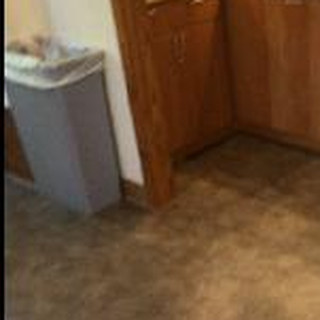} & \qim{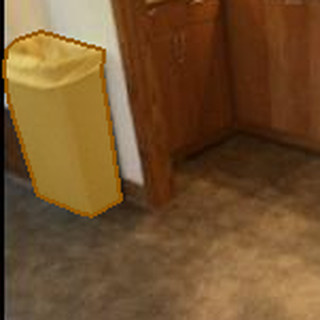} & \qim{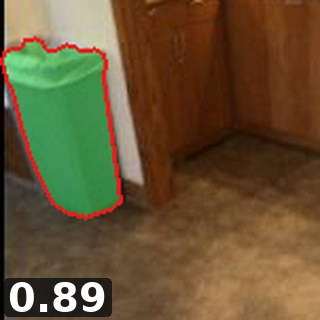} & \qim{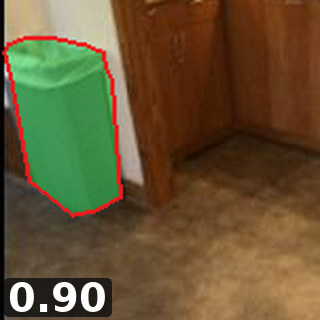} & \qim{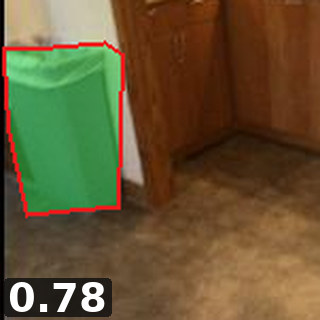} & \qim{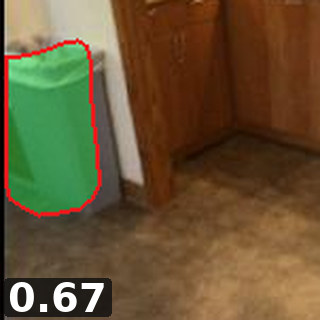} & \qim{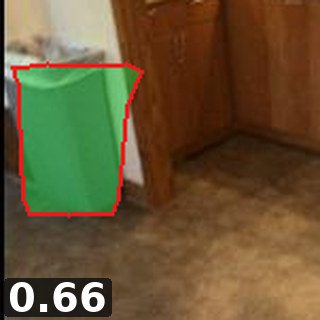} & \qim{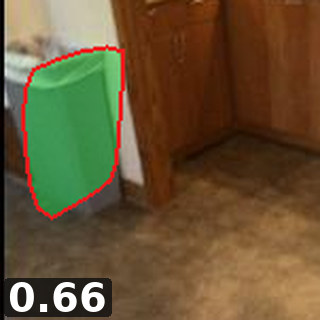} & \qim{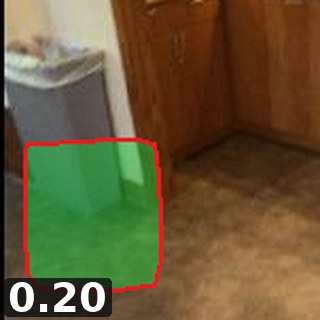} \\
  \end{tabular}
  \vspace{-6pt}
  \caption{\small \textbf{What the MLLM backend gap looks like in \ours.} The
  same pipeline is used throughout, with one column per MLLM backend, ordered
  by split-level stage-2 gIoU, best first. Every panel shows the \emph{stage-2}
  mask, so the comparison is of what each backend traces rather than how it
  refines. Grounding is largely shared, but mask outlines deteriorate down the rows with week MLLMs.}
  \label{fig:qual_backend}
\end{figure}

\subsection{Ablations and design alternatives} \label{app:ablations_other}
\vspace{-2pt}

\tblLabel\ref{tab:ablations} and \tblLabel\ref{tab:ablations_refcoco} test the design choices of
\secLabel\ref{sec:method} on two splits, comparing final-mask quality
(gIoU and cIoU) with MLLM calls per image. Each
split (ReasonSeg val and RefCOCO val) contains 200 items. The reference
(shaded) row is the configuration reported in the paper, with topology
conditioning off, run at the study budget $T{=}12$. The tables report fourteen
variants per split: three change the reported configuration directly (cleanup
and box adjustment), one adds topology conditioning, and the remaining ten
(indented) apply their change on top of topology conditioning, so each of them
is compared with the topology-on row. Both tables give
gIoU, cIoU and calls per image, and the ReasonSeg table adds run-to-run
deviations and wall-clock time. Every ReasonSeg
configuration was run three times with an identical setup, and each cell
reports the mean. Across repeated runs, a fixed configuration's score varies
by $0.1$ to $1.3$ gIoU, which sets the scale for interpreting the comparisons
below. Each RefCOCO configuration was run once, so no deviation is reported.
These results should be read for direction only.
At similar call counts, adding topology conditioning to the reported
configuration reduces observed cIoU by $2.1$ on ReasonSeg and $2.5$ on RefCOCO.
The reductions have the same sign on both splits and for gIoU as well ($1.5$
and $2.6$, respectively), so we leave topology off by default and retain it as
an option for the item type it was designed for (\secLabel\ref{app:holes}). The
largest effect of any component is refinement: removing it costs $5.4$ gIoU
and $5.4$ cIoU on ReasonSeg, and $9.8$ gIoU and $12.7$ cIoU on RefCOCO,
relative to the topology-on row it modifies, and refinement accounts for $12.3$
of that row's $14.3$ calls per image. The remaining changes generally produce
smaller differences. Apart from removing refinement, the largest gIoU drop
relative to the reported configuration is for low image detail: $3.5$ gIoU
($3.2$ cIoU) on ReasonSeg and $4.5$ gIoU ($4.4$ cIoU) on RefCOCO. Most
variants are within two points of the reported configuration on both splits.
No variant outperforms the reported configuration on both metrics across both
splits. Cost varies more across
the alternatives than quality does: calls range from $8.8$ without the detect
stage to $21.3$ with per-instance cropped refinement, while on ReasonSeg the
final gIoU of every other variant stays between $67.2$ and $71.8$. The pipeline
does not appear to be delicately balanced on prompt conditioning or on how its
calls are batched.

\emph{Prompt conditioning.} Cutting the stage-1 morphology or topology
description from the prompts does not measurably change final-mask quality.
The topology-on row includes both descriptions. Relative to its
$69.2 \pm 1.3$ gIoU, removing morphology yields $69.9$ and removing topology,
which gives the reported configuration, yields $70.7$. Both differences are
small relative to the spread across configurations. On RefCOCO, the
corresponding rows are $2.4$ and $2.6$ gIoU above the topology-on row
(\tblLabel\ref{tab:ablations_refcoco}).
Topology conditioning is a design option, not a component on which
the aggregate results depend. We report the configuration without the extra
topology prompt text because describing holes does not measurably improve
split-level scores. Topology information does provide structural control:
Stage~2 uses it for hole carving and gated cleanup. Without it, the model is
not told how many closed curves to fit, and a holed or multi-part object may be
cleaned as if it were one solid blob.
The fraction of benchmark targets for which this control applies ranges
from $0\%$ to $7\%$. On items scored both ways, the paired difference is
$-0.3$ gIoU overall and $+0.1$ on holed items. The option is
retained for explicit structural control, not claimed as a source of aggregate
accuracy (\secLabel\ref{app:holes}).
Reducing the backend's reasoning effort or the image detail sent with each
call leaves quality flat to slightly below the topology-on row it modifies:
the corresponding cIoU scores are $72.3$ and $71.5$ here, against $72.6$. On
RefCOCO, they are $74.3$ and $75.4$, against $77.3$.
Low reasoning effort is the only setting that is cheaper: at the same
call count, backend-call time per image falls by about a quarter, from $175\,s$
to $133\,s$. The deterministic cleanup likewise does not carry the result.
Removing it from the reported configuration gives $74.1$ cIoU against $74.7$
($77.9$ against $79.8$ on RefCOCO), and removing it from the topology-on row
gives $73.8$ against $72.6$ ($76.8$ against $77.3$). This matters because it is
the only step in \ours whose geometry the model did not choose.

\emph{Is the detection stage needed?} Asking the model for splines directly,
with no detect stage and therefore no box to confine control points to, tests
whether stage~1 is worth its call. Without a box, the model must place absolute
coordinates over the whole image, and no invariant prevents a loop from
wandering across the scene.
On ReasonSeg, the no-detection variant is within the run-to-run band of the
topology-on row it modifies ($69.5$ gIoU and $72.1$ cIoU versus $69.2$ and
$72.6$) and uses fewer
calls ($8.8$ vs.\ $14.3$).
RefCOCO is also competitive ($78.5$ gIoU at $10.4$ calls against $76.7$ at
$16.8$). Stage-2 cIoU does favor detection on both splits
(\tblLabel\ref{tab:ablations_main}). The advantage is concentrated on targets
whose ground-truth masks are disconnected (at least two components above the
annotation-speck threshold: $63/200$ ReasonSeg and $18/200$ RefCOCO items). On
ReasonSeg, the initial cIoU on these targets is $52.6$ without the detect stage
and $63.9$ with it. On RefCOCO, it is $61.7$ without and $63.9$ with.
Thus detection provides box confinement (\eqn{eq:raster}), instance identities,
and structural conditioning, not a demonstrated general accuracy advantage on multiple objects.

\emph{Why one batched call rather than per-instance crops?} An obvious
alternative runs the segment stage once per detected instance on a crop of its
box, which gives the model more effective resolution per object at a cost that
grows linearly with instance count, against the fixed single call of
\secLabel\ref{sec:method}. It reaches $75.4$ cIoU ($71.8$ gIoU) at $14.2$
calls per image, the best ReasonSeg cIoU in the table. This is $2.8$ above the
topology-on row it modifies but only $0.7$ above the reported configuration at
the same cost, in part because most
ReasonSeg items have one instance, making one crop equivalent to one call.
On RefCOCO, however, it scores $78.3$ cIoU at $15.4$ calls, a point and a half
\emph{below} the reported configuration ($79.8$). The advantage does not carry over to the second
split.

\emph{Why one batched refine rather than per-instance crops?} The same crop
alternative applied to stage~3: keep the batched segment call, then run the
refinement loop once per instance on a crop of its box, each with its own
overlay. Cost again grows with instance count. The batched refine of
\secLabel\ref{sec:method} stays two calls per pass regardless of $N$. This is
the most expensive variant in the table, $21.3$ calls per image against
$14.6$ for the reported configuration, a $46\%$ increase, for $74.9$ cIoU
against $74.7$, and on RefCOCO it
costs two points ($77.7$ against $79.8$). Its cost
would grow further on multi-instance benchmarks, where the batched refine's
cost does not.

\emph{Why B-splines rather than polygons?} Asking for a dense polygon instead
of spline control points is the natural baseline for our output primitive: it
removes the cubic smoothness prior and allows more coordinates per object. The
variant uses degree-one loops in both stages, with prompted vertex ranges of
$32$ to $160$ initially and $6$ to $24$ in refinement, against $16$ to $64$ and
$4$ to $12$ for the B-spline topology-on row, so this comparison matches neither coordinate
nor token budgets. At similar call cost, the final masks score comparably
($69.8$ gIoU and
$73.1$ cIoU against $69.2$ and $72.6$, with $78.7$ and $79.1$ on RefCOCO), and so do
the stage-2 masks over three runs ($65.3/66.2$ against $65.1/68.3$ gIoU/cIoU).
The ablation is self-limiting because the backend chooses the degree itself
(\secLabel\ref{sec:method}) and chooses $1$ (the polygon) for $125$ of the
$195$ objects it traced. Thus, on most of the split, the ablation changes the
vertex budget and placement rules rather than the primitive. By ground-truth
area, stage-2 polygon traces lead in the smaller two strata, while the
B-spline run leads in the largest quartile on both metrics ($75.6$ against
$73.4$ gIoU and $76.0$ against $69.7$ cIoU). As a pixel-mass metric,
cIoU is dominated by large objects, which is where the B-spline run leads. On
cubic-traced items, it does so with fewer coordinates: it emits
$28.9$ per image, compared with $48.5$ for the polygon variant, a
$1.7\times$ difference. \figLabel\ref{fig:qual_polygon_supp}
shows three examples. These results support a compact parametric
representation, not a uniform or equal-budget accuracy advantage.

\mysection{A different split, the same backend} The rankings of the fifteen
configurations (the reference and the fourteen variants, the same rows in both
tables) by final gIoU agree
between ReasonSeg and RefCOCO val ($\rho = 0.71$, $p = 0.003$), so a variant that helps on one split is likely
to help on the other.

\begin{table}[h]
  \centering
  \small
  \setlength{\tabcolsep}{6pt}
  \begin{tabular}{l cc cc}
  \toprule
  \textbf{Variant} & gIoU & cIoU & calls/image & s/image \\
  \midrule
  \rowcolor[HTML]{EFEFEF}
  \ours (reported configuration)              & 70.7\,\sd{0.74} & 74.7\,\sd{1.62} & 14.6\,\sd{0.32} & 177\,\sd{9} \\
  \midrule
  w/o deterministic postprocessing            & 70.2\,\sd{0.27} & 74.1\,\sd{2.20} & 14.8\,\sd{0.37} & 162\,\sd{12} \\
  + box-adjust pre-pass                       & 70.0\,\sd{0.27} & 70.2\,\sd{0.38} & 15.7\,\sd{0.34} & 188\,\sd{17} \\
  w/o postprocessing + box-adjust             & 69.5\,\sd{0.11} & 73.2\,\sd{1.64} & 15.9\,\sd{0.10} & 169\,\sd{12} \\
  \midrule
  + topology conditioning                     & 69.2\,\sd{1.34} & 72.6\,\sd{0.23} & 14.3\,\sd{0.59} & 175\,\sd{17} \\
  \quad w/o refinement ($T{=}0$)              & 63.8\,\sd{0.68} & 67.2\,\sd{2.72} & 2.0\,\sd{0.00} & 24\,\sd{2} \\
  \quad w/o morphology conditioning           & 69.9\,\sd{0.47} & 73.4\,\sd{2.69} & 14.9\,\sd{0.32} & 179\,\sd{7} \\
  \quad low backend reasoning effort          & 68.5\,\sd{0.30} & 72.3\,\sd{1.59} & 14.4\,\sd{0.58} & 133\,\sd{6} \\
  \quad low image detail                      & 67.2\,\sd{0.88} & 71.5\,\sd{4.60} & 14.6\,\sd{0.22} & 180\,\sd{6} \\
  \quad w/o deterministic postprocessing      & 69.6\,\sd{0.55} & 73.8\,\sd{1.89} & 14.8\,\sd{0.36} & 172\,\sd{29} \\
  \quad w/o postprocessing + box-adjust       & 69.4\,\sd{0.69} & 73.6\,\sd{3.55} & 15.5\,\sd{0.32} & 166\,\sd{12} \\
  \midrule
  \quad w/o detect stage (splines over whole image) & 69.5\,\sd{0.59} & 72.1\,\sd{0.87} & 8.8\,\sd{0.26} & 120\,\sd{19} \\
  \quad per-instance cropped calls            & 71.8\,\sd{0.44} & 75.4\,\sd{0.44} & 14.2\,\sd{0.24} & 169\,\sd{3} \\
  \quad polygon loops instead of B-splines    & 69.8\,\sd{0.57} & 73.1\,\sd{3.16} & 14.7\,\sd{0.09} & 186\,\sd{7} \\
  \quad per-instance cropped refine           & 71.7\,\sd{0.23} & 74.9\,\sd{0.91} & 21.3\,\sd{0.49} & 185\,\sd{3} \\
  \bottomrule
  \end{tabular}
  \caption{\small \textbf{Ablations on ReasonSeg val} (all 200 items, \gfl backend,
  $T{=}12$). The shaded row is the configuration reported in the paper (topology
  conditioning off). Cells are three-run means and sample deviations. \emph{s/image} is backend-call wall-clock time per image.}
  \label{tab:ablations}
  \end{table}
  
  \begin{table}[h]
  \centering
  \small
  \setlength{\tabcolsep}{6pt}
  \begin{tabular}{l cc c}
  \toprule
  \textbf{Variant} & gIoU & cIoU & calls/image \\
  \midrule
  \rowcolor[HTML]{EFEFEF}
  \ours (reported configuration)              & 79.3 & 79.8 & 17.5 \\
  \midrule
  w/o deterministic postprocessing            & 76.6 & 77.9 & 17.1 \\
  + box-adjust pre-pass                       & 79.2 & 79.7 & 18.4 \\
  w/o postprocessing + box-adjust             & 77.9 & 78.3 & 19.1 \\
  \midrule
  + topology conditioning                     & 76.7 & 77.3 & 16.8 \\
  \quad w/o refinement ($T{=}0$)              & 66.9 & 64.6 & 2.0 \\
  \quad w/o morphology conditioning           & 79.1 & 78.8 & 17.0 \\
  \quad low backend reasoning effort          & 75.7 & 74.3 & 16.6 \\
  \quad low image detail                      & 74.8 & 75.4 & 15.8 \\
  \quad w/o deterministic postprocessing      & 76.4 & 76.8 & 16.8 \\
  \quad w/o postprocessing + box-adjust       & 76.9 & 77.7 & 18.4 \\
  \midrule
  \quad w/o detect stage (splines over whole image) & 78.5 & 78.8 & 10.4 \\
  \quad per-instance cropped calls            & 78.8 & 78.3 & 15.4 \\
  \quad polygon loops instead of B-splines    & 78.7 & 79.1 & 16.6 \\
  \quad per-instance cropped refine           & 77.4 & 77.7 & 18.8 \\
  \bottomrule
  \end{tabular}
  \caption{\small \textbf{Ablations on RefCOCO val} (all 200 items, \gfl backend,
  $T{=}12$, one run per configuration). Rows and blocks as in the ReasonSeg
  ablation: the shaded row is the reported configuration.}
  \label{tab:ablations_refcoco}
  \end{table}
  
\begin{figure}[h]
  \centering
  \setlength{\tabcolsep}{1pt}
  \setlength{\qgw}{0.185\linewidth}
  \renewcommand{\arraystretch}{1.0}
  \begin{tabular}{@{}m{0.20\linewidth}@{\hspace{3pt}}cccc@{}}
    & \qhd{Input} & \qhd{Ground truth} & \qhd{Polygon loops} & \qhd{B-splines (\ours)} \\[1pt]
    \qlab{ReasonSeg}{\ldots what would the speaker use to amplify their voice?} & \qim{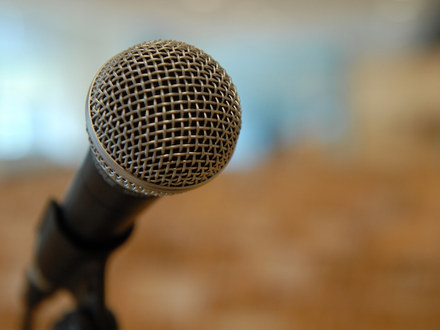} & \qim{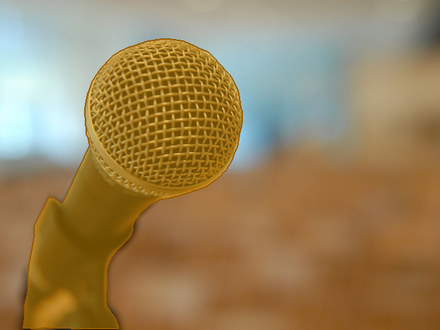} & \qim{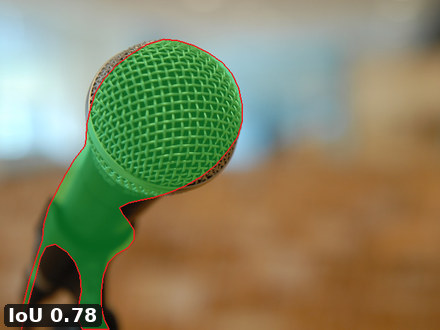} & \qim{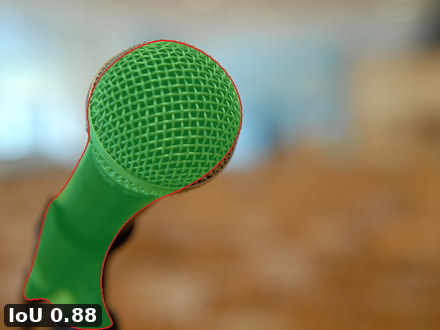} \\
    \qlab{ReasonSeg}{\ldots what could be used to rotate or spin other parts?} & \qim{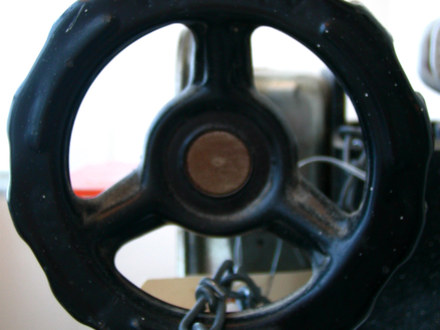} & \qim{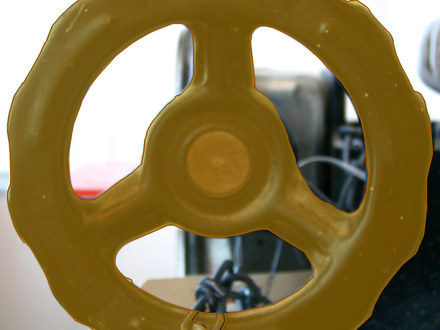} & \qim{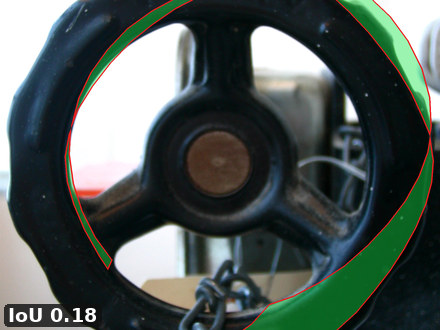} & \qim{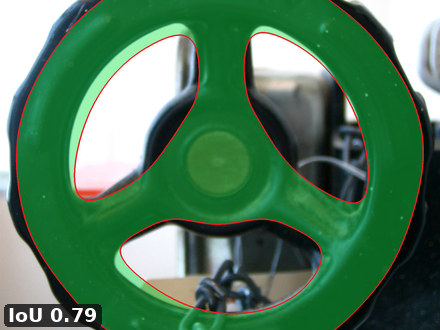} \\
    \qlab{ReasonSeg}{\ldots what part of the lion is a defining characteristic of males?} & \qim{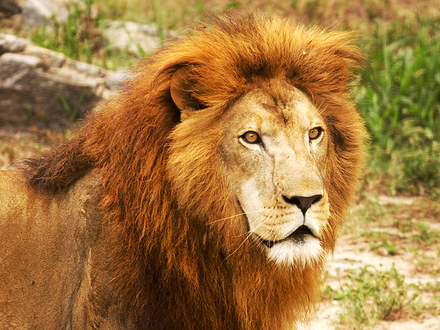} & \qim{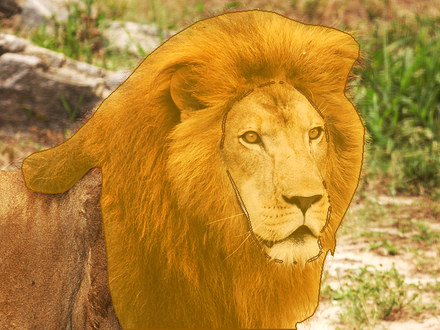} & \qim{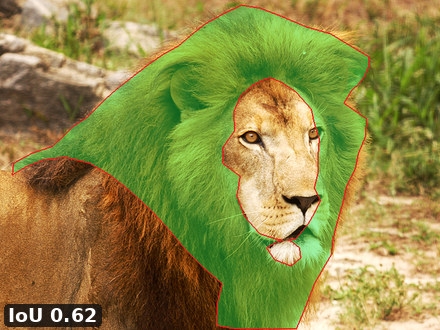} & \qim{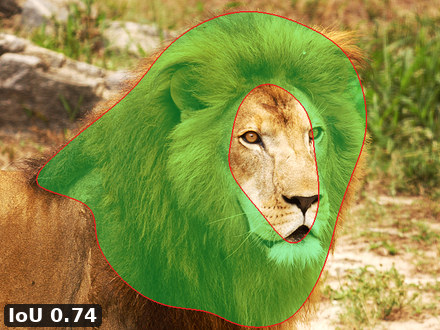} \\
  \end{tabular}
  \vspace{-6pt}
  \caption{\small \textbf{Polygons vs. B-splines in \ours.} Stage-2
  traces from the polygon-loop ablation and the B-spline run on identical items
  with the same backend. Each chip shows that run's recorded stage-2 IoU. Cubic B-splines cover curved boundaries better than polygons with fewer control points.}
  \label{fig:qual_polygon_supp}
\end{figure}

\subsection{Where the ground truth has holes} \label{app:holes}
\begin{figure}[ht]
  \centering
  \setlength{\tabcolsep}{1pt}
  \begin{tabular}{@{}ccccc@{}}
    \includegraphics[width=0.195\linewidth]{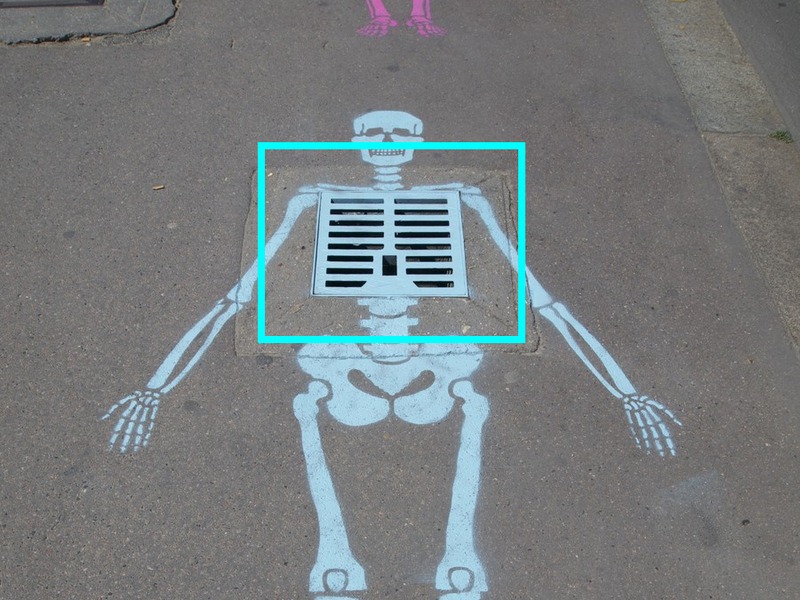} &
    \includegraphics[width=0.195\linewidth]{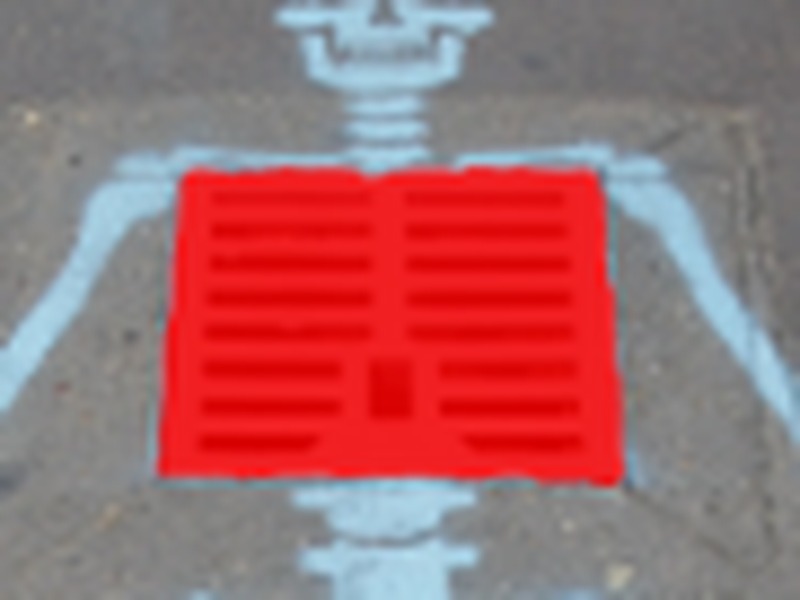} &
    \includegraphics[width=0.195\linewidth]{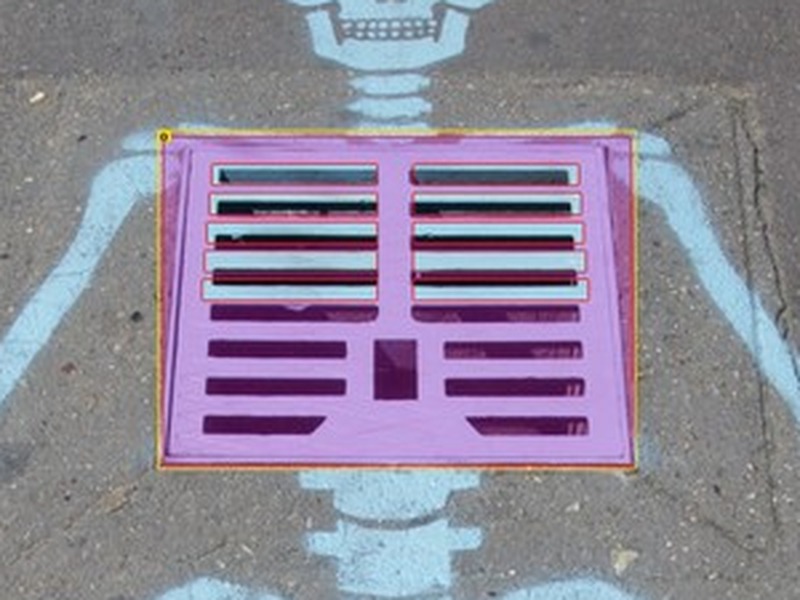} &
    \includegraphics[width=0.195\linewidth]{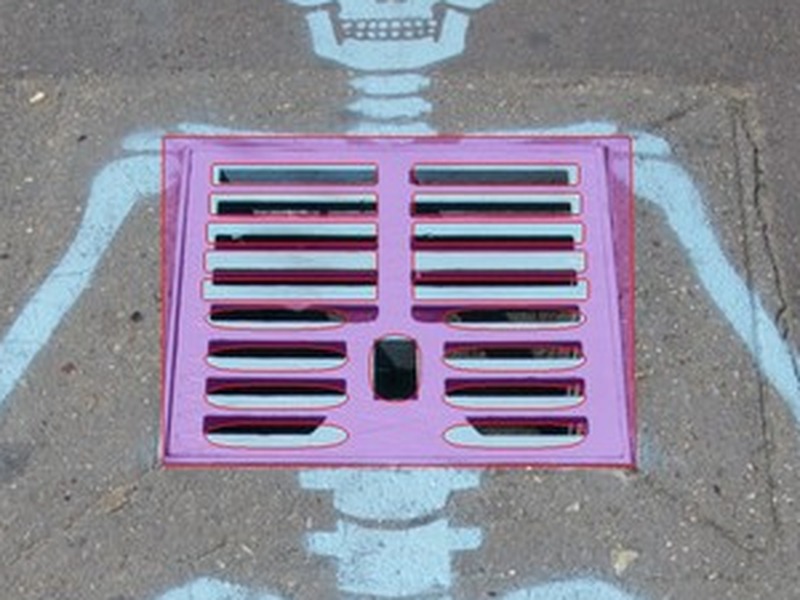} &
    \includegraphics[width=0.195\linewidth]{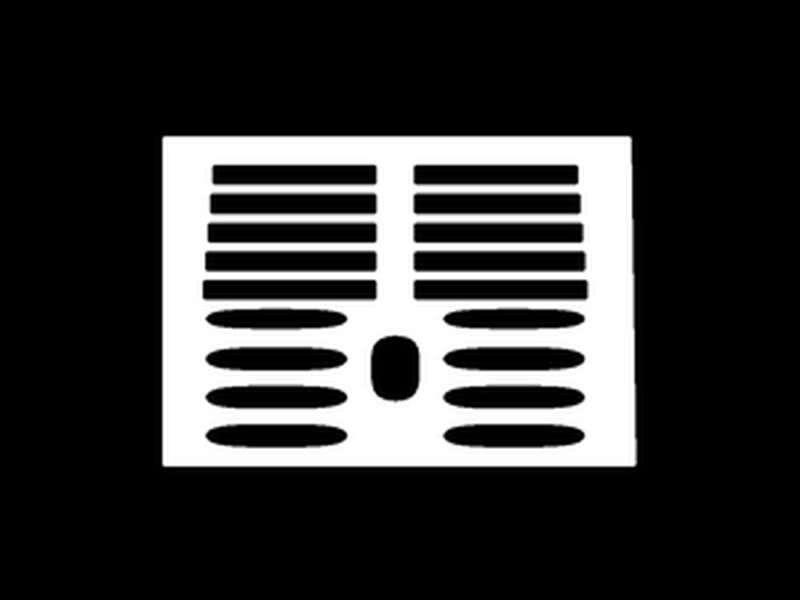} \\
    \small input & \small \samthree & \small w/o refine & \small \ours & \small \ours mask \\
  \end{tabular}
  \vspace{-6pt}
  \caption{\small \textbf{Explicitly carved holes.} The query is \emph{``the
  sewage outlet''}. The cyan-box region is enlarged. \samthree fills the grate
  solidly, whereas \ours carves one hole per slot.}
  \label{fig:topology}
\end{figure}
\vspace{-2pt}

The topology ablation of \tblLabel\ref{tab:ablations} removes the hole and
component description from all three prompts and produces no measurable change
($70.7$ vs.\ $69.2$ gIoU, within the run-to-run band). That average is taken
over a split where almost no target has a hole, so it cannot distinguish
``topology does not help'' from ``too few items benefit for the effect to show
up.'' We measure each item's ground truth directly to find the holed
targets. A hole is an enclosed background component that never touches the
image border and covers at least $0.5\%$ of the object's area. Anything smaller
is annotation speckle. Items with an empty ground truth are excluded, since a hole
needs an object.

The first finding is prevalence. Over the eleven splits measured,
holed targets are $7.1\%$ of ReasonSeg val and $6.6\%$ of ReasonSeg test,
$4.7\%$ of LVIS positives, and between $0\%$ and $1.4\%$ of every
RefCOCO/RefCOCO+/RefCOCOg split: COCO's polygon annotations essentially never
carve one. The subset the topology machinery exists for is a small minority of
what these benchmarks contain, which helps explain the null split-level
ablation.

The second finding closes the ablation. On the $396$ ReasonSeg items scored by
both the topology-on and topology-off runs, the mean per-item IoU
difference is $-0.3$ ($[-1.8, 1.2]$) overall and $+0.1$ ($[-5.3, 5.3]$) on the
$19$ holed ones: targeting the subset does not reveal a hidden effect. This is
the measurement behind the conclusion in \secLabel\ref{sec:analysis}:
topology conditioning makes explicit hole-loop specification possible when a
target has holes, but it is not a source of the accuracy we report, and we do
not claim it as one. Two caveats limit this result's interpretation: the runs
also differ in pass budget (\secLabel\ref{app:config}), and $19$ holed items
make up a small sample, as reflected in the interval.

\subsection{Localization or boundary?} \label{app:apiou}
\vspace{-2pt}

\begin{figure}[t]
  \centering
  \includegraphics[width=\textwidth]{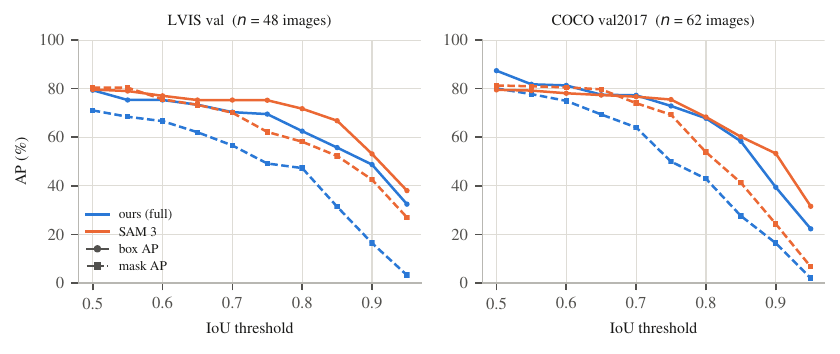}
  \vspace{-6pt}
  \caption{\small \textbf{\ours localizes like \samthree and loses the boundary.}
  COCO-style AP at each IoU threshold on AP-eligible images in the matched
  LVIS and COCO subsets, with boxes taken from the predicted masks.
  \ours is the reported configuration. Solid lines are box AP and dashed
  lines mask AP. The box curves run together while the mask curves fan apart, indicating good localization but loose boundary for \ours errors.}
  \label{fig:apiou}
\end{figure}

\tblLabel\ref{tab:openvocab} reports the gap to \samthree as two scalars per
benchmark, each averaged over the ten COCO IoU thresholds
$0.50{:}0.05{:}0.95$. This averaging hides the shape of the effect: a system
that finds every object but traces it loosely can report the same mean AP as
one that finds fewer objects but traces them tightly. Their behavior becomes
clear from how AP changes \emph{across the threshold axis}. To show this,
\figLabel\ref{fig:apiou} plots AP at each threshold for the main-results
pipeline configuration (\gfl, $T{=}8$, topology conditioning off), using the
same matched subsets and COCOeval pass.

At AP$_{50}$ on LVIS, box scores are nearly tied ($79.3$ vs.\ $79.6$). On
COCO, \ours leads in box AP ($87.4$ vs.\ $79.5$) and is close in mask AP
($80.0$ vs.\ $81.3$). The mask AP gap widens at tighter thresholds. By IoU
$0.90$, \ours scores $16.4$ on both benchmarks, while \samthree scores $42.5$
and $24.3$, respectively. Averaged over the ten thresholds, the mask gap is
$3$ to $6\times$ the box gap. On COCO, the interval for the box gap includes
zero. This pattern is consistent with masks that identify the right object
and place it within the right box, but have filled, closed-spline boundaries
that remain a few pixels loose. A threshold sweep reveals this boundary error,
whereas mean AP obscures it. Refinement improves the boundary but does not
eliminate the error: relative to the unrefined stage-2 mask, mask AP improves
by $+10.4$ on LVIS and $+12.5$ on COCO across thresholds up to $0.85$, peaking
at $+14$ to $+20$ in the $0.65$ to $0.85$ band. By IoU $0.95$, the gains fall
to $+2.4$ and $+1.5$, respectively. Refinement moves the outline substantially
closer to the object but does not produce pixel-tight masks. The remaining gap
is concentrated at high overlap thresholds. These subset results do not
establish full-dataset AP parity.

\subsection{What a mask costs} \label{app:cost}
\vspace{-2pt}

\begin{table}[h]
  \centering
  \footnotesize
  \setlength{\tabcolsep}{14pt}
\begin{tabular}{l|ccc|c}
\toprule
& \multicolumn{3}{c|}{measured / estimated cost per image} & \\
Method & calls & median & est.\ \$ & final \\
& MLLM\,/\,SAM\,3 & s & /\,1k img & gIoU \\
\midrule
\rowcolor[HTML]{EFEFEF} \ours (GPT-6 Astra) & 11.7\,/\,0 & 218 & 323.9 & 72.9 \\
\rowcolor[HTML]{EFEFEF} \ours (Gemini 3.7 Flash) & 15.0\,/\,0 & 153 & 15.6 & 69.4 \\
\rowcolor[HTML]{EFEFEF} \ours (Gemini 3.8 Flash) & 16.2\,/\,0 & 419 & 33.7 & 66.7 \\
\rowcolor[HTML]{EFEFEF} \ours (Gemini 3.5 Flash) & 17.6\,/\,0 & 246 & 77.9 & 66.2 \\
\rowcolor[HTML]{EFEFEF} \ours (GPT-5.6 Sol) & 15.0\,/\,0 & 665 & 83.0 & 61.6 \\
\rowcolor[HTML]{EFEFEF} \ours (Gemini 3 Flash) & 24.4\,/\,0 & 588 & 36.0 & 58.6 \\
\rowcolor[HTML]{EFEFEF} \ours (Gemma 4 31B) & 21.8\,/\,0 & 1883 & 5.0 & 56.2 \\
\rowcolor[HTML]{EFEFEF} \ours (Gemma 4 26B) & 19.4\,/\,0 & 2151 & 3.7 & 56.1 \\
\rowcolor[HTML]{EFEFEF} \ours (GPT-5.6 Terra) & 10.6\,/\,0 & 264 & 62.3 & 55.9 \\
\rowcolor[HTML]{EFEFEF} \ours (GPT-5.6 Luna) & 14.6\,/\,0 & 368 & 8.6 & 55.1 \\
\rowcolor[HTML]{EFEFEF} \ours (Qwen3.8 27B) & 12.3\,/\,0 & 847 & 15.4 & 51.0 \\
\rowcolor[HTML]{EFEFEF} \ours (GLM 5.3 Flash) & 7.3\,/\,0 & 1100 & 1.4 & 49.6 \\
\rowcolor[HTML]{EFEFEF} \ours (Qwen3.5 397B) & 12.4\,/\,0 & 2920 & 14.3 & 49.5 \\
\rowcolor[HTML]{EFEFEF} \ours (Claude Opus 5) & 23.1\,/\,0 & 706 & 321.2 & 46.6 \\
\rowcolor[HTML]{EFEFEF} \ours (Gemini 3.1 Flash-Lite) & 23.8\,/\,0 & 100 & 17.6 & 44.5 \\
\rowcolor[HTML]{EFEFEF} \ours (Muse Spark 1.2) & 24.4\,/\,0 & 837 & 76.2 & 43.2 \\
\rowcolor[HTML]{EFEFEF} \ours (Gemini 2.5 Flash-Lite) & 21.4\,/\,0 & 113 & 5.6 & 28.8 \\
\rowcolor[HTML]{EFEFEF} \ours (GPT-5.4) & 24.0\,/\,0 & 211 & 176.8 & 27.1 \\
\rowcolor[HTML]{EFEFEF} \ours (Qwen3 VL 235B) & 9.0\,/\,0 & 128 & 6.6 & 25.1 \\
\rowcolor[HTML]{EFEFEF} \ours (Muse Glimmer 30B) & 14.7\,/\,0 & 1478 & 13.6 & 23.5 \\
\rowcolor[HTML]{EFEFEF} \ours (GLM 4.6V) & 15.7\,/\,0 & 1046 & 11.4 & 14.2 \\
\midrule
\samagent (Gemma 4 31B) & 2.3\,/\,1.4 & 27 & 6.4 & 73.1 \\
\samagent (Gemini 3.5 Flash) & 2.5\,/\,1.5 & 13 & 17.5 & 72.5 \\
\samagent (Gemini 3.7 Flash) & 2.4\,/\,1.4 & 11 & 8.3 & 72.4 \\
\samthree & 0\,/\,1 & 2 & 4.3 & 13.1 \\
\bottomrule
\end{tabular}
    \caption{\small \textbf{What a mask costs.} ReasonSeg validation set (200 items),
  covering all 21 \ours backends and the \samthree and \samagent baselines.
  Call counts and median latency are measured per image. Estimated spend is per
  1,000 images, using one token budget shared across all rows and a fixed
  serving price per \samthree forward pass. \ours rows are shaded.}
  \label{tab:cost}
\end{table}

\mysection{Where the time goes}
Refinement dominates: stage 3 accounts for $76$ to $94\%$ of wall-clock time on
every backend ($89\%$ on \gfl), consistent with a loop of up to $T{=}12$ passes
at two calls per pass, compared with one detection and one segmentation call.
The cost comes from the number of calls, not their content: at a published
$2.3$\,s time-to-first-token, \gfl spends about $34$ of its $153$ median seconds
waiting for first tokens alone across its $15$ calls. Call count is not a proxy
for useful effort: the best backend averages $11.7$ calls per image, while the
three highest-call backends average $24.0$ to $24.4$ and score $58.6$, $43.2$
and $27.1$. This reflects stopping behavior: a sequence may stop while its mask
is still changing, while an unproductive sequence may run to the budget.

\mysection{\ours is the expensive way to reach this quality}
On this benchmark \samagent is both better and cheaper. Driven by the same
\gfl, it reaches $72.4$ gIoU against our $69.4$ in roughly one-fourteenth of
the wall-clock time ($11$\,s against $153$\,s) and at just over half the estimated
serving cost ($\$8.31$ against $\$15.61$ per thousand images). It makes $2.4$
agent rounds and $1.4$ \samthree passes where we make $15$ vision calls, and
the mask itself comes from a $860$\,M-parameter forward pass rather than from
coordinates the MLLM has to write out. Charging for the
\samthree passes narrows that gap rather than closing it: the agent's
$1.4$ passes cost it $\$5.85$ of its $\$8.31$, and \samthree on its own costs
$\$4.30$, a cost undercut by only two \ours backends. That advantage is
specific to what \samthree can already segment: \samagent inherits \samthree's
mask, so where \samthree collapses (\secLabel\ref{sec:results}) the agent
collapses with it, while \ours is bounded only by its detect box. In summary, predicting a mask as text buys open-vocabulary reach at the
price of calls, latency and dollars, and the price is set by the backend
rather than by the pipeline: across the twenty-one backends, spend per thousand
images ranges from $1.4$ to $324$, a factor of $238$, across a gIoU range of
$14.2$ to $72.9$, with cost only weakly tracking gIoU. The most expensive
backend is also the best, but costs $21\times$ as much as the second-best for
a $3.5$-point gIoU gain, and the two cheapest backends outscore nine dearer ones.

\subsection{Where the LLM route wins and loses} \label{app:limitations}
\vspace{-2pt}

The results support a consistent picture. \ours wins where (i) the query
needs \emph{reasoning} (ReasonSeg, \figLabel\ref{fig:reasoning_qual}),
(ii) the domain is \emph{rare} (RF-100VL, $2.2\times$ \samthree's AP$_0$),
including medical and industrial imagery, (iii) \emph{presence} judgment
matters (hard-negative \cgf on LVIS and COCO), where the LLM can reason about
whether the concept is in the image. This does not carry over to SA-Co/Gold,
where \ours trails \samthree by $9.0$ \cgf while \pmf is level, so the loss
lies in the presence decision. The last case is (iv) a \emph{regular shape}:
such shapes admit compact cubic or straight-edged traces, and, with topology
enabled, holes can be specified by interior loops. Neither final boundary
smoothness nor hole count is guaranteed. Holed targets are a small
minority in these benchmarks (see \figLabel\ref{fig:topology}), and prevalence is quantified in \secLabel\ref{app:holes}. \ours loses on pixel-tight mask AP
and on thin, amorphous or highly irregular regions
(\figLabel\ref{fig:failures}), where compact geometric traces lose fine
boundary detail. In practice, \ours takes minutes per image and makes multiple
API calls, compared with a median of $2$\,s per image for \samthree on a local GPU (\tblLabel\ref{tab:cost}). However, it runs on the heavily
optimized LLM serving stack, needs no GPU of its own, and produces inspectable
intermediate records (boxes, optional topology, per-loop splines, and focus
boxes with a type and a short shape description). These records are not verified explanations of the
model's internal reasoning.

\mysection{Limitations}
\ours inherits its backend's failure modes: hallucinated detections on
ambiguous prompts, occasional coordinate-format instability (mitigated but
not eliminated by the repairs of \S\ref{app:pipeline}), and per-call latency
of seconds to minutes, which makes video and real-time use out of scope.
Spline loops are a poor primitive for thin or fractal structures
(\figLabel\ref{fig:failures}), and segmentation quality is upper-bounded by
detect-box quality due to the confinement invariant. Finally, API-served
backends make exact reproducibility subject to provider model updates. We
pin model versions and release all records.

\subsection{Failure modes} \label{app:failures}
\vspace{-2pt}

\begin{figure}[!ht]
  \centering
  \begin{subfigure}[b]{0.235\textwidth}
    \includegraphics[width=\linewidth]{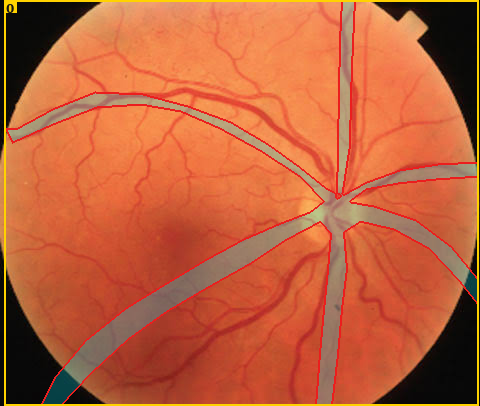}
    \caption*{w/o refine}
  \end{subfigure}\hfill
  \begin{subfigure}[b]{0.235\textwidth}
    \includegraphics[width=\linewidth]{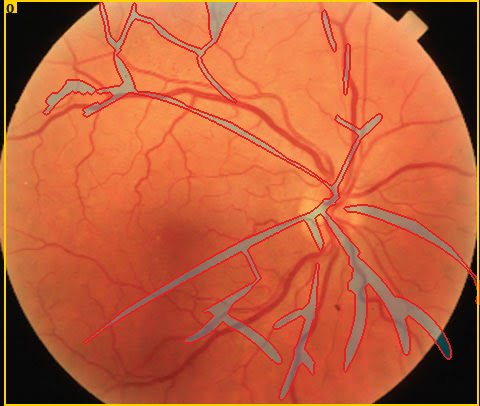}
    \caption*{full (7 passes)}
  \end{subfigure}\hfill
  \begin{subfigure}[b]{0.245\textwidth}
    \includegraphics[width=\linewidth]{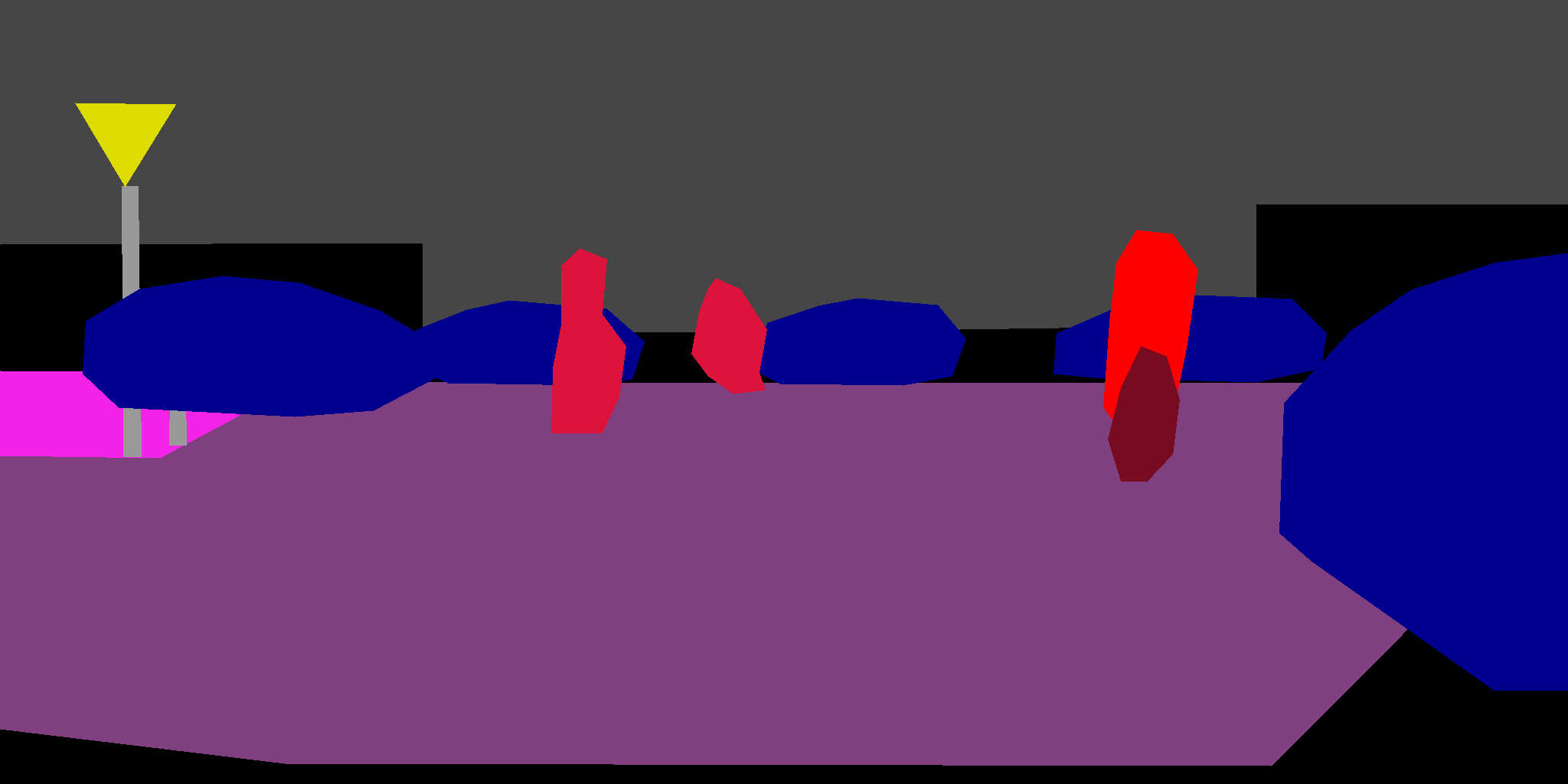}
    \caption*{\ours label map}
  \end{subfigure}\hfill
  \begin{subfigure}[b]{0.245\textwidth}
    \includegraphics[width=\linewidth]{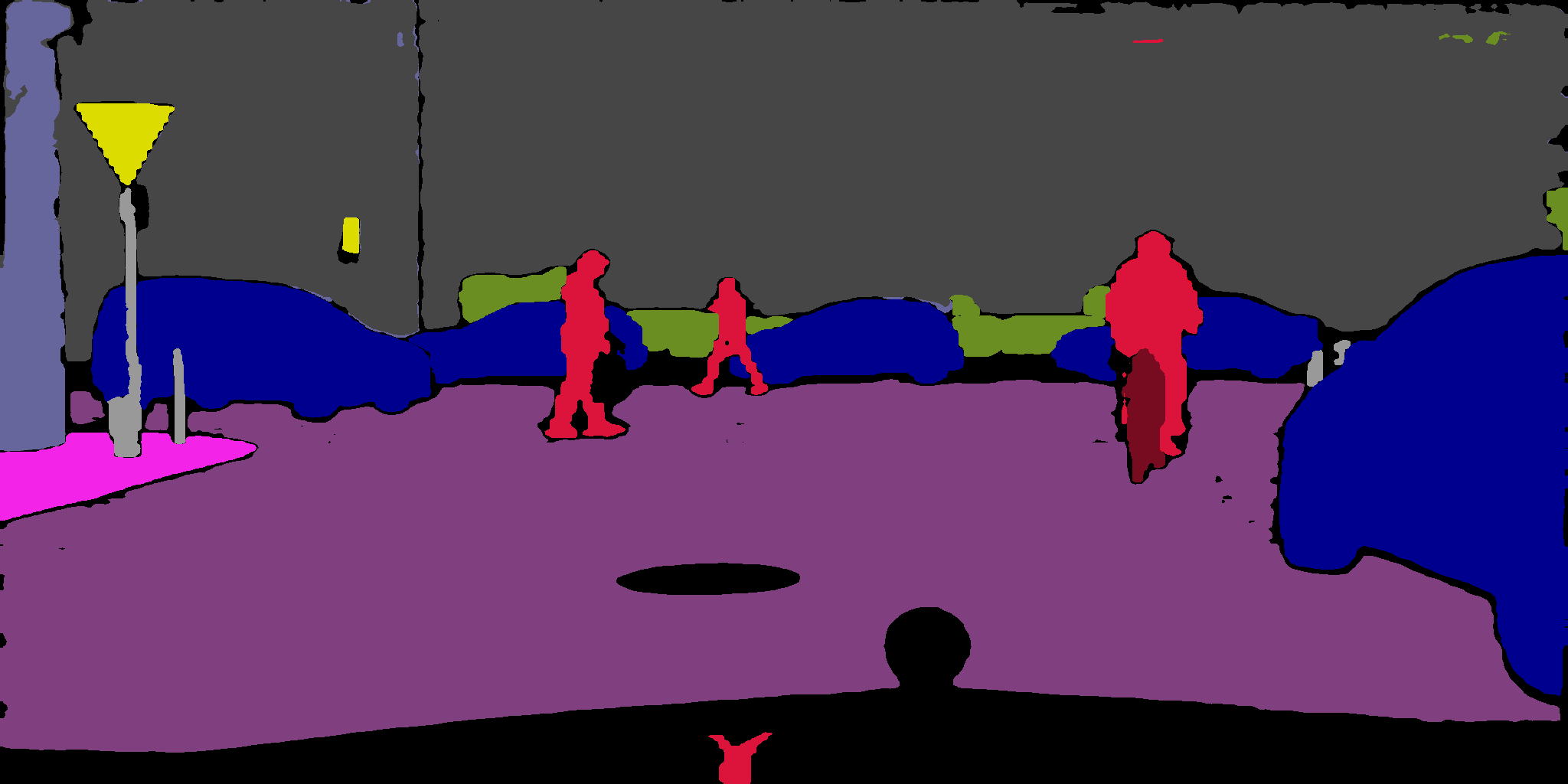}
    \caption*{\samthree label map}
  \end{subfigure}\\[4pt]
  \begin{subfigure}[b]{0.235\textwidth}
    \includegraphics[width=\linewidth]{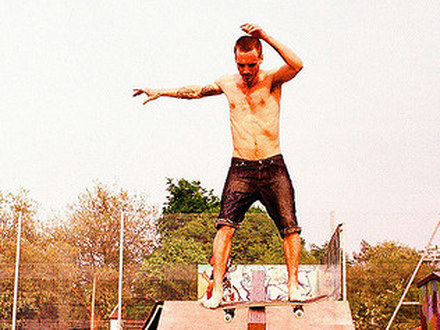}
    \caption*{input}
  \end{subfigure}\hfill
  \begin{subfigure}[b]{0.235\textwidth}
    \includegraphics[width=\linewidth]{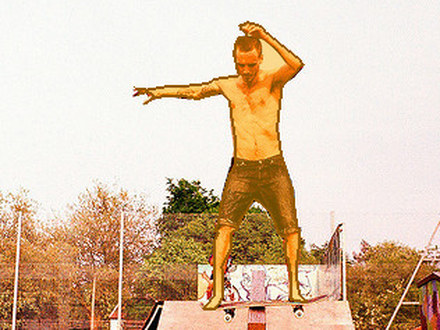}
    \caption*{ground truth}
  \end{subfigure}\hfill
  \begin{subfigure}[b]{0.245\textwidth}
    \includegraphics[width=\linewidth]{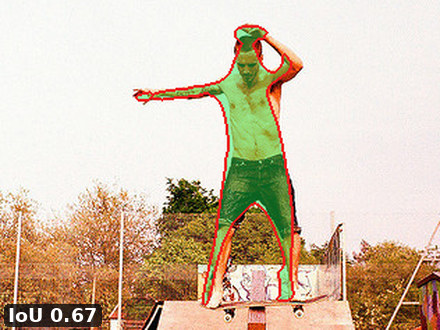}
    \caption*{\ours}
  \end{subfigure}\hfill
  \begin{subfigure}[b]{0.245\textwidth}
    \includegraphics[width=\linewidth]{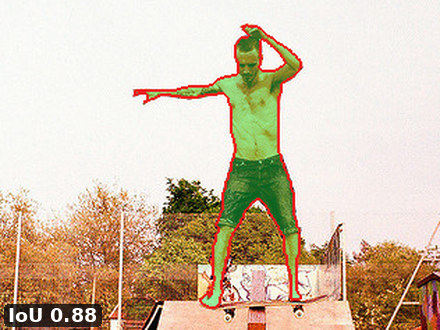}
    \caption*{\samthree}
  \end{subfigure}
  \vspace{-6pt}
  \caption{\small \textbf{Failure modes.} Top left pair: thin structures (retinal vessels)
  defeat spline outlines even after refinement. Top right pair: on Cityscapes, amorphous and heavily-occluded background
  classes favour pixel decoders over compact parametric loops. Bottom row: the
  query is grounded correctly, but a closed loop cannot cut into the deep,
  narrow concavities of an articulated hands.}
  \label{fig:failures}
\end{figure}

\figLabel\ref{fig:failures} collects the failure modes discussed in
\secLabel\ref{app:limitations}: thin structures, amorphous background
classes, and the deep, narrow concavities of articulated bodies.

\clearpage
\section{Prompt Templates} \label{app:prompts}
\vspace{-4pt}

This section gives the three prompt templates that drive the pipeline, in the
form the model receives them: box~1 is the detection call
of \secLabel\ref{sec:method}, box~2 the initial tracing call, and box~3
the focus-box proposal call. Text in angle brackets is filled in at run time. The
instance and focus-region blocks are shown filled for a two-instance example,
since their layout is part of what the model reads. Long lines are re-wrapped
to the column width. Everything in black is template text. The
{\color{PromptNote}\texttt{\#}}-prefixed green lines are ours, not the
model's, so a box can be copied whole and the notes stay comments. The green
leading line stands for the image, which is attached to the request
alongside the prompt text. The boxes show the study configuration, which
includes the stage-1 topology description. The reported configuration drops
that description from all three prompts: the topology block of box~1's schema,
the topology lines of the instance and detection-box listings in boxes~2 and~3,
and the sentences that refer to topology. The fourth call, the re-tracing half of a
refinement pass, is box~2's spline rules restated
once per focus region. It repeats the coordinate paragraph and the overlay
description of box~3, lists the surviving focus regions with their type and
reported shape, fixes the degree to cubic, asks for four to twelve control
points per region instead of sixteen to sixty-four, and returns the same
id-keyed list of closed loops.

The templates share three properties. First,
every call states the coordinate space explicitly and every call that returns
points restates the box those points must fall inside as a numeric inequality.
The code clamps regardless, but a clamped-in point is a worse outline than a
correctly placed one. Second, each prompt offers an escape hatch, an empty
detection list, an empty loop list for an instance whose box holds nothing, an
empty proposal list for a pass that finds no error, so that a failure to find
something is expressible and does not have to be invented. Third, batched
calls pin the emission order to ascending instance id, which is what makes a
truncated reply salvageable.

\vspace{0.3em}
\begin{promptbox}{\normalfont\bfseries Prompt \#1: Instance Detection
(Stage 1)}
\begin{lstlisting}[style=promptstyle]
# <the input image>

The image is <W>x<H> pixels (origin top-left, x right, y down).
ALL coordinates are absolute PIXELS with 0 <= x <= <W> and 0 <= y <= <H>.

Find every DISTINCT instance of the following in the image: "<query>"

Return ONLY a JSON object (no prose, no markdown fences) with this exact schema:
{"detections": [{
  "id": <int starting at 0>,
  "label": "<short label>",
  "box": [x1, y1, x2, y2],
  "confidence": <0..1>,
  "morphology": {
    "shape": "<overall silhouette in a few words, e.g. 'compact rounded body with two upright
              protrusions'>",
    "aspect": "<tall|wide|square>",
    "convexity": "<convex|slightly-concave|highly-irregular>",
    "parts": ["<protruding/notable parts, e.g. 'left protrusion'>", "<'base contact'>", "..."]
  },
  "topology": {
    "components": <int number of disconnected pieces of the object silhouette>,
    "holes": <int number of enclosed background holes inside the silhouette>,
    "boundary_loops": <int total closed boundary curves = components + holes>,
    "notes": "<one short phrase, e.g. 'single blob, no holes'>"
  }
}]}

where box is [x1, y1, x2, y2] in the coordinate space described above (x1 < x2, y1 < y2, all values
inside the valid range for that space).
- One entry per DISTINCT instance -- do NOT return duplicate or nested boxes for the same object. Order
  detections in reading order (top-to-bottom, then left-to-right); ids will be reassigned by code, but
  keep them contiguous from 0.
- morphology.aspect must describe the instance as you SEE it, independently of the numbers you wrote in
  box; it is cross-checked against them.
- Make each box LARGE enough to ENTIRELY encompass that instance -- leave generous margin so no part of
  the silhouette (protrusions, extremities, soft edges, head, ears, paws) is clipped by the box. Prefer
  a roomy box over a tight crop; a later stage confines all outline tracing to this box, so an
  undersized box truncates the segmentation.
- morphology describes the object's FORM (shape of its outline) so a later stage can trace it; topology
  describes its STRUCTURE (how many separate blobs and how many holes) so a later stage knows how many
  closed curves to fit.
If nothing matches, return {"detections": []}.
Output the JSON and nothing else.
\end{lstlisting}
\end{promptbox}

\vspace{0.3em}
\begin{promptbox}{\normalfont\bfseries Prompt \#2: Outline Tracing
(Stage 2)}
\begin{lstlisting}[style=promptstyle]
# <the input image>

The image is <W>x<H> pixels (origin top-left, x right, y down).
ALL coordinates are absolute PIXELS with 0 <= x <= <W> and 0 <= y <= <H>.

We have already DETECTED <N> instance(s) of "<query>" (same query / class). Each instance has its own
bounding box; a later stage confines that instance's outline to ITS box only.

INSTANCES (trace EVERY one):
- id=0 label="<label>" box=[120, 80, 540, 610]
    every x,y you return for id=0 must satisfy: 120 <= x <= 540 and 80 <= y <= 610
    morphology: {"shape": "compact rounded body with two upright ears", "aspect": "tall",
                 "convexity": "slightly-concave", "parts": ["left ear", "tail", "front paws"]}
    topology:   {"components": 1, "holes": 0, "boundary_loops": 1, "notes": "single blob, no holes"}
- id=1 label="<label>" box=[600, 140, 930, 585]
    # ... (one such entry per detected instance)

Your job has TWO parts:
  (1) pick ONE shared overlay tint color for the mask (same for all instances),
  (2) CHOOSE a spline degree (1 or 3) shared by every instance of this class, then for EACH instance
      trace its silhouette with piece-wise CLOSED 2D B-splines of that degree so the spline(s) tightly
      SURROUND that instance.

MASK COLOR (part 1):
Pick ONE color name from this fixed list:
  "lime", "green", "spring_green", "mint", "teal", "cyan", "turquoise", "sky_blue", "blue",
  "periwinkle", "indigo", "violet", "purple", "magenta", "hot_pink", "pink", "coral", "orange",
  "chartreuse"
Choose the color that maximizes contrast of a light translucent tint against BOTH the "<query>" AND its
surrounding background, so a later pass can clearly see what is covered vs not. Rules for the choice:
- ONE tint is shared by every instance: pick a hue that reads clearly on ALL of them and against each
  one's own local background, not just the first.
- Prefer a hue that is NOT already dominant in the object or in the nearby background.
- Do NOT pick a color that would wash out or blend into the object's own colors or hide fine edge
  detail under the tint.
- Return the name EXACTLY as listed (lowercase, underscores).

SPLINE RULES (part 2):
- Return exactly one entry per instance, keyed by the same integer "id".
- Every control point for instance id=K MUST lie inside THAT instance's box=[x1, y1, x2, y2] listed
  above.
- Fit ONE closed B-spline per connected blob of that instance (topology.components), plus ONE
  additional closed B-spline per interior hole (topology.holes), marking holes with "role":"hole".
  Outer silhouette loops use "role":"outer".
- Use each instance's morphology to place control points on the true outline (follow protrusions,
  concavities, corners, and curves named in the morphology).
- If one instance's box is empty, unreadable, or holds nothing matching the "<query>", return an EMPTY
  "splines" list for that id rather than inventing an outline. Still return its id, and still trace
  every other instance normally.
- Your whole reply must fit in ONE response, and <N> instances is a lot of coordinates. Emit entries in
  ASCENDING id order and FINISH each entry before starting the next; stay near the LOW end of the
  control-point range for the simple ones so that nothing gets cut off.
CHOOSE THE SPLINE DEGREE (1 or 3) from the object's form, then place the control points the way that
degree requires:
- Pick DEGREE 1 (closed polyline) for angular / polygonal / straight-edged objects, or simple
  silhouettes where a polygon is enough. Its control points are vertices ON the true object outline
  (not a looser control polygon): place them directly on the edge, ordered around the loop, denser at
  corners/bends and sparser on long straight spans.
- Pick DEGREE 3 (closed cubic B-spline) for organic / rounded / smoothly curved silhouettes. Its
  control points are the B-spline CONTROL POLYGON (the curve is smoother than and sits inside the
  polygon), so push them slightly OUTWARD toward the object edge, ordered around the loop.
- Use the SAME degree for EVERY spline (outer loops and holes). Report it in the top-level "degree"
  field AND set "degree" on each spline to that value.
- Choose the NUMBER of control points from the object's morphology and what you actually see inside the
  box -- typically 16-64 per loop. Use MORE points for intricate / irregular / many-parted outlines
  (closer to 64) and FEWER for simple, smooth, or rounded ones (near 16). Do not aim for a fixed count
  -- let the shape's complexity decide within that range; under-sampling a complex outline is worse
  than a few extra points on a simple one.

Return ONLY this JSON (no prose, no markdown fences):
{"mask_color": "<one name from the list>",
  "mask_color_reason": "<short why this hue contrasts with object+background>",
  "degree": <1 or 3>,
  "degree_reason": "<short why this degree fits the object's form>",
  "instances": [
  {"id": 0, "splines": [
    {"role": "outer", "degree": <same as top-level degree>, "closed": true,
      "control_points": [[x, y], [x, y], ...]}
  ]}
]}
Output the JSON and nothing else.
\end{lstlisting}
\end{promptbox}

\vspace{0.3em}
\begin{promptbox}{\normalfont\bfseries Prompt \#3: Proposing Focus Regions
(Stage 3)}
\begin{lstlisting}[style=promptstyle]
# <the current overlay image>

This image shows the CURRENT segmentation of the "<query>":
  - the translucent <tint> tint marks the current MASK REGION (coverage),
  - the RED outline is the current mask BOUNDARY -- the main edge to compare,
  - each YELLOW rectangle is one instance's detection box (ids drawn beside them).
The <tint> fill is light on purpose: you can still see the <query>'s real edges/texture underneath. Use
the <tint> tint to tell what is already covered; use the red outline + the underlying pixels to judge
where the TRUE edge runs. Do NOT treat the soft tinted edge as the object boundary.
The image is <W>x<H> pixels (origin top-left, x right, y down).
ALL coordinates are absolute PIXELS with 0 <= x <= <W> and 0 <= y <= <H>.

DETECTION BOXES (each focus box MUST lie inside ONE of these):
- id=0 label="<label>" box=[120, 80, 540, 610]
    topology:   {"components": 1, "holes": 0, "boundary_loops": 1, "notes": "single blob, no holes"}
- id=1 label="<label>" box=[600, 140, 930, 585]
    # ... (one such entry per detected instance)

Your task is a COMPARISON: follow the RED outline and, at each part of it, look through the <tint> tint
at the actual object pixels to see where the TRUE edge of the <query> is. Report the places where the
two disagree.

Find FOCUS BOXES that mark where the mask is wrong and needs local fixing:
  - type "positive": a box that FULLY encloses an area that IS part of the <query> but is NOT tinted /
    lies OUTSIDE the red outline (MISSED, e.g. a protruding part or tip left outside the mask) -> to be
    ADDED.
  - type "negative": a box that FULLY encloses an area that IS tinted / INSIDE the red outline but is
    actually background (wrongly included, e.g. background bleeding beside the silhouette, or a gap
    that is not the <query>) -> to be REMOVED.
Check BOTH error directions along the red outline: missed object (positive) AND tinted background
(negative). Do not only hunt for misses.

Rules:
- Tag every focus box with "instance_id" naming which detection box it belongs to.
- Each focus box [x1, y1, x2, y2] MUST lie inside that instance's detection box.
- Make each box LARGE enough to ENTIRELY encompass the region to be fixed -- leave generous margin so
  no part of the miss / over-include is clipped by the box edge. Prefer one roomy box over a tight
  crop; undersized boxes cause the next-step spline to miss the true outline. Propose 1-6 boxes PER
  instance (up to 6 x <N> total), one per distinct problem area.
- Check EVERY one of the <N> instances listed above, not only the largest or the most obviously wrong
  one. Propose up to 6 boxes PER instance (do not dump the whole budget on one).
- Anchor each box on what you can actually SEE: it should cover the full gap between the red outline
  (or the tinted coverage) and the true edge of the <query> at that spot, plus margin. Place box edges
  on real image features (the object's boundary, a protrusion, the adjacent background), not on the
  soft tint alone.
- A pass may be ALL positive, ALL negative, or a mix -- there is NO requirement to include both types.
  Return only boxes that are genuinely wrong; never add a box of the other type just to have one of
  each.
- Only flag areas that are CLEARLY wrong in the current image; if a spot already looks correct, do not
  propose a box there (earlier fixes already show in the tint).
- For each box, describe the SHAPE of the region to be added/removed in a few words, and how intricate
  its outline is (e.g. "thin curved sliver", "small round bump", "jagged notch", "simple triangular
  wedge"). A later step uses this to decide how many control points to spend on the box, so be honest
  about whether the local edge is simple or wiggly.

Return ONLY this JSON (no prose, no markdown fences). Example shape (mix of types; your list may be
all-positive, all-negative, or mixed):
{"focus_boxes": [
  {"type": "positive", "instance_id": 0, "shape": "<shape + complexity of the region to add>",
    "box": [x1, y1, x2, y2]},
  {"type": "negative", "instance_id": 0, "shape": "<shape + complexity of the region to remove>",
    "box": [x1, y1, x2, y2]}
]}
Output the JSON and nothing else.
\end{lstlisting}
\end{promptbox}